\documentclass[11pt]{article}
\usepackage[utf8]{inputenc}
\usepackage[T1]{fontenc}
\usepackage{lmodern}

\usepackage{geometry}
\usepackage{authblk}

\usepackage{graphicx}
\usepackage{caption}
\usepackage{float}

\usepackage{amsmath, amssymb}

\usepackage{booktabs}
\usepackage{multirow}
\usepackage{makecell}

\usepackage{rotating}
\usepackage{pdflscape}

\usepackage[numbers,sort&compress]{natbib}
\usepackage{hyperref}

\usepackage{pdfpages} %append pdf file to the end (suppl)

\newcommand{\edit}[1]{\begingroup\color{black}#1\endgroup}

\hypersetup{
  colorlinks=true,
  linkcolor=blue,
  citecolor=blue,
  urlcolor=blue,
  pdftitle={Continual Learning with Single-Layer Extension in Fourier Neural Operators},
  pdfauthor={Mahmoud Elhadidy, Amirhoesseni Arzani}
}

\title{SLE-FNO: Single-Layer Extensions for Task-Agnostic Continual Learning in Fourier Neural Operators}

\author[1,2]{Mahmoud Elhadidy}
\author[3]{Roshan M. D'Souza}
\author[1,2,4]{Amirhossein Arzani\thanks{Corresponding author: \texttt{amir.arzani@sci.utah.edu}}}

\affil[1]{Department of Mechanical Engineering, University of Utah, Salt Lake City, UT, USA}
\affil[2]{Scientific Computing and Imaging Institute, University of Utah, Salt Lake City, UT, USA}
\affil[3]{Department of Mechanical Engineering, University of Wisconsin--Milwaukee, Milwaukee, WI, USA}
\affil[4]{Department of Biomedical Engineering, University of Utah, Salt Lake City, UT, USA}

\date{}
\begin{document}
\maketitle

\begin{abstract}

    Scientific machine learning is increasingly used to build surrogate models, yet most models are trained under a restrictive assumption in which future data follow the same distribution as the training set. In practice, new experimental conditions or simulation regimes may differ significantly, requiring extrapolation and model updates without re-access to prior data. This creates a need for continual learning (CL) frameworks that can adapt to distribution shifts while preventing catastrophic forgetting. Such challenges are pronounced in fluid dynamics, where changes in geometry, boundary conditions, or flow regimes induce non-trivial changes to the solution. Here, we introduce a new architecture-based approach (SLE-FNO) combining a Single-Layer Extension (SLE) with the Fourier Neural Operator (FNO) to support efficient CL. SLE-FNO was compared with a range of established CL methods, including Elastic Weight Consolidation (EWC), Learning without Forgetting (LwF), replay-based approaches, Orthogonal Gradient Descent (OGD), Gradient Episodic Memory (GEM), PiggyBack, and Low-Rank \edit{Adaptation} (LoRA), within \edit{a spatial field-to-field regression setting}. The models were trained to map transient concentration fields to time-averaged wall shear stress (TAWSS) in pulsatile aneurysmal blood flow. Tasks were derived from 230 computational fluid dynamics simulations grouped into four sequential and out-of-distribution configurations. Results show that replay-based methods and architecture-based approaches (PiggyBack, LoRA, and SLE-FNO) achieve the best retention, with SLE-FNO providing the strongest overall balance between plasticity and stability, achieving accuracy with zero forgetting and minimal additional parameters. Our findings highlight key differences between CL algorithms and introduce SLE-FNO as a promising strategy for adapting baseline models when extrapolation is required. 

\end{abstract}

\newpage

\section{Introduction}

Scientific machine learning (SciML) deals with combining machine learning with scientific data and physical models~\cite{thiyagalingam2022scientific}. In SciML models, a range of data availability from sparse to large data and various physical priors could be considered~\cite{karniadakis2021physics}. The data typically arise from numerical simulations and laboratory experiments and are governed by physical laws~\cite{brunton2022data}. A popular category of SciML problems focuses on developing surrogate models designed to emulate computationally costly numerical workflows. Recently, operator learning, due to its efficient incorporation of functional data, has become a popular approach for building surrogate models~\cite{li2020fourier,lu2021learning,kovachki2023neural}.  

While physical priors could be incorporated into deep learning architectures (e.g., physics-informed neural networks~\cite{raissi2019physics} and physics-informed neural operators~\cite{goswami2023physics}), most of these techniques do not incorporate physics in a by-construction manner and instead rely on soft constraints and regularizers. As a result, for the network to learn the physical equations, it needs to be probed across the desired physics. That is, the network is not guaranteed to respect the physical equations for all parameters and boundary/initial conditions unless it was probed during training. Unfortunately, training parametric physics-informed deep learning models is itself a challenging and computationally expensive task due to challenging optimization landscapes~\cite{wang2021understanding,kiyani2025optimizer}. Consequently, most surrogate models rely on purely data-driven strategies and large data often generated by numerical simulations.

As a more ambitious goal, inspired by the success of large language models, the community is starting to consider the concept of scientific foundation models (SciFM)~\cite{choi2025defining,menonscientific}. These are ideally scalable models trained across tasks and modalities, designed for transfer and fine-tuning to specific problems. However, we are still far from true practical SciFM models. At a simpler level, while the field has witnessed a large and growing body of work dedicated to training and testing neural surrogates, a simple question has largely been overlooked: So what is next? Trained neural surrogate models are rarely used in follow-up studies, after the original paper, to solve a practical problem of interest, even by the same group. Given tremendous progress over the past few years, the time is ripe to consider practical deployment questions related to SciML models.

A key challenge is that trained SciML models are trained under the assumption that inference data follow the same distribution as the training data~\cite{quinonero2008dataset}, an assumption that mostly holds only during carefully designed training, validation, and testing stages. Once the model is deployed, the input data provided by the user has to adhere to the same training distribution, which is difficult to control for multiple reasons. First, the training landscape may not be clearly defined, particularly in complex settings. Second, it is not clear if the user is interested in the same training landscape. Finally, in practical downstream tasks such as control, optimization, and design, the user does not necessarily have full control over the inferred input data, which could easily drift away from the offline training regime. Collectively, these represent a well-known problem in machine learning where inputs can be out-of-distribution (OOD) relative to training data \cite{hendrycks2016baseline,yang2024generalized,yuan2022towards,arzani2025interpreting}, where zero-shot inference with a pretrained model often yields low accuracy.

A trivial remedy to the OOD issue is to fine-tune on the new distribution with transfer learning~\cite{church2021emerging}, but this has limited success as a full transfer learning based fine-tuning with new data typically causes catastrophic forgetting on the prior learned data distribution~\cite{french1999catastrophic}, while limited fine-tuning will still under-perform on the new OOD task. This challenge motivates the development of methods that can detect distribution shifts~\cite{yang2024generalized}, identify OOD inputs, and monitor or preserve model reliability under evolving desired input conditions~\cite{gama2014survey,hendrycks2016baseline,liang2017enhancing,ovadia2019can,koh2021wilds}. Such challenges are particularly significant in SciML applications where small distributional changes can reflect meaningful physical differences due to nonlinearity, and computational costs or data access limitations often prohibit retraining from scratch. This naturally connects to the broader need for continual learning (CL)~\cite{hadsell2020embracing,de2021continual}, robust adaptation, and efficient mechanisms to update surrogate models as new input regimes are encountered.

To address catastrophic forgetting, CL methods are commonly grouped into five families: regularization-based, replay-based, optimization-based, architecture-based, and representation-based \cite{wang2024comprehensive,hadsell2020embracing,vandeven2019three}. Regularization-based methods constrain updates to protect past knowledge, either by penalizing changes to parameters important for earlier tasks such as in Elastic Weight Consolidation (EWC) or by matching a frozen model’s outputs while learning the new data such as in Learning without Forgetting (LwF)~\cite{li2017learning,kirkpatrick2017overcoming}.
Replay-based methods keep a small buffer of past samples (or the corresponding \edit{generative model}) and mix them with the new OOD data, but have the limitation of storing and accessing samples from original data, which requires memory and potentially privacy/access issues~\cite{rolnick2019experience,shin2017continual}. Optimization-based methods steer gradients to reduce interference as in Orthogonal Gradient Descent (OGD) or enforce constraints such that losses on replayed exemplars do not increase as in Gradient Episodic Memory (GEM)~\cite{farajtabar2020ogd,lopez2017gradient,saha2021gradient}. Architecture-based approaches protect or allocate parameters in two main ways. In static methods, parameters with small or negligible values can be effectively pruned or deactivated, allowing the model to reuse its capacity for new tasks while preserving previously learned representations. In contrast, dynamic-based methods grow the network as new tasks arrive by adding new neurons, layers, masks, or branches dedicated to the new task, while keeping earlier components fixed. This separation reduces interference between tasks but increases model size over time (e.g., the PiggyBack method)~\cite{mallya2018packnet,mallya2018PiggyBack,rusu2016progressive}. Representation-based methods stabilize features so earlier decision boundaries remain usable while learning new ones, using techniques like feature distillation, cosine-normalized classifiers, and bias correction \cite{rao2019continual,cha2021co2l}. In our study, we compare methods from all major CL categories except representation-based approaches. This exclusion is motivated by the fact that representation-based methods typically rely on pre-trained or self-supervised encoders and aim to learn task-agnostic features, which differs fundamentally from the task or class-incremental CL setting considered here. As noted in prior work, their strong performance often reflects the quality of frozen representations rather than effective mitigation of catastrophic forgetting, making direct comparison potentially misleading \cite{Zhou_2024,van2022three}.

Most work on CL focuses on classification, with fewer studies looking into regression tasks~\cite{he2021clear,besnard2024forecasting,ding2024understanding}. CL work in SciML and operator learning is even more overlooked. As an example, a multifidelity CL framework for physical systems was developed in~\cite{howard2024multifidelity}. A new operator learning framework was proposed in~\cite{tripura2023ncwno} that uses a gating mechanism to mix local wavelet experts and produce task-specific solutions. Finally, a benchmark study on regression data demonstrates challenges in CL for engineering problems~\cite{samuel2025cl3d}. In the context of non-scientific operator learning, Kang et al.~\cite{kang2024continual} proposed a CL framework that combines neural operators with the lottery ticket hypothesis. Their approach identifies task-specific winning subnetworks through iterative pruning, effectively selecting the most suitable computational pathways for each task. The method was demonstrated on video representation learning.
The multifidelity approach introduced by \cite{howard2024multifidelity} helps reduce forgetting in physics-informed neural networks (PINNs) by using the output of the previous model as an input to the new model for each task. Their results show reduced forgetting, but their comparison is limited to only two CL methods: experience replay and Memory Aware Synapses. The approach proposed by \cite{tripura2023ncwno} uses a more sophisticated mechanism based on gating and probabilistic routing, and it requires access to task labels during training and inference. In the comparative study by \cite{samuel2025cl3d}, only one method from each CL category is selected, and the task is limited to mapping geometric features to a scalar quantity (the drag coefficient). Our work builds on these recent studies by introducing a new method that completely mitigates forgetting, is simple to implement, and is task-agnostic, requiring no task labels. In addition, we provide a more detailed comparison by evaluating two methods from each CL category and provide a practical and more complex function-to-function mapping scenario, focusing on both qualitative and quantitative results, offering deeper insight into the behavior of different methods.

We consider the setting of functional mapping with Fourier Neural Operator (FNO) and target a practical cardiovascular flow problem where the goal is to utilize time-resolved concentration fields away from the vessel wall (easier to measure experimentally) and quantify time-averaged wall shear stress (TAWSS), an important metric in cardiovascular disease~\cite{mahmoudi2021story}, which is not easy to measure experimentally. We consider a longitudinal scenario where multiple times OOD input data are provided and CL needs to be repeated. We develop a dataset to benchmark established CL baselines (EWC, LwF, replay, GEM, OGD, PiggyBack, and LoRA) documenting their shortcomings and introduce a lightweight Single-Layer Extension (SLE) adapted to FNO (SLE-FNO) with OOD detection for efficient task-agnostic CL.

Our contributions are summarized as follows:

\begin{itemize}

\item The underexplored setting of CL for SciML function-based regression problems is directly addressed, where the objective is to predict continuous physical fields. This setting introduces unique challenges not captured by standard classification-based CL benchmarks.

\item We introduce SLE-FNO, a parameter-efficient CL architecture tailored to FNO and readily extensible to other neural architectures. SLE-FNO enables effective sequential adaptation while preserving the operator-learning and spectral structure of FNO, avoiding full network duplication or uncontrolled parameter growth. In our examples, each new task added between 1.5-4.4\% of the total parameters.

\item A task-agnostic CL framework with an integrated OOD detector is developed, enabling automatic task identification during inference without prior task labels. In the reported experiments, this mechanism achieves perfect task-identification accuracy \edit{in most stages}, supporting reliable model selection under task shifts.

\item A detailed comparative evaluation of standard CL algorithms applied to FNO on scientific data is presented, revealing differences in forgetting behavior, data efficiency, and predictive performance on new tasks. Multiple stages of CL are also considered to provide a more detailed analysis of longitudinal CL. 

%\item Model performance is evaluated using both quantitative error metrics and qualitative field visualizations, allowing assessment not only of numerical accuracy but also of physically meaningful spatial error patterns in the reconstructed fields.

\item Low-sample fine-tuning regimes and their impact on model performance during CL are analyzed, providing insight into how limited data availability in new SciML tasks affects 
adaptation, knowledge retention, and predictive robustness.

\item A practical problem (blood flow in aneurysms) is considered where the goal is to predict TAWSS from time-resolved concentration fields considering changes in geometry and pulsatile blood flow rates.

\end{itemize}

\section{Methods}

\begin{figure}[h!]
    \centering
    %\vspace{-4pt}
    \includegraphics[width=.8\textwidth]{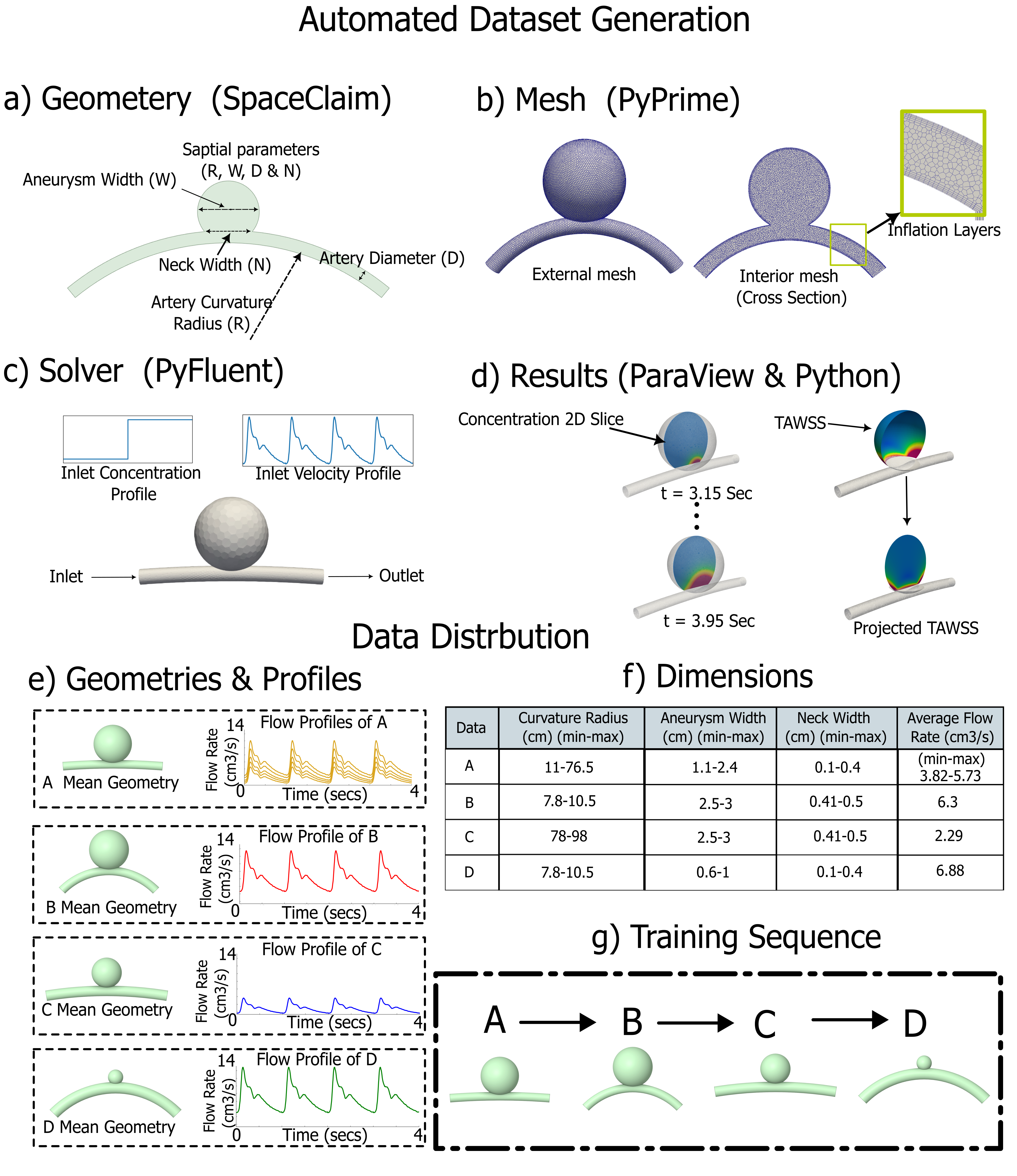}
    \caption{(a) Geometry sketch with four spatial parameters \(W, D, N, R\). We fix \(D\) and vary three non-dimensional ratios \(W/D\), \(N/D\), and \(R/D\). (b) Polyhedral mesh of the geometry with a zoom on the inflation layers. (c) The transient boundary conditions used. (d) Extracted results: a 2D slice of concentration at five time steps from 3.15\,s to 3.95\,s (within the last cardiac cycle), and the corresponding TAWSS projected onto the same slice. (e) Mean spatial parameters for each dataset with the associated inlet flow profiles. (f) Table of parameter ranges and flow rates for each dataset/task. (g) Training/finetuning sequence used throughout the study based on the four datasets/tasks.}
    \label{fig:data_generation}
\end{figure}

\subsection{Problem Statement}

In this work, we evaluate several benchmark CL algorithms covering major categories of CL techniques~\cite{de2021continual}, and we propose a new CL method termed SLE-FNO, equipped with efficient OOD detection. The dataset consists of time-resolved 2D concentration slices inside a 3D idealized aneurysm geometry, mapped to the corresponding TAWSS projection on a 2D domain. As the main neural architecture, we use FNO due to its strong capability for learning function-to-function mappings in SciML tasks~\cite{li2020fourier}.

Let $C(x,y,t)$ denote the concentration field and $\mathrm{TAWSS}(x,y)$ denote the target field. The surrogate learning problem is formulated as
\begin{equation}
\mathcal{F}: \{C(x,y,t_k)\}_{k=1}^{K} \;\longrightarrow\; \mathrm{TAWSS}(x,y) \;,
\end{equation}
where $K$ is the number of temporal snapshots.

We generate four different datasets (tasks) that differ in the geometry and pulsatile inlet flow rates. Each task is generated using computational fluid dynamics (CFD) simulations for a 3D curved vasculature with an idealized aneurysm (sudden enlargement), where the aneurysm is the region of interest. Using FNO, as a surrogate model that maps the time-resolved concentration slices to TAWSS projection is developed.

In practical SciML applications, one typically starts with a large initial dataset covering a wide range of conditions, followed by new tasks. Ideally, one wants to adapt the model to the new tasks with a limited number of new simulations. Our experimental setup follows this scenario, and we evaluate CL performance under this progressively data-limited regime. Additionally, we consider a longitudinal CL scenario where the goal is to continuously adapt the model to new OOD regimes. Specifically, we consider 3 adaptions after the baseline training and for each adaption quantify the error on the new dataset as well as the prior regimes. An overview of the study design is shown in Fig.~\ref{fig:data_generation}.

%To make the framework more compatible with real-life inference and deployment, we note that labels are typically unavailable during inference. Therefore, the full framework is constructed using an OOD detector that determines, at inference time, which task-specific model routing should be used. If the incoming sample is detected as OOD with respect to all previously learned tasks, the framework triggers a fine-tuning stage before reliable inference can be performed. In this work, Kernel PCA is used as the OOD detector to perform this task-aware routing and decision making during inference.

\subsection{Dataset}
\subsubsection{Overview}
We generated 230 CFD simulations where each simulation considers an idealized aneurysm \edit{shown in} Fig.~\ref{fig:data_generation}. 
We varied three geometric parameters, namely the artery curvature, aneurysm neck width, and aneurysm width~\cite{raghavan2005quantified} (Fig.~\ref{fig:data_generation}a). 
The inlet boundary condition, defined as a pulsatile flow waveform, was also varied to span along physiologic conditions~\cite{womersley1955method,ku1985pulsatile,malek1999hemodynamic}. To ensure periodic convergence, we injected the concentration after two full cardiac cycles and used data from the \edit{fourth} cycle to train the model. We split the dataset into four groups \{A, B, C, D\}:
\begin{itemize}
  \item Group A (baseline data): 200 simulations generated from 40 geometries each with 5 inlet flow profiles.
  \item Groups B, C, D (OOD data for finetuning): each contains 10 geometries with one inlet profile each.
\end{itemize}

All geometries and inlet profiles in groups B, C, and D were extrapolating with respect to dataset A and also each other. We followed the 80\% training and 20\% test split of the datasets, as summarized in Table~\ref{tab:train_test_split}. 

\begin{table}[h!]
    \centering
    \caption{The number of training and test samples per task.}
    \label{tab:train_test_split}
    \begin{tabular}{ccc}
        \hline
        Task            & Train Samples & Test Samples \\
        \hline
        A               & 160           & 40           \\
        B, C, D (each) & 8             & 2            \\
        \hline
    \end{tabular}
\end{table}

\subsubsection{CFD simulations}

Blood was modeled as a Newtonian fluid~\cite{arzani2018accounting} with density $\rho=1.06~\text{g/cm}^3$, dynamic viscosity $\mu=0.04~\tfrac{\text{g}}{\text{cm}\cdot\text{s}}$, and the passive scalar diffusion coefficient was selected as $D_m=2\times10^{-4}~\text{cm}^2/\text{s}$. Given a characteristic velocity $U$ (mean velocity) and length $L$ (arterial diameter), the Reynolds and Péclet numbers are defined as
\begin{equation}
\mathrm{Re}=\frac{\rho U L}{\mu}\;, \qquad 
\mathrm{Pe}=\frac{U L}{D_m} \;.
\end{equation}
Across the flow-rate range considered, we obtain $\mathrm{Re}\in[85,\,255]$ and $\mathrm{Pe}\in[2.4\times 10^{4},\,7.2\times 10^{4}]$. Ansys Fluent was used for the simulations. The computational domain was discretized with polyhedral control volumes, which are known to improve convergence and accuracy for finite-volume CFD while reducing cell count compared with purely tetrahedral meshes. We used a minimum element size of $0.01~\text{cm}$ and a maximum element size of $0.05~\text{cm}$, with 5 prism (inflation) layers near the wall. The transient solver used a time‐step size $\Delta t=0.001~\text{s}$ for $4000$ steps (4 cardiac cycles). Instantaneous wall shear stress (WSS) vector, $\boldsymbol{\tau}_w(t)$, was used to define TAWSS as
\begin{equation}
\mathrm{TAWSS} \;=\; \frac{1}{T}\int_{0}^{T} \big\lVert \boldsymbol{\tau}_w(t)\big\rVert \, dt \;,
\label{eq:tawss}
\end{equation}
where $\mathrm{TAWSS}$ denotes the time-averaged magnitude of the WSS acting on the vessel wall over one cardiac cycle, $T$ is the cycle duration, $\boldsymbol{\tau}_w(t)$ is the instantaneous WSS vector evaluated on the vessel wall at time $t$, and $\lVert \cdot \rVert$ denotes its Euclidean norm. TAWSS was calculated based on the \edit{fourth} cardiac cycle $T$. TAWSS is a standard hemodynamic parameter widely used to characterize the role of blood flow and hemodynamics in cardiovascular disease~\cite{malek1999hemodynamic,staarmann2019shear,mahmoudi2021story}. 

The 3D unsteady incompressible Navier–Stokes equations were solved and the corresponding velocity field was coupled to the unsteady advection–diffusion equation to obtain time-resolved passive scalar concentrations. No–slip condition was imposed at vessel walls for velocity, with prescribed inflow and zero–gauge pressure at the outlet and for the scalar, a specified constant concentration at the inlet and zero–flux wall condition were used.

% for the velocity–pressure field $(\mathbf{u},p)$,
% \begin{equation}
% \nabla\!\cdot\!\mathbf{u} = 0,\qquad
% \rho\left(\frac{\partial \mathbf{u}}{\partial t} + (\mathbf{u}\!\cdot\!\nabla)\mathbf{u}\right)
% = -\,\nabla p + \mu\,\nabla^{2}\mathbf{u},
% \end{equation}
% assuming a Newtonian fluid with constant density $\rho$ and viscosity $\mu$. Using this unsteady flow field, we then advanced an unsteady advection–diffusion equation for the passive scalar $c(\mathbf{x},t)$,
% \begin{equation}
% \frac{\partial c}{\partial t} + \mathbf{u}\!\cdot\!\nabla c \;=\; D\,\nabla^{2}c,
% \end{equation}

\subsubsection{Automated Pipeline}

We automated geometry generation in Ansys SpaceClaim using a Python script and exported each model as a STEP file (ISO~10303). After the script was configured, all geometries were generated within a few minutes. Meshing was performed with PyPrimeMesh, an Ansys Python-based automated meshing tool. The simulations were solved in Ansys Fluent through its Python interface (PyFluent), which is based on the finite volume method (Ansys, Inc., Canonsburg, PA, USA). The results were post-processed using ParaView and the Python VTK library~\cite{ayachit2015paraview}. A planar slice was extracted at the aneurysm midsection, and from the transient concentration field we saved five time instants over one cardiac cycle. TAWSS was projected onto the same 2D plane to serve as the target output. Thus, the input consists of five 2D concentration slices within the aneurysm and the spatial coordinates $(x,y)$, and the output is a single 2D TAWSS projection.

\begin{figure}[h!]
    \centering
    \includegraphics[width=1\textwidth]{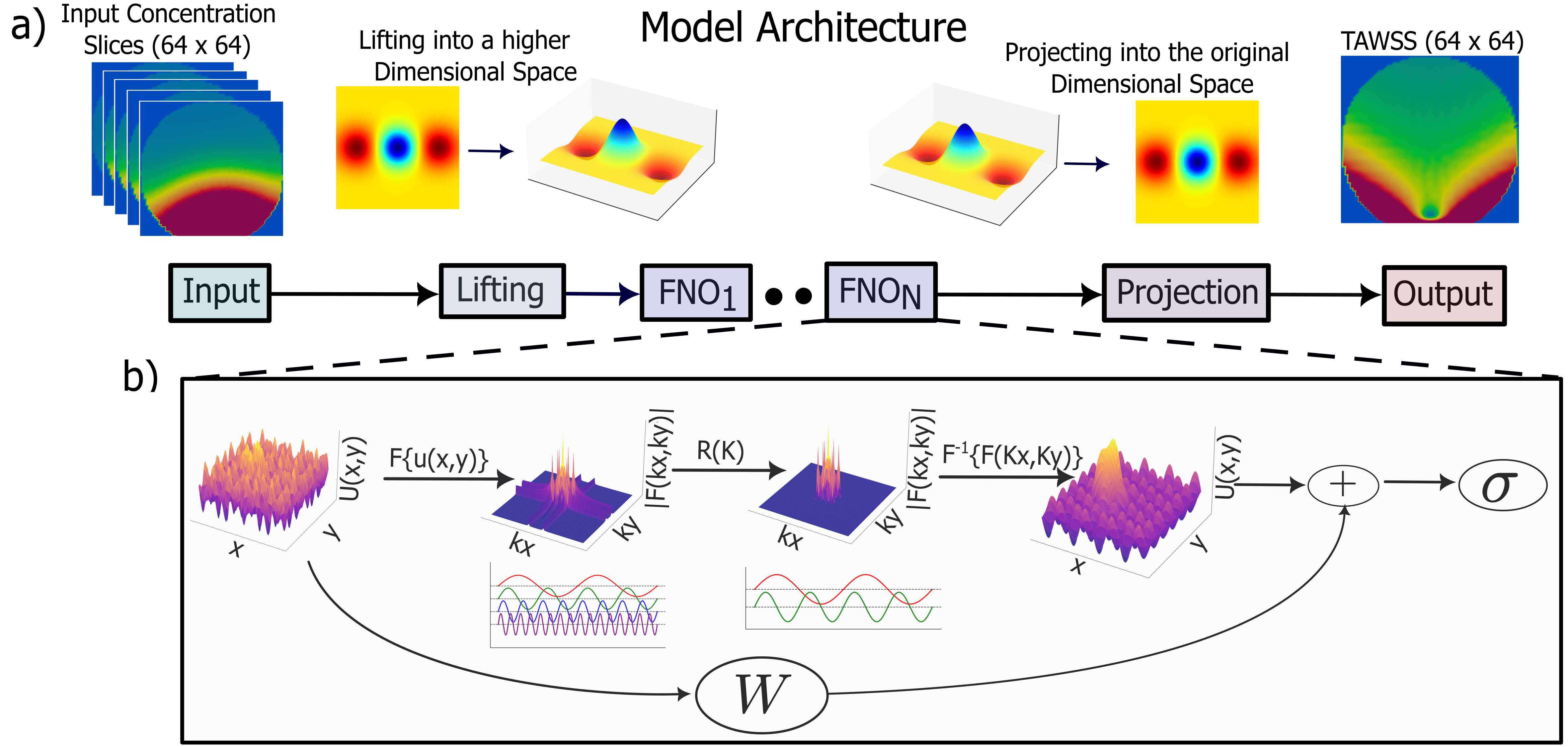}
    \caption{The overall Fourier neural operator (FNO) architecture. (a) FNO lifts the input to a higher-dimensional channel space through a neural lifting, applies a stack of Fourier operator layers with nonlinear activations, then projects back to the target dimension with a neural projection to obtain the output. (b) Fourier layer overview: Given input in the feature space, the upper branch applies \edit{Fast Fourier Transform (FFT)} \(\mathcal{F}\), a learned linear map \(R\) on the retained low-frequency modes while discarding higher modes, then applying an inverse transform \(\mathcal{F}^{-1}\). The lower branch applies a local linear transform \(W\) and acts as a residual connection.}
    \label{fig:main_model}
\end{figure}

\subsection{Neural Operator Baseline Training}

% In general, for order $s\in\mathbb{N}$ the Sobolev norm is
% \begin{equation}
% \|f\|_{H^s(\Omega)}^2=\sum_{|\alpha|\le s}\left\|D^{\alpha} f\right\|_{L^2(\Omega)}^2,
% \end{equation}
% and our training loss applies this to the prediction error. A practical first-order ($s=1$) instantiation is

We chose FNO as the primary neural architecture~\cite{li2020fourier}. The configuration used 4 Fourier layers with 64 hidden channels, lifting and projection ratios of 2, and 16 retained Fourier modes. Training was performed for 3000 epochs \edit{and batch size of 8 was used}. The learning rate started at $10^{-3}$ and was reduced to $10^{-5}$. The Sobolev-norm loss was used, which penalizes discrepancies in both the field values and their gradients~\cite{czarnecki2017sobolev}. Considering just first order derivatives, the Sobolev loss used is defined as
\begin{equation}
\mathcal{L}_{\mathrm{Sob}}=\frac{1}{N}\sum_{i=1}^{N}
\left(
\| \hat{y}_i - y_i \|_2^2
+ \lambda\, \| \nabla \hat{y}_i - \nabla y_i \|_2^2
\right) \;,
\end{equation}
where $\hat{y}_i$ and $y_i$ are the predicted and true fields for sample $i$, $\nabla(\cdot)$ denotes spatial gradients, and $\lambda>0$ balances value and gradient matching.

% To compare predictions against ground truth, we first compute the relative $L^2$ error on the discretized 2D field (vectorized over all pixels):
% \begin{equation}
% \mathrm{Rel}\text{-}L^2
% = \frac{\left\|\hat{\mathbf{y}}-\mathbf{y}\right\|_{2}}
% {\left\|\mathbf{y}\right\|_{2}} \;.
% \end{equation}
% The error is also converted into an accuracy ($R$) metric defined as
% \begin{equation}
% \mathrm{R} = 1 - \mathrm{Rel}\text{-}L^2 \;,
% \end{equation}
% where higher values correspond to better agreement between prediction and ground truth.

All fields are discretized at $64\times64$ resolution. The model input has 7 channels: 5 time slices of concentration within a cardiac cycle window and two coordinate channels $(x,y)$. The target has a single channel corresponding to the 2D TAWSS projection. An overview of the FNO architecture is shown in Fig.~\ref{fig:main_model}.

\begin{figure}[h!]
    \centering
    \includegraphics[width=0.9\textwidth]{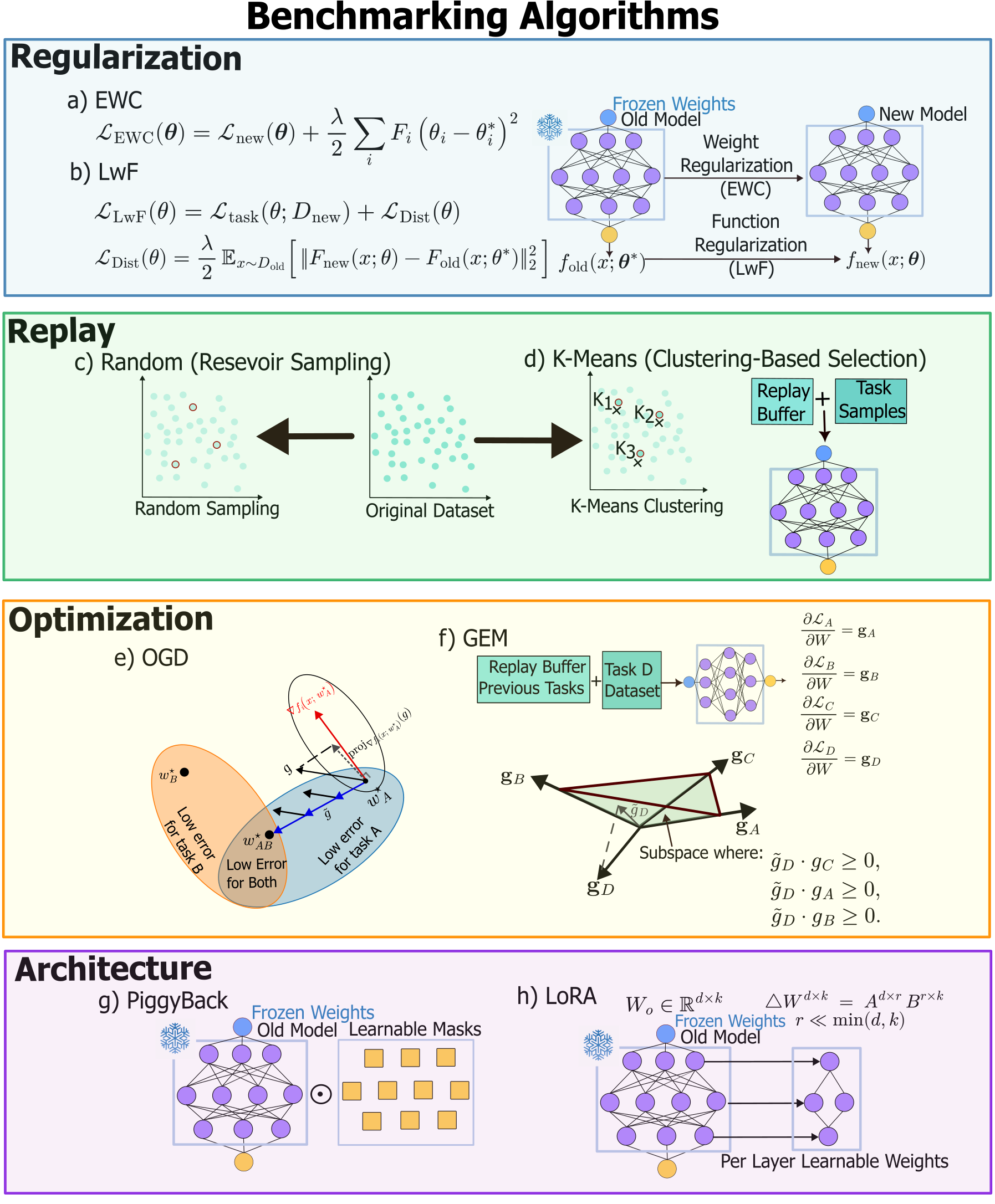}
    \caption{Regularization-based: (a) Elastic Weight Consolidation (EWC) constrains updates using a Fisher-weighted penalty. (b) Learning without Forgetting (LwF) applies distillation from a frozen previous model.
    Replay-based: (c) Random reservoir sampling. (d) K-means exemplar replay. Optimization-based: (e) Orthogonal Gradient Descent (OGD) projects gradients to avoid interference with past tasks (recreated from~\cite{farajtabar2020ogd}). (f) Gradient Episodic Memory (GEM) constrains updates to prevent increases in past-task loss. Architecture-based: (g) PiggyBack learns task-specific masks over a frozen backbone. (h) Low-Rank Adaptation (LoRA) trains per-layer low-rank adapters on a frozen backbone.}
    \label{fig:continual_learning_algorithms}
\end{figure}

\subsection{Continual Learning (CL) Algorithms}

First, we evaluate several established CL algorithms, alongside naive transfer learning based sequential fine-tuning, which serves as a lower-bound baseline for comparison. Most methods follow a sequential training paradigm in which, at each stage, the model is updated using only the training data corresponding to the new task, without access to previous task data (unless stated otherwise in some methods like replay-based methods). As illustrated in Fig.~\ref{fig:data_generation}(g), the training schedule is consistent across methods and proceeds in the order: Task A $\rightarrow$ Task B $\rightarrow$ Task C $\rightarrow$ Task D. A brief summary of each CL method is provided below. An overview of all considered methods is provided in Fig.~\ref{fig:continual_learning_algorithms}.
For each of the following methods, the hyperparameters used during training are listed in the Appendix~\ref{subsec:HT}.

\edit{In the following, we provide a summary of each of the used methods while a more detailed explanation and mathematical formulation can be found in Appendix~\ref{app:cl}.}

\subsubsection{Naive Fine-Tuning (no CL)}

In this baseline, the model is sequentially fine-tuned using transfer learning, where the parameters learned from the previous task are reused as initialization for the next task, without imposing any constraint or CL-specific mechanism. This naive fine-tuning strategy serves as a standard lower bound for knowledge retention, as it typically suffers from severe catastrophic forgetting of earlier tasks. At the same time, it often provides strong learning plasticity on new tasks, since all parameters are freely updated. This baseline reflects the behavior of unconstrained transfer learning in a sequential setting, where knowledge is transferred forward but not preserved across tasks~\cite{goodfellow2013empirical,luo2023empirical,pratt1992discriminability}.

\edit{

\subsubsection{Elastic Weight Consolidation (EWC)}

Elastic Weight Consolidation (EWC)~\cite{kirkpatrick2017overcoming} is a regularization-based CL method that mitigates catastrophic forgetting by constraining important parameters from previous tasks. The key idea is that not all parameters are equally important; therefore, updates to parameters that strongly influence past tasks should be restricted.

To quantify parameter importance, EWC uses the Fisher Information Matrix (FIM), which measures the sensitivity of the model likelihood to parameter changes:
\begin{equation}
F_i \approx \mathbb{E}\left[\left(\frac{\partial}{\partial \theta_i} \log p(y \mid x, \theta)\right)^2\right] \;,
\end{equation}
where $F_i$ represents the (diagonal) importance of parameter $\theta_i$. In practice, a diagonal approximation is used for scalability.

When learning a new task, EWC adds a quadratic penalty that discourages deviation from previously learned parameters:
\begin{equation}
\mathcal{L}_{\mathrm{EWC}}(\theta)
=
\mathcal{L}_{\text{new}}(\theta)
+
\frac{\lambda}{2}\sum_i F_i \left(\theta_i - \theta_i^*\right)^2 \;,
\end{equation}
where $\theta^*$ denotes the parameters after training on previous tasks, and $\lambda$ controls the stability–plasticity trade-off. This formulation preserves previously acquired knowledge by slowing updates to important parameters, while allowing less important ones to adapt to new tasks.
}

\edit{
\subsubsection{Learning without Forgetting (LwF)}

Learning without Forgetting (LwF)~\cite{li2017learning} is a distillation-based CL method that preserves previously learned knowledge by encouraging the current model to match the predictions of an earlier model. Instead of constraining parameters, LwF constrains the model outputs, allowing greater flexibility in adapting to new tasks.

In this framework, the model trained on previous tasks is treated as a fixed teacher, while the current model acts as a student. During training on a new task, the student is optimized to both fit the new data and remain consistent with the teacher’s predictions on the same inputs. The resulting objective combines the task loss with a distillation loss:
\begin{equation}
\mathcal{L}_{\mathrm{LwF}}
=
\mathcal{L}_{\text{task}}(\theta;\mathcal{D}_k)
+
\lambda \,\mathbb{E}_{x \sim \mathcal{D}_k}
\left[\| f_{\theta}(x) - f_{\theta^*}(x) \|_2^2 \right] \;,
\end{equation}
where $\theta^*$ denotes the frozen parameters from previous tasks (teacher), $\theta$ are the current parameters (student), and $\lambda$ controls the trade-off between learning new information and preserving past behavior. By aligning outputs rather than parameters, LwF avoids storing past data while still reducing catastrophic forgetting.

}

\edit{
\subsubsection{Replay-Based}
\label{subsub:replay}

Replay-based methods mitigate catastrophic forgetting by storing a subset of samples from previous tasks in a buffer $\mathcal{B}$ and reusing them during training. In this work, we adopt a fraction-based replay strategy, where approximately $10\%$ of the data is stored (16 samples for Task A), following common practice in CL \cite{buzzega2020dark,prabhu2020gdumb}.

During training, replay samples are combined with current-task data to reinforce past knowledge while learning new information. For Tasks B and C, a strict $10\%$ allocation would result in fewer than one sample; therefore, we store one representative sample per task as a practical discretization of the fraction-based policy. Samples can be selected using strategies such as reservoir sampling \cite{chaudhry2018efficient} or representative selection. While effective, replay methods introduce challenges including memory constraints, potential privacy concerns, and reduced representation as the number of tasks grows.
}

\edit{

In this study, we adopted two replay selection strategies: reservoir (random) sampling and K-means clustering–based selection. Each data sample occupies approximately 132.03 KB when stored in float32 precision. 
The replay buffer composition remains compact across all CL stages. 
During training on task $B$, the buffer stores 16 samples from task $A$, corresponding to a total storage of 2.06 MB. 
For task $C$, the buffer includes 16 samples from $A$ and 1 sample from $B$, resulting in 2.19 MB. 
Finally, during training on task $D$, the buffer contains 16 samples from $A$, 1 from $B$, and 1 from $C$, reaching a peak memory usage of only 2.32 MB (18 samples in total). 
%These results confirm the negligible memory overhead introduced by the used replay strategy.

}

\edit{
\subsubsection{Orthogonal Gradient Descent (OGD)}

Orthogonal Gradient Descent (OGD)~\cite{farajtabar2020ogd} is an optimization-based CL method that prevents interference with previously learned tasks by constraining the direction of parameter updates. Instead of storing past data, OGD preserves knowledge by ensuring that updates for new tasks do not conflict with gradients from earlier tasks.

The key idea is to modify the current gradient so that it is orthogonal to the subspace spanned by past task gradients. This is achieved by projecting the gradient onto the orthogonal complement:
\begin{equation}
\mathbf{g}_k^{\perp}
=
\mathbf{g}_k -
\sum_{t=1}^{k-1}
\frac{\langle \mathbf{g}_k, \mathbf{g}_t \rangle}{\|\mathbf{g}_t\|_2^2} \, \mathbf{g}_t \;,
\end{equation}
where $\mathbf{g}_k$ is the current-task gradient and $\mathbf{g}_t$ are gradients associated with previous tasks.

The model is then updated using the projected gradient, ensuring that new learning does not overwrite directions important to past tasks. By enforcing orthogonality in gradient space, OGD reduces catastrophic forgetting without requiring access to previous data. A limitation of OGD is the need to store and project onto past gradients, which can become computationally expensive as the number of tasks or model parameters grows.

}

\edit{

\subsubsection{Gradient Episodic Memory (GEM)}

Gradient Episodic Memory (GEM)~\cite{lopez2017gradient} is a hybrid CL method that combines replay with gradient-based constraints to prevent forgetting. A small memory buffer of past samples is maintained, and during training on a new task, gradients are computed for both current data and stored memory.

The key idea is to ensure that updates on the new task do not increase the loss of previous tasks. This is enforced by constraining the update direction to have non-negative alignment with past gradients. When this condition is violated, GEM modifies the gradient by solving the following projection problem:
\begin{equation}
\tilde{g} = \arg\min_{g} \frac{1}{2} \|g - g_k\|^2 
\quad \text{s.t.} \quad 
\langle g, g_t \rangle \ge 0 \;\; \forall t<k \;,
\end{equation}
where $g_k$ is the gradient of the current task and $g_t$ are gradients associated with previous tasks. In practice, efficient variants such as A-GEM approximate this constraint using a single reference gradient computed from the memory buffer, significantly reducing computational cost. By directly constraining gradient updates, GEM provides stronger protection against forgetting than standard replay, while remaining more flexible than strictly orthogonal methods such as OGD.

}

\edit{
\subsubsection{PiggyBack}

PiggyBack~\cite{mallya2018PiggyBack} is an architecture-based CL method that avoids forgetting by freezing a shared backbone and learning task-specific masks. Instead of modifying the backbone parameters, the model learns which connections should be active for each task.

Given fixed backbone parameters $\theta$, a task-specific mask $m_k$ is learned and applied through elementwise modulation. The mask is represented as a point-wise scalar with the same dimensionality as $\theta$:
\begin{equation}
\theta_k^{\mathrm{eff}} = m_k \odot \theta \;,
\end{equation}
where $\theta_k^{\mathrm{eff}}$ are the effective parameters for task $k$. This approach leverages the representational capacity of a pretrained network while isolating task-specific adaptations, thereby preventing interference between tasks. Since the backbone remains unchanged, previously learned knowledge is preserved by design. In our implementation, we retain continuous mask values rather than binarizing them, which improves flexibility and avoids loss of fine-grained connectivity patterns.

}

\edit{
\subsubsection{Low-Rank Adaptation (LoRA)}

Low-Rank Adaptation (LoRA)~\cite{hu2022lora} is a parameter-efficient method that adapts pretrained models by learning low-rank updates while keeping the backbone weights fixed. Instead of updating the full weight matrix, LoRA assumes that task-specific changes lie in a low-dimensional subspace.

Given a pretrained weight matrix $W$, LoRA models the adapted weights as:
\begin{equation}
W' = W + \frac{\alpha}{r} A B \;,
\end{equation}
where $A \in \mathbb{R}^{k \times r}$ and $B \in \mathbb{R}^{r \times d}$ are low-rank trainable matrices with $r \ll \min(k,d)$, and $\alpha$ controls the adaptation strength. In CL settings, each task is assigned its own low-rank parameters $(A_k, B_k)$ while sharing the frozen backbone:
\begin{equation}
W_k' = W + \frac{\alpha}{r} A_k B_k \;.
\end{equation}

This modular design isolates task-specific updates, preventing interference between tasks without requiring replay or explicit constraints. As a result, LoRA provides an efficient balance between stability and adaptability, particularly for large pretrained models.

}

\subsection{Single-Layer Extension with Fourier Neural Operator (SLE-FNO)}

We propose a new CL strategy coupled with FNO called Single-Layer Extension (SLE-FNO).  The method belongs to the same category as architecture-based parameter-efficient approaches such as PiggyBack \cite{mallya2018PiggyBack}, LoRA \cite{hu2022lora}, and Side-Tuning \cite{zhang2020side}, where the backbone network is kept frozen and only task-specific components are trained. However, unlike previous generic adapter strategies, SLE-FNO is designed specifically around the structure of FNO and leverages their spectral properties.

FNO can be summarized as:
% \begin{equation}
% x \;\xrightarrow{\phi_{\mathrm{lift}}}\; z_0 
% \;\xrightarrow{\text{FNO blocks}}\; z_L 
% \;\xrightarrow{\phi_{\mathrm{proj}}}\; y,
% \end{equation}

\begin{align}
z_0 &= \phi_{\mathrm{lift}}(x) \;, \\
z_{\ell+1} &= \mathcal{F}^{(\ell)}(z_\ell), \qquad \ell = 0, \dots, L-1 \;, \\
y &= \phi_{\mathrm{proj}}(z_L) \;,
\end{align}
where $x$ denotes the input field to the model, the lifting operator $\phi_{\mathrm{lift}}(\cdot)$ maps the input $x$ into a higher-dimensional latent representation $z_0$, the variables $z_\ell$ represent the latent feature maps at layer $\ell$ ($\ell = 0,\dots,L$), and $\mathcal{F}^{(\ell)}$ denotes the $\ell$-th FNO block that transforms $z_\ell$ to $z_{\ell+1}$. The total number of FNO blocks is $L$. Finally, the projection operator $\phi_{\mathrm{proj}}(\cdot)$ maps the final latent representation $z_L$ back to the output space, producing the model prediction $y$. Due to strong spectral bias, pretrained FNOs tend to learn low-frequency structure \cite{rahaman2019spectral}, meaning only small high-frequency corrections are typically required for new tasks.

The core idea of SLE-FNO is to keep the pretrained FNO backbone completely frozen and adapt to each new task by introducing a single, task-specific FNO layer. Rather than modifying existing model weights, this additional layer learns a residual correction that adjusts the frozen backbone output for the current task. This design enables efficient adaptation with minimal parameter overhead while preventing interference with previously learned tasks. SLE-FNO updates the FNO as

\begin{align}
z_0 &= \phi_{\mathrm{lift}}^{\text{(frozen)}}(x) \;, \\[4pt]
\tilde{z}_L &= f_{\mathrm{base}}^{\text{(frozen)}}(z_0) \;, \\[6pt]
Z_{\mathrm{SLE\text{-}FNO}}^{(k)} &= g_k(z_0) \;, \\[4pt]
z_L^{(k)} &= \tilde{z}_L + Z_{\mathrm{SLE\text{-}FNO}}^{(k)} \;, \\[6pt]
f_k(x) &= \phi_{\mathrm{proj}}^{\text{(frozen)}}\!\left(z_L^{(k)}\right) \;,
\end{align}
where the function $f_{\mathrm{base}}^{\text{(frozen)}}(\cdot)$ represents the frozen pretrained FNO backbone, composed of multiple FNO blocks whose parameters remain fixed during CL, and produces the frozen backbone latent output $\tilde{z}_L$. For each task $k$, SLE-FNO introduces a task-specific FNO layer $g_k(\cdot)$ that operates on the lifted representation $z_0$ and produces a residual feature map $Z_{\mathrm{SLE\text{-}FNO}}^{(k)}$. This residual is added to the frozen backbone output to form the task-adapted latent representation $z_L^{(k)}$. Finally, the frozen projection operator $\phi_{\mathrm{proj}}^{\text{(frozen)}}(\cdot)$ maps $z_L^{(k)}$ back to the output space, yielding the task-specific prediction $f_k(x)$.

During training on task $k$, only the parameters of the task-specific layer $g_k$ are updated, while the frozen backbone $f_{\mathrm{base}}^{\text{(frozen)}}$, the frozen lifting operator $\phi_{\mathrm{lift}}^{\text{(frozen)}}$, and the frozen projection operator  $\phi_{\mathrm{proj}}^{\text{(frozen)}}$ remain unchanged. This strategy enables effective CL by allowing task adaptation without modifying the main model weights.

During training on task $k$, the parameters of the frozen backbone remain fixed and only the task-specific SLE-FNO parameters are updated:
\begin{equation}
\nabla_{\theta_{\mathrm{base}}}\mathcal{L}_k = \mathbf{0} \;, \qquad
\nabla_{\theta_{\mathrm{SLE\text{-}FNO}}}\mathcal{L}_k \neq \mathbf{0} \;.
\end{equation}

Compared to PiggyBack, which learns binary masks over all backbone weights~\cite{mallya2018PiggyBack}, LoRA, which attaches low-rank matrices across multiple layers~\cite{hu2022lora}, and Side-Tuning, which adds full side networks~\cite{zhang2020side}, SLE-FNO concentrates all task-specific learning capacity into a single FNO operator acting on the lifted feature space. As a result, SLE-FNO remains extremely lightweight in both memory and computation, increasing the total parameter count by only $1.5\%$ for tasks B and C, and $4.4\%$ for the more challenging task D in our problem.

\begin{figure}[h!]
    \centering
    \includegraphics[width=.8\textwidth]{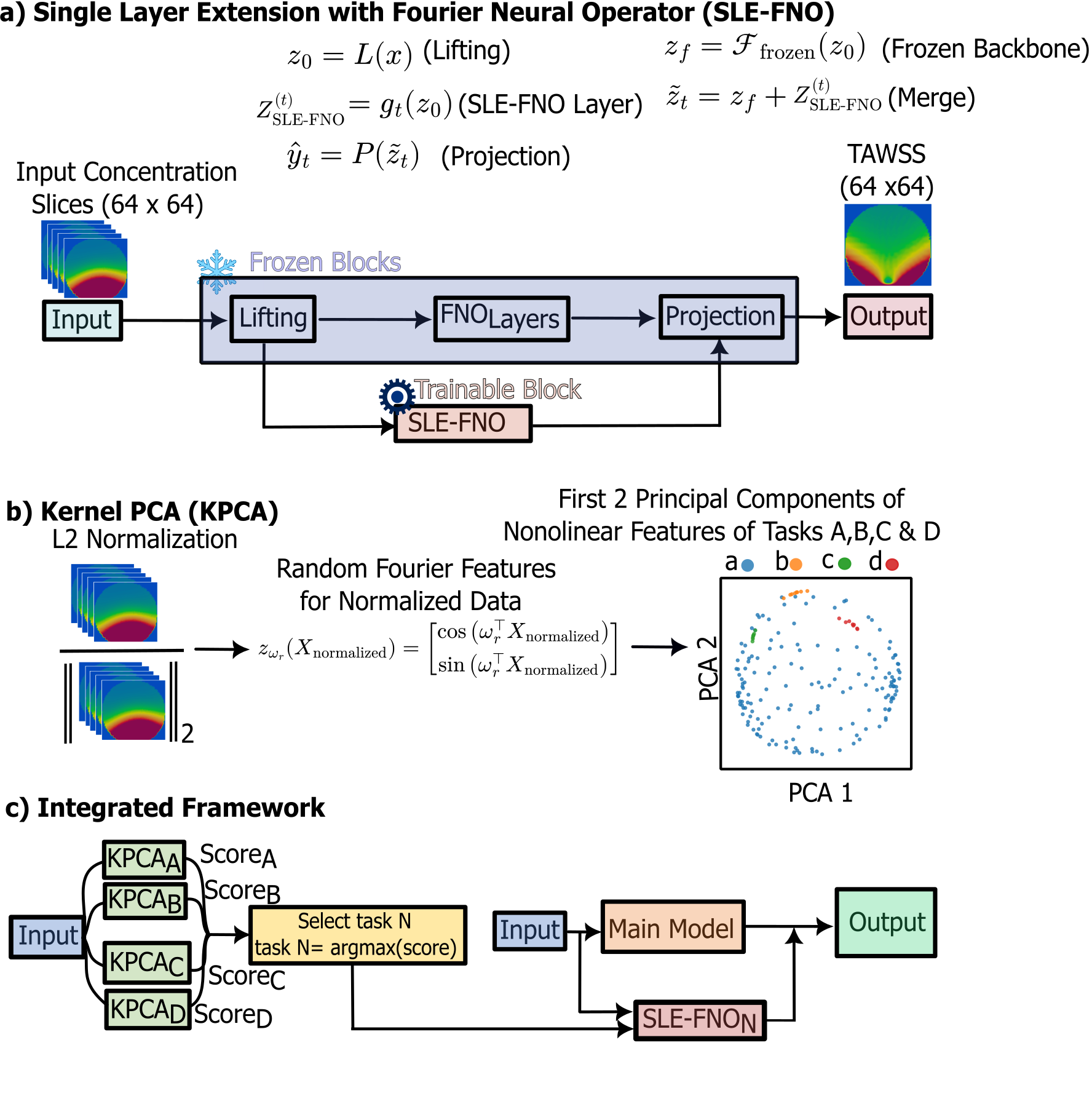}
    \caption{Single-Layer Extension with Fourier Neural Operator (SLE-FNO). (a) During training, only a single Fourier layer is trainable while the backbone remains frozen. (b) Kernel PCA is used for \edit{routing and} out-of-distribution (OOD) detection. (c) Complete framework at inference: \edit{KPCA modules} select the route, and the prediction is based on the frozen backbone and the SLE-FNO layer.}
    \label{fig:SLE-FNO}
\end{figure}

SLE-FNO also differs fundamentally from constraint-based CL methods such as EWC, OGD and GEM, which rely on penalty terms, replay buffers, or gradient projection mechanisms to protect previous tasks. SLE-FNO avoids these mechanisms entirely, preventing optimization stiffness, reducing overhead, and removing the need for stored past data or constrained gradient updates. An overview is shown in Fig.~\ref{fig:SLE-FNO}a.

After training on $N$ tasks, the model consists of a shared frozen backbone and $N$ task-specific SLE-FNO layers. Given an input $x$, the inference process proceeds as follows.  
First, the input is passed through the frozen lifting operator, which maps it to the latent space $z_0$. Next, the lifted representation is processed in parallel by the frozen backbone and the selected task-specific SLE-FNO. A routing function selects the task index,
\begin{equation}
r(x) \in \{1,\dots,N\}.
\end{equation}

The frozen backbone produces its latent output
\begin{equation}
z_{\mathrm{base}} = f_{\mathrm{base}}^{\text{(frozen)}}(z_0),
\end{equation}
and in parallel, the selected SLE-FNO produces a task-specific latent correction
\begin{equation}
Z_{\mathrm{SLE\text{-}FNO}} = g_{r(x)}^{\text{(frozen)}}(z_0).
\end{equation}

Finally, the outputs of the frozen backbone and the SLE-FNO are combined and passed through the frozen projection operator, which maps the latent representation to the output space:
\begin{equation}
\hat{y} = \phi_{\mathrm{proj}}^{\text{(frozen)}}\!\left(z_{\mathrm{base}} + Z_{\mathrm{SLE\text{-}FNO}}\right).
\end{equation}

An overview \edit{of} the inference is sketched in Fig.~\ref{fig:SLE-FNO}c.

\subsection{Task-agnostic routing}

Architecture-based CL methods that rely on task-specific modules mainly differ in how they handle task identity at inference. Early approaches are explicitly task-aware, assuming that the task label is given and can be used to select the appropriate parameters. Representative examples include Progressive Neural Networks (PNN)~\cite{rusu2016progressive}, which add a new column per task and select it via the known task ID, and mask-based method such as PiggyBack~\cite{mallya2018PiggyBack} and Hard Attention to the Task (HAT)~\cite{serra2018overcoming}, all of which require the task label to activate the correct mask or output head. Hypernetwork-based approaches similarly condition parameter generation on an explicit task code~\cite{ha2016hypernetworks}. More recent work targets task-agnostic settings by introducing mechanisms to infer or approximate task identity from data. One line of work uses auxiliary models for task inference, such as Expert Gate~\cite{aljundi2017expert}. Another strategy formulates task selection as a routing or classification problem, as in task agnostic CL using multiple experts (TAME)~\cite{zhu2024tame}, where a selector network is trained online, or CL based on OOD detection and task masking (CLOM)~\cite{kim2022continual}, which relies on OOD detection and confidence comparison across task-specific heads. A different perspective is taken by optimization-based inference methods such as Supermasks in Superposition (SupSup)~\cite{wortsman2020supermasks}, which recover the task by searching over task-specific masks at inference time.

In our introduced method SLE-FNO, we follow the task-agnostic setting. This aligns with real-world deployment, where the data stream does not indicate task boundaries and the model must operate continuously without external supervision. Detectors can be kernel PCA (KPCA) reconstruction error \cite{scholkopf1998nonlinear,scholkopf1999kpca}, Mahalanobis distance in feature space \cite{lee2018simple}, energy-based scores \cite{liu2020energy}, or Out-of-Distribution detector for Neural network (ODIN)-style maximum-softmax baselines combining temperature-scaled softmax confidence with a small input perturbation~\cite{liang2017enhancing,hendrycks2016baseline}. The previously mentioned methods when added to the CL algorithms, help in accurately determining the task, and ultimately using task-specific parameters/branches.

\subsubsection{Kernel Principal Component Analysis (KPCA)}
Kernel principal component analysis (KPCA) was introduced as an OOD detector~\cite{fang2024kernel}. The motivation was that standard PCA is inherently linear and therefore struggles to detect nonlinear shifts in data distributions. In contrast, KPCA performs PCA in a nonlinear feature space generated by a kernel, making it a more accurate OOD detector.

Given input samples \(x \in \mathbb{R}^d\), PCA constructs the covariance matrix
\begin{equation}
C = \frac{1}{N} \sum_{i=1}^{N} (x_i - \mu)(x_i - \mu)^\top \;,
\end{equation}
where $C$ denotes the sample covariance matrix of the data. The set $\{x_i\}_{i=1}^N$ represents $N$ data samples, where each $x_i \in \mathbb{R}^d$ is a $d$-dimensional feature vector. The quantity $\mu = \frac{1}{N}\sum_{i=1}^N x_i$ is the sample mean of the data. The term $(x_i - \mu)$ denotes the mean-centered data vector, and the outer product $(x_i - \mu)(x_i - \mu)^\top$ captures pairwise correlations between features. Averaging these outer products over all samples yields the empirical covariance matrix $C$, which extracts principal directions via eigendecomposition. 

KPCA replaces the covariance in the input space with the one computed in a feature space \(\Phi(x)\) induced by a kernel function:
\begin{equation}
k(x_i, x_j) = \langle \Phi(x_i), \Phi(x_j) \rangle \;.
\end{equation}
Instead of computing \(\Phi(x)\) explicitly, KPCA relies on a pre-defined kernel matrix $K_{ij} = k(x_i, x_j)$ and solves an eigenvalue problem based on this kernel. Samples that deviate strongly from the learned subspace yield large reconstruction errors, which are used as the OOD score (as explained in the next subsection).

In \cite{fang2024kernel}, the kernel design consists of two preprocessing steps before PCA is applied. First, each input is normalized. Next, the normalized samples are mapped using Random Fourier Features (RFF)~\cite{rahimi2007random} to approximate a Gaussian kernel and PCA is then applied to the transformed features. \edit{An overview of the KPCA approach is shown in Fig.~\ref{fig:SLE-FNO}b.}

\subsection{Reconstruction Error Computation}
We describe how KPCA is fit during training and how the reconstruction error (used as an OOD score) is computed at inference. \edit{The KPCA modes are fitted using the same task-specific training inputs, with an identical number of samples and the same number of temporal slices per sample.}
Given training inputs $\{x_i\}_{i=1}^N$ for task $t$, we apply the same preprocessing as in~\cite{fang2024kernel} with $\ell_{2}$ normalization followed by an RFF mapping
\begin{equation}
x_i' \;=\; \frac{x_i}{\|x_i\|_2} \;,
\qquad
z_i \;=\; \mathrm{RFF}(x_i') \in \mathbb{R}^D \;,
\qquad i=1,\dots,N  \;.
\end{equation}
Let $Z = [z_1,\dots,z_N]$ denote the collection of feature vectors. We compute the feature mean and center each feature vector as
\begin{equation}
\bar{z} \;=\; \frac{1}{N}\sum_{i=1}^{N} z_i \in \mathbb{R}^D \;,
\qquad
\tilde{z}_i \;=\; z_i - \bar{z} \;,
\qquad i=1,\dots,N  \;.
\end{equation}
KPCA (computed in this feature space) returns the top $K$ principal directions $\{u_k\}_{k=1}^{K}$, which define the task-$t$ subspace.

\edit{At inference, the same test input samples and corresponding temporal slices are used to compute the inference score.}
For a new input $x$, we apply the same preprocessing and feature mapping:
\begin{equation}
x' = \frac{x}{\|x\|_2},
\qquad
z_{\mathrm{infer}} = \mathrm{RFF}(x') \in \mathbb{R}^{D}.
\label{eq:kpca_infer_feature}
\end{equation}

We then project $z_{\mathrm{infer}}$ onto the task-$t$ KPCA subspace:
\begin{equation}
\Pi^{\mathrm{KPCA}}_{t}(z_{\mathrm{infer}})
=
\bar{z}
+
\sum_{k=1}^{K}
\left\langle z_{\mathrm{infer}} - \bar{z},\, u_k \right\rangle u_k .
\label{eq:kpca_projection}
\end{equation}

Finally, the reconstruction error (OOD score) is defined as the distance between
$z_{\mathrm{infer}}$ and its projection:
\begin{equation}
s_t(x)
=
\left\|
z_{\mathrm{infer}} - \Pi^{\mathrm{KPCA}}_{t}(z_{\mathrm{infer}})
\right\|_2 .
\label{eq:kpca_score}
\end{equation}
Here, $\bar{z}$ and $\{u_k\}_{k=1}^{K}$ are the task-$t$ KPCA mean and principal directions learned during training. A small $s_t(x)$ indicates that the sample lies close to the task-$t$ feature distribution, while a large $s_t(x)$ indicates OOD sample with respect to task $t$. This score is subsequently used by the SLE-FNO router.

\subsection{SLE-FNO Routing}
% We combine the above KPCA-based distribution detector with task‐specific SLE-FNO adapters. During training for each task $t$, a KPCA model is fit on the task’s inputs to capture its distribution, and a SLE-FNO branch attached to the frozen backbone is fine‐tuned for that task. At inference, a new sample $x$ is scored by all saved KPCA detectors via the reconstruction error defined in Eq.~\eqref{eqn:score}
% and routed to the SLE-FNO associated with $\arg\min_t s_t(x)$. If $\min_t s_t(x)>\tau$ for a calibrated threshold $\tau$, the sample is treated as OOD relative to all known tasks, a new SLE-FNO branch is allocated, and fine‐tuning proceeds on the new task (Fig.~\ref{fig:SLE-FNO}c).

We combine the above KPCA-based distribution detector with task-specific SLE-FNO adapters. 
During training for each task $t$, a KPCA model is fit on the task’s input data to capture its distribution, 
and a SLE-FNO branch attached to the frozen backbone is fine-tuned for that task. 

At inference, a new sample $x$ is evaluated by all saved KPCA detectors, yielding reconstruction errors 
$s_t(x)$ as defined in Eq.~\eqref{eq:kpca_score}, 
\edit{and then the sample is assigned to the task}
\begin{equation}
t = \arg\min_t s_t(x) \;,
\end{equation}
where t is corresponding to the KPCA model that best reconstructs $x$, and is routed to the associated SLE-FNO branch. 
If the minimum reconstruction error satisfies
\begin{equation}
\min_t s_t(x) > \tau \;,
\end{equation}
for a calibrated threshold $\tau$, the sample is considered out-of-distribution with respect to all known tasks. 
In this case, a new SLE-FNO branch is allocated and fine-tuned on the new task (Fig.~\ref{fig:SLE-FNO}c).

\edit{
The threshold $\tau$ is calibrated as a percentile of the training sample reconstruction error distribution for each task. More specifically, after fitting KPCA on the task-specific training data, each training sample is reconstructed using the retained number of modes; further details on the number of retained modes are provided in Appendix~\ref{subsec:KPCAmodes}. The resulting reconstruction errors form a task-specific training error distribution, from which $\tau$ is selected as a percentile-based threshold.

During inference, the reconstruction error is computed in the same manner as described in the previous inference section and compared against the corresponding $\tau$ value. If the reconstruction error exceeds this threshold, the sample is considered not to belong to that task distribution. If the sample is rejected by all fitted KPCA models, it is identified as OOD, indicating the need to fine-tune a new SLE-FNO branch for the newly encountered distribution. In practice, increasing $\tau$ generally improves routing accuracy by allowing more samples to be assigned to an existing task, whereas decreasing $\tau$ tends to improve OOD detection accuracy at the cost of lower routing accuracy, since more in-distribution samples may be rejected as OOD. The trade-off between routing accuracy and OOD detection accuracy is illustrated in Supplementary Material Figure~S.12. 

In the main results, we assume that the task identity is known at inference time in order to ensure a fair comparison with baseline methods that make the same assumption. Under this setting, SLE-FNO is evaluated under identical inference conditions, so performance differences reflect the learning capability of the methods rather than differences in routing assumptions. To further assess the practicality of SLE-FNO beyond this standard evaluation protocol, Appendix~\ref{subsec:endtoend} removes this known-task assumption and considers task-agnostic inference, under the assumption that all tasks have already been observed during training or fine-tuning. In this setting, the KPCA-based routing mechanism is used to automatically assign each sample to its corresponding task-specific layer. The results show that this task-agnostic inference procedure preserves the exact same accuracy across all stages, with the exception of Stage D for dataset D, where a performance drop is observed.

We also consider the more challenging open-set setting, in which a test sample may belong to a completely unseen distribution. In this case, the percentile-based threshold $\tau$ is used to reject samples that do not match any previously learned task distribution. As shown in Supplementary Figure~S12, enforcing 100\% OOD detection leads to only an approximately 7\% reduction in routing accuracy. This result indicates that the proposed KPCA-based routing mechanism remains effective even under a strict OOD detection requirement, while introducing only a modest trade-off in task assignment performance.
}

\edit{

\subsection{Training Protocol}
\label{subsec:training_protocol}

All methods are evaluated under a multi-run training protocol using five distinct data splits and two random seeds, resulting in ten runs per method (all splits and seeds are fixed across the methods). Each split corresponds to a different 80\%/20\% train/test partition across all tasks, ensuring variability in data allocation while maintaining a consistent ratio. Results are reported as mean $\pm$ standard deviation across runs. The random seeds control the initialization of learnable parameters in trainable components, the stochastic aspects of optimization (e.g., mini-batch ordering), and, where applicable, the initialization of representative centers in clustering-based methods. This protocol follows best practices for reliable model evaluation, where multiple random seeds and data splits are used to capture variability due to both stochastic optimization and dataset partitioning~\cite{takamoto2022pdebench,semmelrock2025reproducibility}.
}

\subsection{Error Metrics}

Following standard practice in neural operator--based surrogate modeling, we evaluate predictions using the post hoc relative $\ell_{2}$ error computed on the discretized \edit{projected} two-dimensional output field, vectorized over all spatial grid points~\cite{li2020fourier} to define a relative $\ell_{2}$ error

\begin{equation}
\mathrm{Rel}\text{-}\ell_{2}
= \frac{\left\|\hat{\mathbf{y}}-\mathbf{y}\right\|_{2}}
{\left\|\mathbf{y}\right\|_{2}} \;,
\end{equation}
where $\hat{\mathbf{y}} \in \mathbb{R}^M$ is the model prediction,
$\mathbf{y} \in \mathbb{R}^M$ is the ground-truth target, and $M$ is the output dimension. \edit{The relative $\ell_{2}$ error is reported in Supplementary Material Table S1 and an accuracy-based metric is used in the results.} We convert the relative error into an accuracy-style metric using a monotonic exponential transformation~\cite{chaudhry2018riemannian}
\begin{equation}
R = \exp\!\left(-\alpha \, \frac{\mathrm{Rel\text{-}\ell_{2}}}{L_{\max}}\right) \;,
\label{eq:R_metric}
\end{equation}
where the parameter $\alpha$ controls the sensitivity of the metric to small error variations and is fixed to $\alpha=3$ for all methods and training stages, and \(L_{\max}\) denotes the maximum expected value of the relative error used for normalization. 
In our experiments, we set \(L_{\max} = 5\). The use of an exponential mapping is further motivated by general machine learning principles, where monotonic nonlinear transformations are commonly employed to reshape sensitivity to skewed error distributions while preserving the relative ordering of model performance~\cite{bengio2017deep}.

Let \(T\) denote the total number of tasks in the CL sequence, and let \(R_{i,j}\) denote the accuracy on task \(j\) after completing training on task \(i\), where \(i \in \{1,\dots,T\}\) and \(j \in \{1,\dots,i\}\). 
Each entry \(R_{i,j}\) is computed using the accuracy metric \(R\) defined in Eq.~\ref{eq:R_metric}. 
This evaluation matrix therefore captures the model’s performance on all previously learned tasks at each training stage and is used to define the following metrics.

\paragraph{Average Accuracy (AA).}
At training stage $i$, the average accuracy is defined as
\begin{equation}
\mathrm{AA}_i = \frac{1}{i} \sum_{j=1}^{i} R_{i,j} \;.
\label{eq:AA_i}
\end{equation}
This metric, adapted from the CL evaluation protocol in~\cite{lopez2017gradient},
measures the mean performance across all tasks learned up to the current training stage $i$.
When $i = T$, this definition reduces to the standard final average accuracy.

\paragraph{Forgetting Measure (F).}
At training stage $i$, we define the task-wise forgetting for each previously learned task
$j \in \{1,\dots,i-1\}$ as
\begin{equation}
\mathrm{F}_{i}^{(j)} =
\max_{k \in \{j,\dots,i\}} R_{k,j} - R_{i,j} \;,
\label{eq:FM_taskwise}
\end{equation}
where $R_{k,j}$ denotes the accuracy on task $j$ after completing training on task $k \in \{j,\dots,i\}$, which includes all training stages from the time task $j$ up to the current stage $i$,
and $R_{i,j}$ is the accuracy on task $j$ after completing training on the current task $i$. The operator $\max(\cdot)$ therefore captures the best observed accuracy achieved on task $j$
over all training stages up to $i$.

The overall forgetting measure at training stage $i$ is then obtained by averaging the task-wise
forgetting values across all previously learned tasks:
\begin{equation}
\mathrm{MeanF}_i =
\frac{1}{i-1} \sum_{j=1}^{i-1} \mathrm{F}_{i}^{(j)} \;,
\label{eq:MeanF}
\end{equation}
where $i$ denotes the index of the current training stage. This formulation, originally introduced in~\cite{chaudhry2018riemannian},
quantifies the average degradation in performance on previously learned tasks up to the current training stage.
Lower values of $\mathrm{F}_i$ indicate reduced catastrophic forgetting.

\section{Results}
\label{sec:Results}

We evaluated the above CL approaches on a four-task sequential setting 
\(A \rightarrow B \rightarrow C \rightarrow D\) (see Fig.~\ref{fig:data_generation}g). 
At each stage \(t\), the model is trained only using task-\(t\) data without access to earlier tasks (unless otherwise stated). In the following sections, we present the results obtained at each training stage (A, B, C, and D), including both quantitative evaluation metrics and qualitative assessment. \edit{In addition to the results presented here, further visualizations of accuracy and forgetting are provided in the Supplementary Material (Figs. S1–S9). We also include a comprehensive sensitivity analysis of key hyperparameters for SLE-FNO, including the learning rate, batch size, data size, and weight decay in the Supplementary Material (Figs. S10–S11).}

\subsection{Stage 1: Training on Task A}

\begin{figure}[h!]
    \centering
    \includegraphics[width=1.0\textwidth]{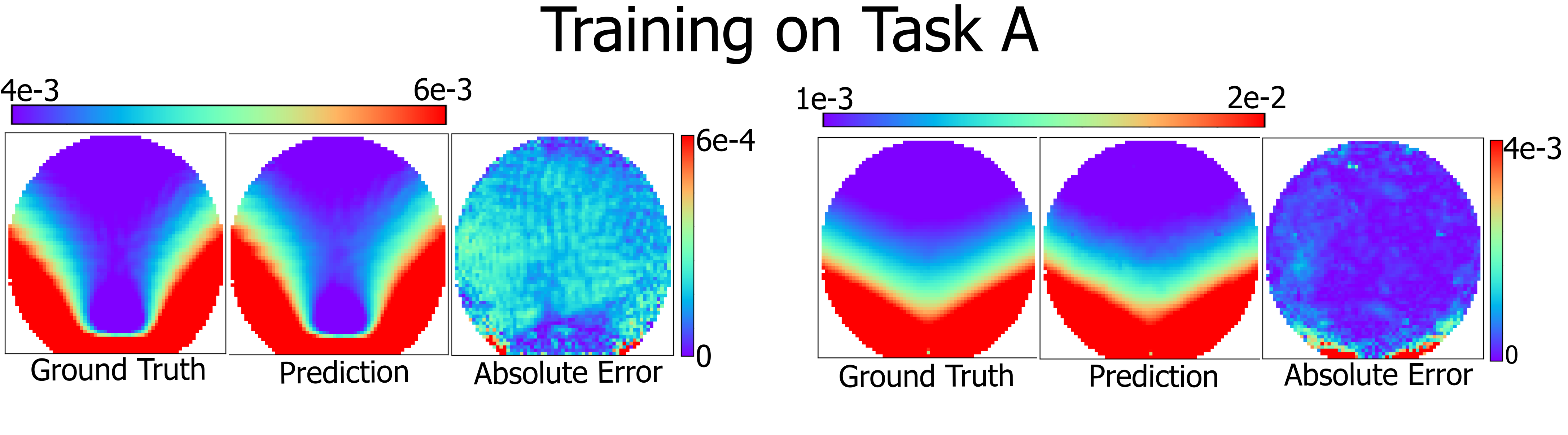}
    \caption{TAWSS results after the first stage (Training on A) for two randomly selected samples from task A. All units are in dynes/cm$^2$.}
    \label{fig:StageA}
\end{figure}

\edit{Table~\ref{tab:accuracies}--\ref{tab:forgetting} summarize the quantitative accuracy and forgetting metrics.} In the first stage, the model is trained on task A, which is the largest dataset (160 samples for training and 40 for testing). The model was trained for 3000 epochs, and the final accuracy on the test set was $R=$0.985. Since this is the first task in the sequence, there is no forgetting yet, so accuracy is reported only once.

For qualitative evaluation, Figure~\ref{fig:StageA} shows results for two randomly selected samples from the test set. The model demonstrates good qualitative performance with a small localized elevated quantitative error region near the junction between the aneurysm and the parent vessel.

%When examining the ground truth TAWSS distribution in these samples, the domain can be separated into three clear regions: a high-TAWSS region, a low-TAWSS region, and a transition region between them, consistent with typical hemodynamics patterns. The model performs well across all three areas, including the transition region, which is usually the most challenging due to the sharp spatial gradients . Only a small localized area contains a few pixels with higher error. This is expected, as the region corresponds to very high shear values close to the vessel wall, where predictions are more sensitive to boundary effects and mesh resolution. However, this does not significantly affect overall accuracy, and the high accuracy (0.9881) confirms this.

\subsection{Stage 2: Fine-Tuning on Task B}

\edit{
\begin{table}[h!]
\centering
\caption{\edit{Stage-wise accuracies and average accuracy, as defined in Eq.~\eqref{eq:AA_i}, for each method. Results are reported as mean $\pm$ standard deviation. Best values per stage are highlighted in bold. Each row (A--D) corresponds to a training stage, while each column represents the task accuracy after sequentially fine-tuning the model up to that task.}}
\label{tab:accuracies}
\begingroup
\edit{
\begin{tabular}{llccccc}
\hline
\textbf{Method} & \textbf{Stage} & \textbf{A} & \textbf{B} & \textbf{C} & \textbf{D} & \textbf{Avg. Acc.} \\
\hline
\multirow{4}{*}{Baseline}
 & A & 0.985 $\pm$ 0.000& - & - & - & 0.985\\
 & B & 0.663 $\pm$ 0.004 & \textbf{0.910} $\pm$ 0.027& - & - & 0.787 \\
 & C & 0.641 $\pm$ 0.011 & 0.718 $\pm$ 0.028 & \textbf{0.794} $\pm$ 0.088& - & 0.718 \\
 & D & 0.572 $\pm$ 0.018 & 0.265 $\pm$ 0.079 & 0.192 $\pm$ 0.103 & \textbf{0.864} $\pm$ 0.032& 0.473 \\
\hline
\multirow{4}{*}{SLE-FNO}
 & A & 0.985 $\pm$ 0.000& - & - & - & 0.985 \\
 & B & \textbf{0.985}  $\pm$ 0.000& 0.898 $\pm$ 0.024 & - & - & \textbf{0.942}\\
 & C & \textbf{0.985}  $\pm$ 0.000& \textbf{0.898} $\pm$ 0.024& 0.789 $\pm$ 0.050 & - & \textbf{0.891}\\
 & D & \textbf{0.985}  $\pm$ 0.000& \textbf{0.898} $\pm$ 0.024& \textbf{0.789} $\pm$ 0.050& 0.830 $\pm$ 0.022 & \textbf{0.876}\\
\hline
\multirow{4}{*}{LwF}
 & A & 0.985 $\pm$ 0.000& - & - & - & 0.985\\
 & B & 0.757 $\pm$ 0.011 & 0.888 $\pm$ 0.046 & - & - & 0.823 \\
 & C & 0.696 $\pm$ 0.012 & 0.748 $\pm$ 0.056 & 0.765 $\pm$ 0.058 & - & 0.736 \\
 & D & 0.653 $\pm$ 0.010 & 0.408 $\pm$ 0.032 & 0.370 $\pm$ 0.104 & 0.829 $\pm$ 0.019 & 0.565 \\
\hline
\multirow{4}{*}{EWC}
 & A & 0.985 $\pm$ 0.000& - & - & - & 0.985\\
 & B & 0.808 $\pm$ 0.015 & 0.895 $\pm$ 0.028 & - & - & 0.852 \\
 & C & 0.736 $\pm$ 0.012 & 0.763 $\pm$ 0.042 & 0.766 $\pm$ 0.067 & - & 0.755 \\
 & D & 0.756 $\pm$ 0.011 & 0.794 $\pm$ 0.047 & 0.750 $\pm$ 0.089 & 0.785 $\pm$ 0.020 & 0.771 \\
\hline
\multirow{4}{*}{Replay Reservoir}
 & A & 0.985 $\pm$ 0.000& - & - & - & 0.985\\
 & B & 0.898 $\pm$ 0.007 & 0.907 $\pm$ 0.026& - & - & 0.903\\
 & C & 0.852 $\pm$ 0.007 & 0.824 $\pm$ 0.074 & 0.766 $\pm$ 0.066 & - & 0.814 \\
 & D & 0.837 $\pm$ 0.006 & 0.830 $\pm$ 0.078 & 0.724 $\pm$ 0.089 & 0.862 $\pm$ 0.036 & 0.813 \\
\hline
\multirow{4}{*}{Replay K-means}
 & A & 0.985 $\pm$ 0.000& - & - & - & 0.985\\
 & B & 0.894 $\pm$ 0.005 & 0.908 $\pm$ 0.164& - & - & 0.901 \\
 & C & 0.857 $\pm$ 0.006 & 0.831 $\pm$ 0.036 & 0.774 $\pm$ 0.051 & - & 0.821\\
 & D & 0.841 $\pm$ 0.008& 0.854 $\pm$ 0.046 & 0.740 $\pm$ 0.027 & 0.860 $\pm$ 0.037 & 0.824\\
\hline
\multirow{4}{*}{OGD}
 & A & 0.985 $\pm$ 0.000& - & - & - & 0.985\\
 & B & 0.836 $\pm$ 0.013 & 0.901 $\pm$ 0.024 & - & - & 0.869 \\
 & C & 0.797 $\pm$ 0.022 & 0.725 $\pm$ 0.052 & 0.772 $\pm$ 0.087 & - & 0.765 \\
 & D & 0.654 $\pm$ 0.015 & 0.390 $\pm$ 0.054 & 0.308 $\pm$ 0.115 & 0.859 $\pm$ 0.034 & 0.553 \\
\hline
\multirow{4}{*}{GEM}
 & A & 0.985 $\pm$ 0.000& - & - & - & 0.985\\
 & B & 0.915 $\pm$ 0.002& 0.871 $\pm$ 0.053 & - & - & 0.893 \\
 & C & 0.856 $\pm$ 0.004 & 0.681 $\pm$ 0.098 & 0.768 $\pm$ 0.060 & - & 0.768 \\
 & D & 0.754 $\pm$ 0.012 & 0.500 $\pm$ 0.050 & 0.420 $\pm$ 0.086 & 0.847 $\pm$ 0.032 & 0.630 \\
\hline
\multirow{4}{*}{PiggyBack}
 & A & 0.985  $\pm$ 0.000& - & - & - & 0.985\\
 & B & \textbf{0.985}  $\pm$ 0.000 & 0.891 $\pm$ 0.028 & - & - & 0.938 \\
 & C & \textbf{0.985}  $\pm$ 0.000 & 0.891 $\pm$ 0.028 & 0.781 $\pm$ 0.083 & - & 0.886 \\
 & D & \textbf{0.985}  $\pm$ 0.000 & 0.891 $\pm$ 0.028 & 0.781 $\pm$ 0.083 & 0.832 $\pm$ 0.021 & 0.872 \\
\hline
\multirow{4}{*}{LoRA}
 & A & 0.985  $\pm$ 0.000& - & - & - & 0.985\\
 & B & \textbf{0.985}  $\pm$ 0.000 & 0.894 $\pm$ 0.062 & - & - & 0.940 \\
 & C & \textbf{0.985}  $\pm$ 0.000 & 0.894 $\pm$ 0.062 & 0.784 $\pm$ 0.045& - & 0.888 \\
 & D & \textbf{0.985}  $\pm$ 0.000 & 0.894 $\pm$ 0.062 & 0.784 $\pm$ 0.045& 0.838 $\pm$ 0.033& 0.875\\
\hline
\end{tabular}
}
\endgroup
\end{table}
}

\edit{

\begin{table}[h!]
\centering
\caption{\edit{Forgetting measure $F$ (lower is better), defined in Eq.~\eqref{eq:FM_taskwise}, and the average forgetting measure (MeanF), defined in Eq.~\eqref{eq:MeanF}, are reported for each method and stage. Results are presented as mean $\pm$ standard deviation. MeanF is computed using the mean forgetting values only. Each row (B, C, D) corresponds to a training stage after sequential fine-tuning up to that task.}}
\label{tab:forgetting}
\begingroup
\edit{
\begin{tabular}{llccccc}
\toprule
Method & Stage & $F_A$ & $F_B$ & $F_C$ & $F_D$ & MeanF \\
\midrule
\multirow{3}{*}{Baseline}
 & B & 0.322 $\pm$ 0.004 & - & - & - & 0.322 \\
 & C & 0.343 $\pm$ 0.012 & 0.192 $\pm$ 0.029 & - & - & 0.268 \\
 & D & 0.412 $\pm$ 0.026 & 0.635 $\pm$ 0.079 & 0.579 $\pm$ 0.127 & - & 0.542 \\
\midrule
\multirow{3}{*}{SLE-FNO}
 & B & 0 & - & - & - & 0 \\
 & C & 0 & 0 & - & - & 0 \\
 & D & 0 & 0 & 0 & - & 0 \\
\midrule
\multirow{3}{*}{LwF}
 & B & 0.227 $\pm$ 0.011 & - & - & - & 0.227 \\
 & C & 0.289 $\pm$ 0.012 & 0.139 $\pm$ 0.038 & - & - & 0.214 \\
 & D & 0.332 $\pm$ 0.010 & 0.479 $\pm$ 0.050 & 0.383 $\pm$ 0.071 & - & 0.398 \\
\midrule
\multirow{3}{*}{EWC}
 & B & 0.176 $\pm$ 0.016 & - & - & - & 0.176 \\
 & C & 0.248 $\pm$ 0.013 & 0.131 $\pm$ 0.056 & - & - & 0.190 \\
 & D & 0.229 $\pm$ 0.011 & 0.100 $\pm$ 0.057 & 0.014 $\pm$ 0.030 & - & 0.114 \\
\midrule
\multirow{3}{*}{Replay Reservoir}
 & B & 0.086 $\pm$ 0.007 & - & - & - & 0.086 \\
 & C & 0.132 $\pm$ 0.007 & 0.079 $\pm$ 0.068 & - & - & 0.106 \\
 & D & 0.147 $\pm$ 0.005 & 0.074 $\pm$ 0.066 & 0.039 $\pm$ 0.041 & - & 0.086 \\
\midrule
\multirow{3}{*}{Replay K-means}
 & B & 0.090 $\pm$ 0.005 & - & - & - & 0.090 \\
 & C & 0.128 $\pm$ 0.006 & 0.077 $\pm$ 0.033 & - & - & 0.102 \\
 & D & 0.143 $\pm$ 0.008 & 0.053 $\pm$ 0.037 & 0.035 $\pm$ 0.049 & - & 0.077 \\
\midrule
\multirow{3}{*}{OGD}
 & B & 0.149 $\pm$ 0.013 & - & - & - & 0.149 \\
 & C & 0.188 $\pm$ 0.022 & 0.175 $\pm$ 0.053 & - & - & 0.181 \\
 & D & 0.330 $\pm$ 0.015 & 0.508 $\pm$ 0.073 & 0.449 $\pm$ 0.102 & - & 0.429 \\
\midrule
\multirow{3}{*}{GEM}
 & B & 0.069 $\pm$ 0.002 & - & - & - & 0.069 \\
 & C & 0.129 $\pm$ 0.004 & 0.184 $\pm$ 0.067 & - & - & 0.156 \\
 & D & 0.231 $\pm$ 0.012 & 0.370 $\pm$ 0.074 & 0.341 $\pm$ 0.069 & - & 0.314 \\
\midrule
\multirow{3}{*}{PiggyBack}
 & B & 0 & - & - & - & 0 \\
 & C & 0 & 0 & - & - & 0 \\
 & D & 0 & 0 & 0 & - & 0 \\
\midrule
\multirow{3}{*}{LoRA}
 & B & 0 & - & - & - & 0 \\
 & C & 0 & 0 & - & - & 0 \\
 & D & 0 & 0 & 0 & - & 0 \\
\bottomrule
\end{tabular}
}
\endgroup
\end{table}

}

In the second stage, the pretrained model from task A is fine-tuned on task B. This task is much smaller and contains only 8 training samples and 2 test samples, which makes it a useful case for studying how different CL methods handle low-data adaptation. The model was fine-tuned for 1500 epochs \edit{across five different data splits and two random seeds, ten runs in total}, and the results for accuracy and forgetting are summarized in Table~\ref{tab:accuracies} and Table~\ref{tab:forgetting}.

The baseline model achieves the highest accuracy on task B with a value of \edit{0.910}. However, this comes at the cost of forgetting; the accuracy on task A drops to \edit{0.663}, and the forgetting measure at this stage is \edit{0.322}, which is the worst among all methods. This confirms the typical behavior of naive fine-tuning, where the model quickly adapts to new data but overwrites previously learned knowledge.

Regularization-based methods such as EWC and LwF show a more balanced behavior. Their accuracy on task B (\edit{0.895 and 0.888}, respectively) is close to the baseline, but forgetting is reduced. Their forgetting values for task A are \edit{0.176} (EWC) and \edit{0.227} (LwF), showing that both methods offer mild protection of previously learned knowledge.

Optimization-based methods follow a similar pattern. OGD and GEM perform well on task B (\edit{0.901} and \edit{0.871}, respectively) and \edit{maintain} strong retention of task A. OGD achieves \edit{0.149} forgetting, while GEM performs even better with \edit{0.069} forgetting. This aligns with the design of GEM, which replays gradients from earlier tasks and therefore preserves knowledge more effectively.

Replay-based methods also perform well at this stage. Both reservoir replay and k-means replay achieve accuracies close to the baseline (\edit{0.907} and \edit{0.908}, respectively). Their forgetting values (\edit{0.086} and \edit{0.090}, respectively) \edit{show clear improvement} over baseline.

Architecture-based approaches such as LoRA and PiggyBack perform differently from the above methods. Their accuracy on task B is competitive (\edit{0.894} and \edit{0.891}, respectively), and they show no forgetting at all (F = 0). This is expected, since neither approach modifies the pretrained model weights and instead, they add small trainable components on top of the existing network. SLE-FNO behaves similarly, with an accuracy of \edit{0.898} on task B and no forgetting.
While all three methods prevent forgetting, they differ in expressivity and design complexity. LoRA relies on low-rank adapters that require careful tuning and placement and can suffer when task-specific adaptations are not well captured by a low-rank approximation, leading to reduced expressivity in complex distribution shifts~\cite{hu2022lora}. PiggyBack applies task-specific scalar or binary masks that can restrict model expressivity and, in practice, may lead to reduced accuracy under complex distribution shifts~\cite{mallya2018PiggyBack}, whereas SLE-FNO introduces a single task-specific extension layer aligned with the backbone architecture, offering a favorable balance between expressivity, simplicity, and CL stability.

\begin{figure}[h!]
    \centering
    \includegraphics[width=0.8\textwidth]{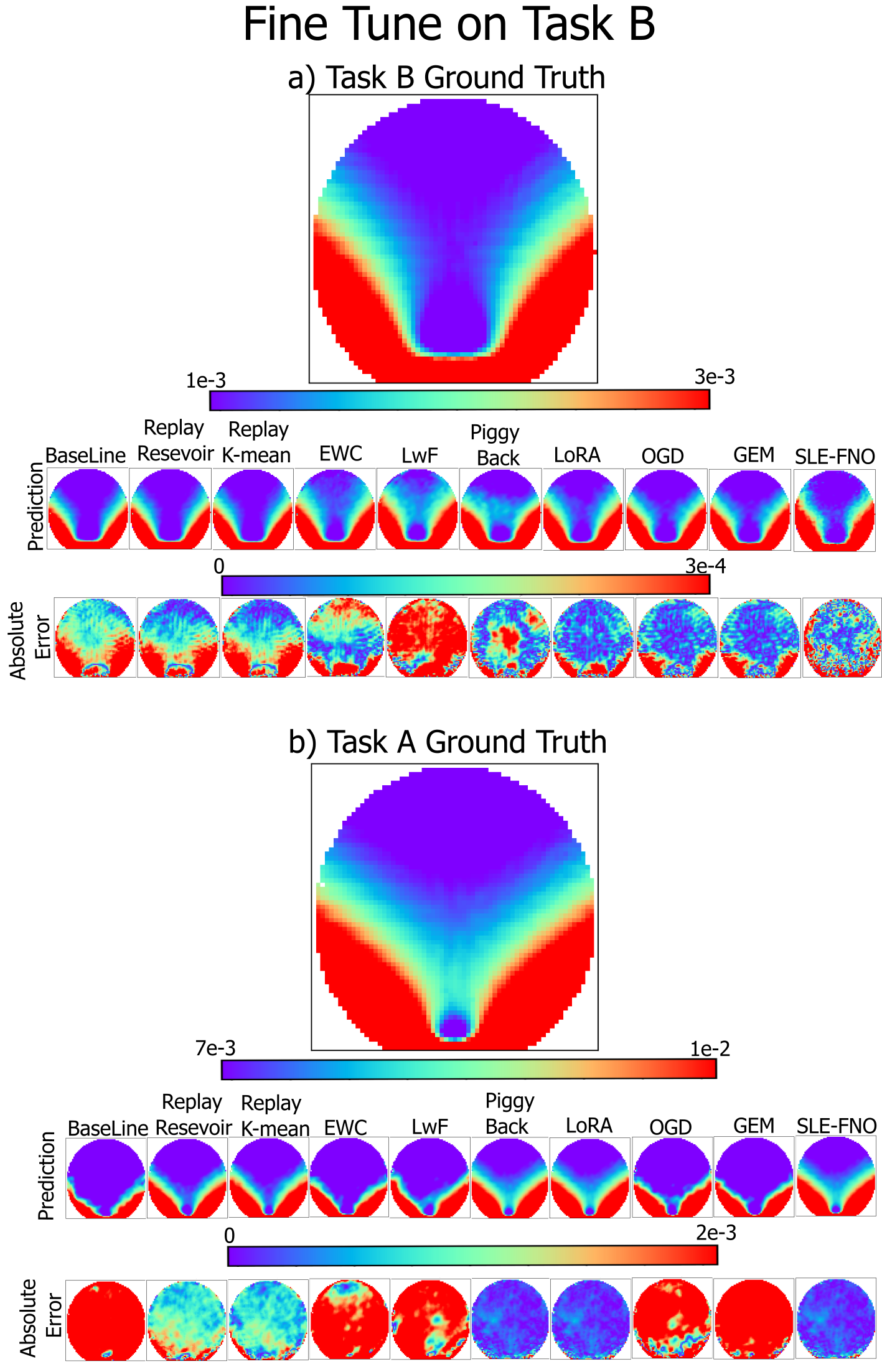}
    \caption{TAWSS results at the second stage (Fine-Tuning at task B). (a) Results for the worst-performing sample from task B. (b) Results for a representative sample from task A. All units are in dynes/cm$^2$.}
    \label{fig:StageB}
\end{figure}

Qualitative results for task B are shown in Fig.~\ref{fig:StageB}. Most methods produce predictions that closely match the ground truth. Small deviations appear in EWC, LwF, and PiggyBack, particularly in low-value regions. The baseline and replay-based methods show the smoothest transitions, while OGD and GEM predictions appear slightly noisier in the transition region. SLE-FNO achieves an excellent qualitative preservation of features of task B although it exhibits slight noise in the transition region. When observing accuracy on task A (the previous task), the qualitative trends match the quantitative results. The baseline shows clear degradation, especially in the transition region between low and high values. EWC and LwF retain structure more successfully, while OGD and GEM recover high-value TAWSS regions more accurately. Replay-based methods produce strong visual consistency across all regions and are the best-performing group among methods that overwrite model weights. Finally, SLE-FNO, LoRA, and PiggyBack preserve task A perfectly, as expected from their zero-forgetting behavior.

\subsection{Stage 3: Fine-Tuning on Task C}

In the third stage, the model from stage 2 is now further fine-tuned on task C. As expected, naive fine-tuning (baseline) achieves the highest accuracy on the new task \edit{(0.794)} but also exhibits the largest mean forgetting \edit{(0.268)}.

\edit{Both regularization-based CL methods demonstrate competitive performance. EWC achieves an accuracy of 0.766 while effectively reducing forgetting (MeanF = 0.190). In comparison, LwF shows similar adaptation at this stage (accuracy = 0.765) but exhibits weaker retention, as indicated by a higher MeanF of 0.214.}

% Among regularization-based methods, EWC performs well, reaching an accuracy of 0.8659 and reducing the mean forgetting to 0.1991. LwF, however, shows weaker adaptation at this stage with an accuracy of 0.8353 and mean forgetting of 0.236.

Replay-based approaches \edit{achieve high learning accuracy for the new task}, with Reservoir and K-means Replay achieving accuracies of \edit{0.766 and 0.774}, respectively. Both replay methods maintain strong retention, with mean forgetting values of \edit{0.106} (Reservoir) and \edit{0.102} (K-means). Optimization-based methods achieve \edit{similar} accuracy on the new task compared to replay approaches. OGD and GEM reach accuracies of \edit{0.772} and \edit{0.768} respectively, while also demonstrating excellent mean forgetting of \edit{0.181} for OGD and \edit{0.156} for GEM.

Architecture-based approaches show \edit{similar} accuracy. PiggyBack obtains an accuracy of \edit{0.781}, and LoRA improves slightly to \edit{0.784}, suggesting that the architectural expansion or adaptation provided is insufficient to fully capture task C. Finally, SLE-FNO performs \edit{slightly better than LoRA and PiggyBack}, achieving the second-highest accuracy \edit{0.789} on task C, while achieving the best average accuracy at this C stage. It also maintains zero forgetting, a key advantage it shares with the other architecture-based methods.

\begin{figure}[h!]
    \centering
    \includegraphics[width=0.5\textheight]{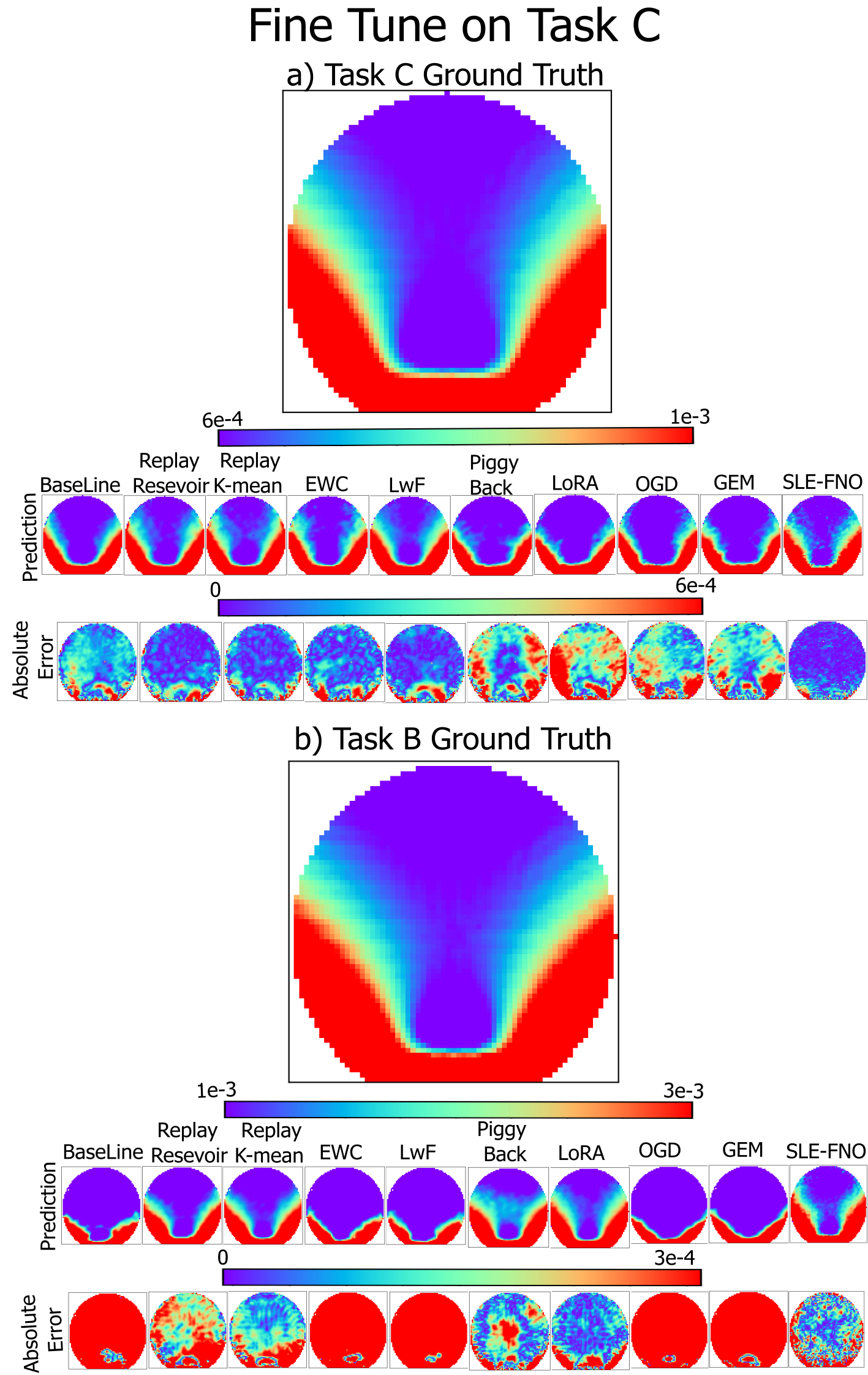}
    \caption{TAWSS results at the third stage (Fine-Tuning on C). (a) Worst-case sample from task C. (b) Worst-case sample from task B. All units are in dynes/cm$^2$.}
    \label{fig:StageC-1}
\end{figure}

\begin{figure}[h!]
    \centering
    \includegraphics[width=0.5\textheight]{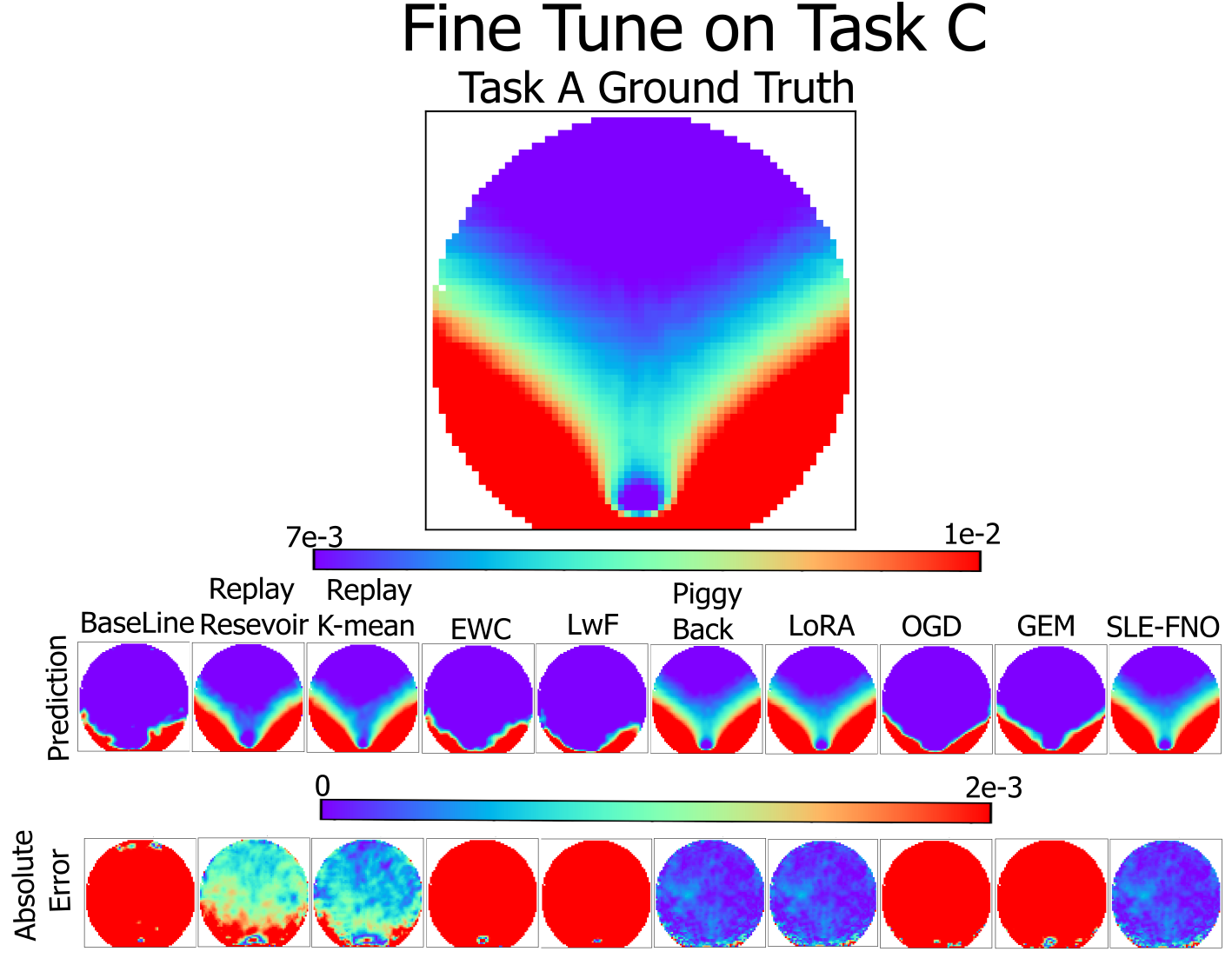}
    \caption{TAWSS results at the third stage (Fine-Tuning on C). A random sample from task A is shown. All units are in dynes/cm$^2$.}
    \label{fig:StageC-2}
\end{figure}

Qualitatively, and as shown in Fig.~\ref{fig:StageC-1} and Fig.~\ref{fig:StageC-2}, the prediction patterns for stage C reveal distinct behaviors across methods. For task C, the baseline results appear noticeably skewed, particularly around the transition zone. In contrast, both replay-based and regularization-based methods produce a smoother and more continuous transition between high and low TAWSS regions. Architecture-based methods, however, deviate slightly in the global contour. The optimization-based methods (OGD and GEM) generate very similar results and they both accurately identify the high TAWSS region but tend to overestimate lower TAWSS values, resulting in a thinner transition boundary. SLE-FNO, on the other hand, successfully preserves the main structural outline, although its output still shows some unevenness and reduced smoothness in the transitional region.

For the previous tasks (A and B), the baseline produces only a coarse outline of the three predominant regions, with a narrow and abrupt transition zone, indicating severe forgetting. A similar pattern appears in the optimization-based methods and, to a lesser extent, in regularization-based approaches. Replay-based algorithms continue to preserve high fidelity for the earlier tasks, demonstrating strong retention. Finally, architecture-based methods and SLE-FNO maintain stable accuracy and exhibit negligible degradation, consistent with their zero-forgetting behavior reported earlier.

\subsection{Stage 4: Fine-Tuning on Task D}

\begin{figure}[h!]
    \centering
    \includegraphics[width=0.8\textwidth]{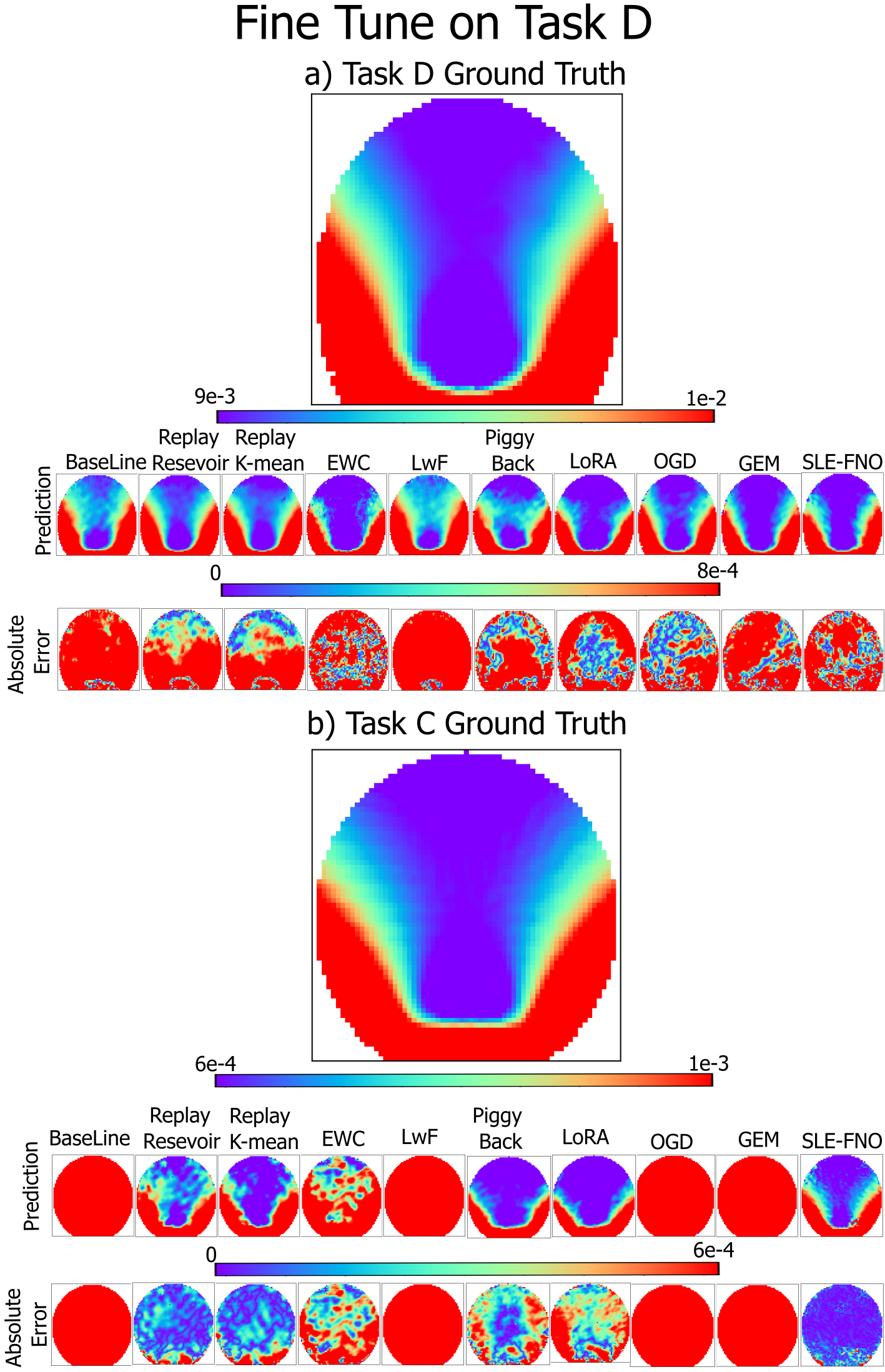}
    \caption{TAWSS results at the fourth stage (Fine-Tuning on task D). (a) Worst-case sample from task D. (b) Worst-case sample from task C. All units are in dynes/cm$^2$.}
    \label{fig:StageD-1}
\end{figure}

\begin{figure}[h!]
    \centering
    \includegraphics[width=0.8\textwidth]{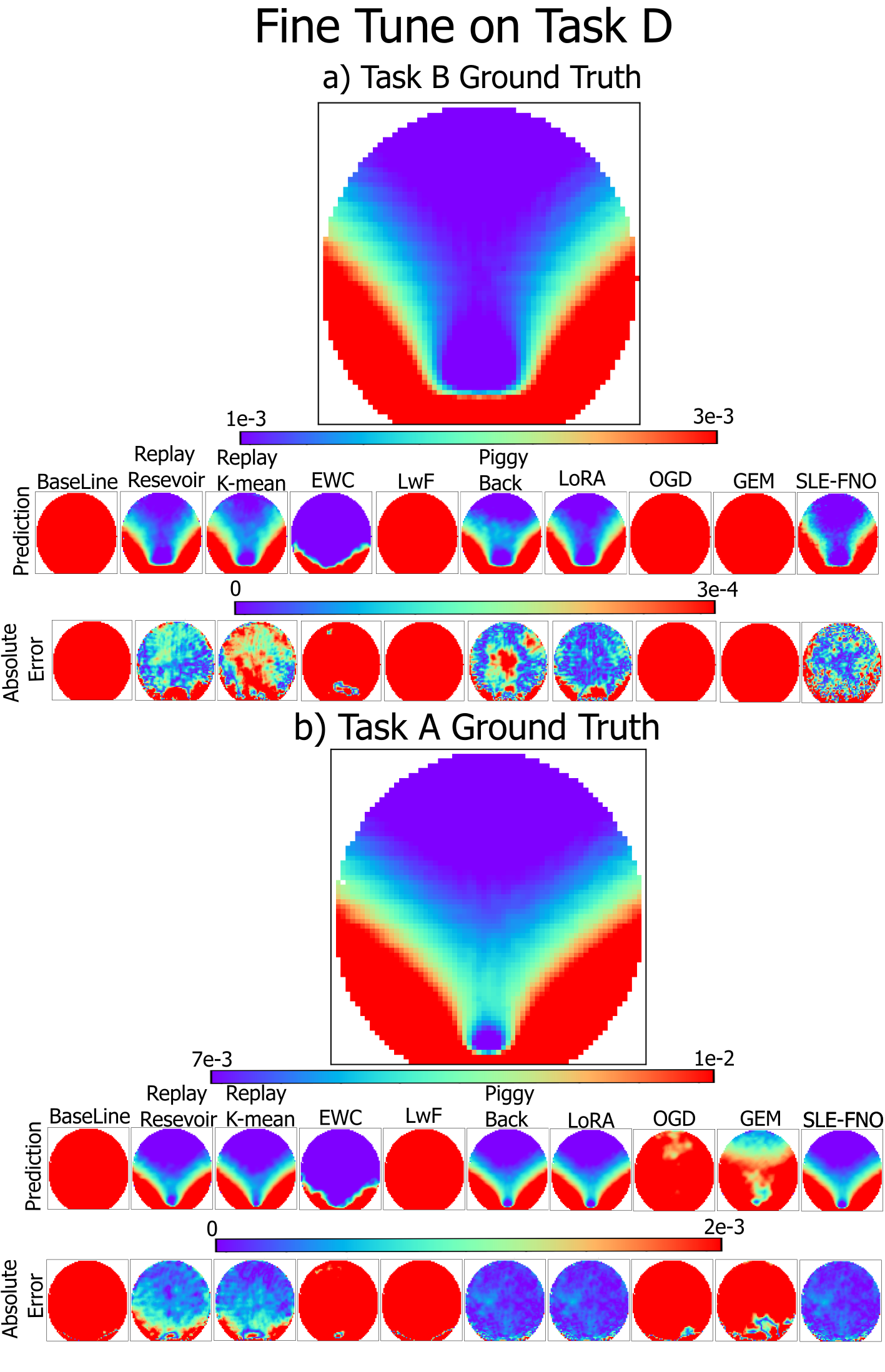}
    \caption{TAWSS results at the fourth stage (Fine-Tuning on task D). (a) Worst-case sample from task B. (b) Random sample from task A. All units are in dynes/cm$^2$.}
    \label{fig:StageD-2}
\end{figure}

The final stage involves further fine-tuning on task D, In terms of accuracy on the new task, the gap between methods becomes smaller compared to previous stages. The baseline achieves an accuracy of \edit{0.864}, representing the highest accuracy on task D, but also suffers from the worst forgetting, with a final mean forgetting score of \edit{0.542}. The second best new-task accuracy is achieved by Replay Reservoir with \edit{0.862}, only marginally \edit{lower} than the baseline. The lowest accuracy among methods is obtained by EWC at \edit{0.785}.

Looking at the final average accuracy across all tasks, architecture-based methods dominate. SLE- achieves the highest overall accuracy with \edit{0.876}, followed by LoRA with \edit{0.875}, and PiggyBack with \edit{0.872}. Their advantage is largely due to the absence of forgetting because of zero interference with previously learned tasks, achieved through task-specific parameter allocations.

Replay-based methods and EWC form a middle tier, with final average accuracies of \edit{0.813} (Reservoir Replay), \edit{0.824} (K-means Replay), and \edit{0.771} (EWC). Their corresponding final mean forgetting measures are \edit{0.114} (EWC), \edit{0.086} (Reservoir), and \edit{0.077} (K-means Replay), showing reasonable but non-zero forgetting. LwF performs poorly at this stage, dropping to a final average accuracy of only \edit{0.565}, \edit{slightly better than the baseline}, and displaying severe final mean forgetting \edit{0.398}, indicating its difficulty adapting to highly shifted task distributions.

Optimization-based methods show \edit{degraded} performance. OGD achieves a final accuracy of \edit{0.553} with mean forgetting of \edit{0.429}, while GEM achieves improved \edit{0.630} average accuracy and \edit{0.314} mean forgetting. This suggests that explicitly retaining sample-based constraints (as in GEM) is more effective than relying solely on gradient-space orthogonalization (as in OGD) when facing highly complex downstream distributions. \edit{Finally, as expected, PiggyBack, LoRA, and SLE-FNO demonstrate zero forgetting, with nearly identical final average accuracies across all methods.}

Qualitative evaluation of the predictions at this stage (Fig.~\ref{fig:StageD-1} and Fig.~\ref{fig:StageD-2}) reveals distinct behavior across methods for task D. The baseline model tends to overestimate the low-TAWSS region, resulting in an exaggerated area of low shear. In contrast, replay-based methods produce more realistic spatial patterns. This improvement is likely due to the presence of replayed OOD samples, which act as a form of implicit regularization and help guide the model toward smoother and more generalizable predictions on previously unseen test samples.

Among the regularization-based methods, EWC produces a noticeably compressed transition region, while LwF behaves similarly to the baseline and also overestimates low-TAWSS values. The architecture-based methods differ where PiggyBack fails to capture the main structural features of the flow patterns, whereas LoRA behaves more similarly to EWC, showing a reduced transition zone but with improved smoothness.

The optimization-based methods (OGD and GEM) perform well qualitatively, with OGD in particular closely matching the ground truth spatial distribution. SLE-FNO successfully captures the overall flow structure and produces the closest qualitative result to the ground truth.

For the previous tasks (C, B, and A), catastrophic forgetting becomes visually apparent. For tasks C and B, the predictions from the baseline, LwF, OGD, and GEM are almost completely misaligned with the ground truth. For task A, the optimization-based methods (OGD and GEM) reconstruct only small portions of the original pattern. Replay-based approaches, especially K-means Replay, recover the closest qualitative match for all three earlier tasks. Reservoir Replay performs similarly but with slightly reduced fidelity, particularly for tasks C and B. Finally, EWC retains partial structural information for task C, though the output remains noisy.

\section{Discussion}

In this study, we evaluated a wide range of CL algorithms on a practical physical operator learning setting consisting of four sequential tasks (A, B, C, and D). Each model was trained in a strictly sequential setting, where access to previous task data was prohibited unless explicitly allowed by the corresponding method. The evaluated approaches covered regularization-based methods (LwF~\cite{li2017learning}, EWC~\cite{kirkpatrick2017overcoming}), optimization-based methods (GEM~\cite{lopez2017gradient}, OGD~\cite{farajtabar2020ogd}), replay-based strategies, and architecture-based parameter-efficient approaches such as (PiggyBack~\cite{mallya2018PiggyBack}, LoRA~\cite{hu2022lora}), in addition to our proposed SLE-FNO framework. Performance was assessed using both quantitative accuracy metrics and qualitative spatial error maps. Unlike many CL benchmarks that assume large and balanced datasets for each task, our setup reflects a more realistic SciML workflow, where training begins with a large pretraining dataset and is followed by smaller OOD datasets corresponding to new operating conditions or geometries. \edit{Importantly, this setting is not specific to CFD, but is representative of a broad class of scientific operator learning problems where the physics evolve over time due to changes in geometry, boundary conditions, or physical parameters.} This uneven and sequential data availability \edit{also} mirrors real-world scenarios in computational biomechanics and digital twin modeling, where new simulations or clinical scenarios may arrive gradually and often sparsely.

During the early stage at task~B, most CL methods perform similarly, since only a single prior task must be retained and forgetting remains limited. Nonetheless, an early ranking is observable, with \edit{replay-based methods and GEM} showing the strongest retention, followed by \edit{OGD}, while regularization-based methods exhibit weaker protection against forgetting. As training progresses to task~C, the problem becomes more challenging due to a significant shift in the flow regime associated with lower Reynolds number and altered hemodynamic patterns. At this stage, method-specific differences become pronounced, where \edit{regularization-based methods} degrades noticeably as constraints from multiple prior tasks restrict model flexibility. Replay-based methods remain effective but require careful buffer composition, particularly to adequately represent the broad distribution of task~A. Optimization-based methods such as GEM and OGD continue to achieve strong accuracy but incur increased \edit{forgetting compared to regularization-based methods}. Architecture-based methods, including LoRA and PiggyBack, successfully eliminate forgetting without replay buffers, though PiggyBack and LoRA show limited expressive capacity especially for qualitative results. SLE-FNO achieves the highest overall \edit{average} accuracy, with minor localized reductions in smoothness qualitatively. By task~D, long-term scalability effects become evident, where LwF fails to retain early knowledge, EWC exhibits \edit{slight} degradation, and optimization-based methods show reduced stability over extended task sequences. Replay-based methods continue to perform strongly but remain impractical for many medical settings due to privacy and storage requirements. In contrast, architecture-based approaches such as LoRA, PiggyBack, and SLE-FNO preserve prior knowledge without replay, making them well-suited for real-world CL scenarios under significant distributional shift.

Our findings align closely with recent comparative studies in both scientific regression and classification. In 3D engineering regression, it has been recently shown that replay-based methods consistently outperformed EWC and GEM, achieving near joint-training accuracy while reducing training time~\cite{samuel2025cl3d}. EWC failed to prevent forgetting in most cases, and GEM showed intermediate performance, while exhibiting some difficulty in adapting to new tasks and incurring high computational costs. This directly matches our observations \edit{across the stages}, where the replay method achieved \edit{stronger retention compared to EWC, except at stage~D}. In building energy forecasting, Li et al.~\cite{li2023large} demonstrated that CL significantly improved long-term prediction accuracy compared to static and cumulative retraining approaches. In their setting, EWC performed well because the underlying dynamics evolved slowly over time. This highlights an important distinction; regularization-based methods are effective when distribution shifts are smooth and gradual, but may become insufficient under large, abrupt shifts, such as those encountered in our blood flow tasks especially \edit{LwF}. In a 3D object classification study~\cite{resani2024miracle3d}, the authors compared EWC and LwF against their proposed shape-model–based method, Their results showed that EWC effectively mitigated catastrophic forgetting, maintaining strong task retention, although its performance degraded in later tasks. In contrast, LwF improved over the non-regularized baseline but performed significantly worse than EWC, particularly in later tasks, where it exhibited a faster increase in prediction error. These observations closely align with our findings where EWC achieves strong overall performance, despite a slight loss of accuracy on earlier tasks, whereas LwF shows substantially weaker retention, ranking just above the baseline method. In a recent study on adaptive learning for building thermal dynamics~\cite{raisch2025adapting}, it was reported that EWC achieved the best overall performance among standard CL methods, providing stable accuracy improvements and robust behavior at a low computational cost. GEM showed competitive accuracy but required significantly higher computation, while LwF demonstrated weaker long-term adaptation and performance degradation as updates accumulated. These observations align with our findings, where EWC and GEM perform well in terms of accuracy, but GEM incurs higher computational overhead due to gradient projection operations, and LwF proves less efficient and less effective in long-term CL scenarios. We extended these previous studies by evaluating two representative methods from each major CL category, enabling a broader and more systematic qualitative comparison. Unlike prior work that often depends on large replay buffers or extensive retraining, our approach fine-tunes each task using a minimal number of samples, emphasizing data efficiency and practical applicability. Moreover, we introduce a fully task-agnostic inference framework based on a KPCA-based task identifier, which automatically selects the appropriate SLE-FNO during inference without requiring task labels. 

%This combination makes the proposed framework ready for real-world deployment, as it couples CL with automatic routing at inference time.

Large-scale vision benchmarks further support these conclusions. De Lange et al.~\cite{de2021continual} showed that replay-based methods consistently achieve the highest final accuracy in class-incremental learning, while regularization-based approaches such as EWC and LwF generally lag behind. Architecture-based methods, including PackNet~\cite{mallya2018packnet}, which iteratively prune task-relevant parameters to reserve capacity for future tasks and retrain the remaining weights to recover performance, and HAT~\cite{serra2018overcoming}, which learns task-specific attention masks to gate neurons and prevent interference between tasks, can almost completely eliminate forgetting. However, both approaches are constrained by finite model capacity and implicitly limit the number of tasks that can be accommodated. These trends match our findings where replay methods achieve the strongest accuracy, regularization methods remain lightweight but comparatively weaker, and architectural approaches such as SLE-FNO strike a strong balance by providing high retention and adaptability without storing raw data.

From a practical standpoint, our results suggest that for long task sequences, architecture-based methods, especially expansion or dynamic ones, such as SLE-FNO are recommended when model size is manageable. If replay is possible, or when generative replay is available, replay-based strategies offer strong accuracy and robustness. Regularization-based methods, \edit{especially LwF,} remain computationally attractive but increasingly struggle as task count and distributional shift grow. It is important to note that we trained our models using a Sobolev norm while reporting accuracy using relative $\ell_{2}$ error. As a result, numerical accuracy alone does not fully reflect qualitative smoothness, since the Sobolev loss explicitly enforces spatial regularity. This explains why some methods show similar $\ell_{2}$ errors but produce different spatial error structures.

From our comparative investigation, we observe trade-offs across CL paradigms. Regularization-based methods occupy a middle ground; EWC provides reasonable retention, while LwF consistently underperforms, particularly under strong distribution shifts. Interestingly, we observe that EWC can occasionally improve performance on earlier tasks. In our experiments, we observe a partial recovery of performance on tasks A and B when progressing from stage 2 (fine-tune on task C) to stage 3 (fine-tune on task D). This behavior is consistent with the formulation of EWC, which constrains updates along directions deemed important for previously learned tasks rather than enforcing strict freezing of parameters. As discussed in~\cite{kirkpatrick2017overcoming}, such elastic regularization can reduce destructive interference and bias optimization toward parameter regions that remain compatible with earlier solutions, allowing limited performance recovery when tasks are related.

Architecture-based approaches, particularly those that isolate task-specific parameters such as PiggyBack and LoRA, mitigate forgetting by design. However, PiggyBack is limited in expressivity due to its scalar masking mechanism, while LoRA introduces additional hyperparameters that require careful tuning, and its low-rank parameterization can limit expressivity when the task-specific adaptations cannot be well represented in a low-rank subspace~\cite{hu2022lora}.
In contrast, our proposed method achieves zero forgetting, similar to LoRA and PiggyBack, while maintaining strong learning on new tasks (with significantly fewer hyperparameters), making it competitive in practice. Moreover, SLE-FNO can be viewed as a parameter-efficient extension of architecture-based model-expansion methods, where task-specific capacity is added in a controlled and lightweight manner. In our example, SLE-FNO increased the total parameter count by only 1.5\% for tasks B and C, and 4.4\% for the more challenging task D. \edit{
Our approach is related to parameter-efficient CL methods that freeze a shared backbone and introduce task-specific modules, including mask-based methods, LoRA variants~\cite{liang2024inflora,he2025cl,wei2025online,smith2023continual}, and prompt-/adapter-based strategies~\cite{wang2022dualprompt,smith2023coda}. In contrast, SLE-FNO performs adaptation in the spectral domain of a FNO, leveraging spectral inductive bias, unlike prior methods operating in weight or feature space. Related spectral approaches~\cite{gao2024parameter,zhang2024spectral,zhang2025f} focus on weight-space adaptation or single-task settings. Moreover, SLE-FNO targets scientific operator learning rather than discrete tasks and employs task-agnostic KPCA-based routing, unlike prior routing methods that rely on task identity or learned gating~\cite{yu2024boosting}.
}

In our example, we focused on the task of TAWSS prediction based on time-resolved concentration data. TAWSS is a well-established hemodynamic biomarker in cardiovascular disease~\cite{mahmoudi2021story}. However, estimating TAWSS directly inside the body using imaging remains challenging, due to low spatial resolutions and imaging artifacts. Computed tomography angiography (CTA) can capture the transport of contrast agents within the vasculature, which may be interpreted as a passive scalar advected by the underlying flow and therefore encodes information about local hemodynamics. Our presented example was motivated by this imaging modality and demonstrates the possibility of estimating TAWSS based on concentration data, which could come from experimental techniques such as dynamic CTA imaging.

%To address this limitation, we propose an adaptive lifelong surrogate model that maps passive scalar concentration measurements obtained from CTA directly to TAWSS. By leveraging contrast dynamics as a proxy for flow behavior, our approach enables rapid, non-invasive estimation of clinically relevant hemodynamic metrics without requiring full CFD simulations. This application highlights how CL architectures such as SLE-FNO can deliver strong retention and adaptability while remaining suitable for real-world clinical deployment.

The proposed framework \edit{has the potential to} support adaptive digital-twin modeling, where a surrogate model must evolve over time as new data, regimes, or operating conditions become available, without sacrificing previously acquired knowledge. In practical scientific settings, such as hemodynamic monitoring, climate modeling, or engineering system surveillance, digital twins are exposed to non-stationary data streams driven by changes in boundary conditions, geometry, or physical parameters~\cite{zohdi2020machine,tasmurzayev2025digital}. Our CL formulation \edit{potentially motivates} enabling the digital twin to incrementally incorporate these shifts by allocating lightweight, task-specific capacity while preserving a shared physical representation in the backbone. This design avoids repeated full retraining and mitigates catastrophic forgetting, which is critical for maintaining long-term consistency. \edit{
From a deployment perspective, this enables practical workflows where models can be incrementally updated as new simulations or measurements become available, without requiring full retraining or access to historical data. This is particularly relevant in settings with data privacy constraints or limited storage, such as clinical environments, where only small amounts of new data may be accessible over time.
}

\edit{
The proposed framework has several limitations that should be acknowledged. First, the evaluation is conducted on a relatively limited dataset, which may not fully capture the variability and complexity encountered in real-world scenarios. In addition, while SLE-FNO provides an efficient parameter-isolation mechanism, it requires introducing a new task-specific branch for each incoming task, leading to a linear growth in model size as the number of tasks increases, which may limit scalability in long CL sequences. Although the method is, in principle, applicable across different scientific domains, the current study is restricted to idealized geometries and clean simulation data without measurement noise, leaving its robustness under noisy or imperfect observations untested. Furthermore, the approach relies heavily on the shared backbone to capture the dominant components of the solution, meaning that under large distribution shifts, where the underlying physics or patterns deviate significantly, the model may struggle to generalize effectively. Finally, the framework is demonstrated on a single application setting, and extending the evaluation to diverse domains and real-world systems would provide stronger evidence of its practical utility.
}

Future work can extend the current framework to more complex, patient-specific aneurysm geometries, where flow patterns may differ significantly from the idealized cases considered in this study. Additionally, exploring the use of generative models for replay could eliminate the need to store raw simulation data, thereby improving the practicality and scalability of replay-based strategies. Finally, systematic ablation studies examining the impact of replay buffer size, particularly the number of samples retained per task, would provide valuable insight into memory–performance trade-offs and their influence on overall model performance. Another important direction is to compare the performance of FNO with other neural operator and deep learning architectures under the same CL setup, in order to better understand how the main architectural choices affect stability, accuracy, and scalability in SciML CL.

\section{Conclusion}
% This work investigated CL for SciML using a realistic multi-stage vascular flow dataset, where a model is first trained on a large dataset and later updated with only a few samples from new physical regimes that extrapolate the original large dataset. Unlike standard CL benchmarks with large balanced tasks, this setting reflects real scientific workflows. We benchmarked major CL families (regularization-based, replay-based, optimization-based, and architecture-based methods) and introduced the SLE-FNO approach. Our results confirm that naïve fine-tuning suffers from severe catastrophic forgetting, while regularization-based methods such as EWC provide limited retention under large distribution shifts. Replay-based methods consistently achieve strong stability-plasticity trade-offs and optimization-based methods become restrictive as task constraints accumulate. Architecture-based methods avoid forgetting by design, and SLE-FNO achieved the best overall accuracy with minimal additional cost by adding a single adaptive spectral layer to a frozen FNO backbone. KPCA enabled reliable task-agnostic inference, supporting practical deployment. Together, these results advance continual operator learning for scientific applications and support adaptive digital-twin modeling and long-term scientific forecasting.

\edit{

This work investigated CL for SciML in a realistic multi-stage setting, where models are pretrained on large datasets and adapted with limited samples from new regimes. Within this idealized blood flow CFD benchmark, we show that naive fine-tuning leads to catastrophic forgetting, regularization-based methods provide limited robustness under large shifts, and replay-based methods offer strong performance at the cost of data storage. Architecture-based approaches avoid forgetting by design, and the proposed SLE-FNO achieves the best overall performance with minimal additional parameters. KPCA-based routing further enables effective task-agnostic inference.

The presented results are established on a controlled benchmark with limited tasks and idealized conditions. Extending the approach to more complex patient-specific data, larger task sequences, and real-world deployment, where noise, partial observability, and stronger distribution shifts are present, remains an important direction for future work.
}

\section*{Conflict of Interest}
The authors declare no conflicts.

\section*{Acknowledgments}
This research was funded by the National Science Foundation (NSF) Award No. 2205265.

\appendix
\section{Appendix}

\edit{

\subsection{Hyperparameter Tuning Protocol}
\label{app:hyperparameter_tuning}
To ensure a fair and well-documented comparison across all methods, we adopt a structured multi-stage hyperparameter tuning strategy. We begin with a coarse grid search over a predefined range of key hyperparameters, including the learning rate, batch size, weight decay, and (when applicable) the weights of additional loss terms. This initial search identifies promising regions in the hyperparameter space. Subsequently, we define a refined search region around the best-performing configurations obtained from the coarse stage and perform a finer-grained exploration within this region.

To account for stochasticity and ensure robust model selection, we perform a randomized search within the refined region. Specifically, for each training stage, 32 hyperparameter configurations are sampled from this space and evaluated, and the best-performing setup is selected based on validation performance. Resource usage, including training time and memory footprint, is also monitored during tuning.

}
\subsection{Hyperparameters}
\label{subsec:HT}
\edit{All hyperparameters are presented in Table~\ref{tab:hyperparameters_overhead}.}
% Required packages in preamble:
% \usepackage{booktabs}
% \usepackage{multirow}
% \usepackage{makecell}
% \usepackage{graphicx}
% \usepackage{rotating}
%\clearpage
\begin{sidewaystable}
\centering
\edit{
\caption{\edit{Hyperparameters and computational overhead for each CL algorithm per training stage (B, C, D). ``---'' denotes that the method does not use the corresponding parameter. $\mathrm{lr_{start}}$: initial learning rate; $\mathrm{lr_{end}}$: final learning rate; $\lambda_A$, $\lambda_B$, $\lambda_C$: regularisation penalty weights for tasks A, B, and C respectively; Replay A/B/C: number of stored samples replayed from each prior task. Training time is reported in seconds, peak memory in MB, and inference time (latency) in milliseconds, defined as the average forward-pass time over 20 runs after 3 warm-up iterations on a small batch of test samples. All computational overhead values are reported as mean $\pm$ std. All training and inference experiments were conducted on a single \texttt{NVIDIA RTX A2000} GPU with 12 GB of memory.}}
\label{tab:hyperparameters_overhead}
\renewcommand{\arraystretch}{1.15}
\resizebox{\textwidth}{!}{%
\begin{tabular}{llccccccccccccccc}
\toprule
\multirow{2}{*}{\textbf{Method}} &
\multirow{2}{*}{\textbf{Stage}} &
\multirow{2}{*}{$\mathrm{lr_{start}}$} &
\multirow{2}{*}{$\mathrm{lr_{end}}$} &
\multirow{2}{*}{\textbf{Batch Size}} &
\multirow{2}{*}{\textbf{Weight Decay}} &
\multirow{2}{*}{\textbf{Grad Clip}} &
\multirow{2}{*}{$\lambda_A$} &
\multirow{2}{*}{$\lambda_B$} &
\multirow{2}{*}{$\lambda_C$} &
\multicolumn{3}{c}{\textbf{Replay Count}} &
\multirow{2}{*}{\textbf{Train Time (s)}} &
\multirow{2}{*}{\textbf{Peak Memory (MB)}} &
\multirow{2}{*}{\textbf{Inference (ms)}} \\
\cmidrule(lr){11-13}
& & & & & & & & & & \textbf{A} & \textbf{B} & \textbf{C} & & & \\
\midrule
% Baseline
\multirow{3}{*}{Baseline}
 & B & $2\times10^{-3}$ & $1\times10^{-5}$ & 2 & $1\times10^{-4}$ & --- & --- & --- & --- & --- & --- & --- & $167.99 \pm 4.11$ & $308.91 \pm 33.82$ & $3.50 \pm 0.11$ \\
 & C & $5\times10^{-4}$ & $1\times10^{-5}$ & 4 & $1\times10^{-3}$ & 0.5 & --- & --- & --- & --- & --- & --- & $88.87 \pm 2.89$ & $370.58 \pm 21.95$ & $3.53 \pm 0.16$ \\
 & D & $2\times10^{-3}$ & $1\times10^{-5}$ & 4 & $1\times10^{-3}$ & 0.5 & --- & --- & --- & --- & --- & --- & $88.97 \pm 2.94$ & $374.18 \pm 11.49$ & $3.52 \pm 0.14$ \\
\midrule
% EWC
\multirow{3}{*}{EWC}
 & B & $5\times10^{-3}$ & $1\times10^{-5}$ & 2 & $1\times10^{-2}$ & --- & $2\times10^{7}$ & --- & --- & --- & --- & --- & $224.90 \pm 2.07$ & $355.99 \pm 34.51$ & $3.52 \pm 0.20$ \\
 & C & $8\times10^{-4}$ & $1\times10^{-5}$ & 2 & $1\times10^{-2}$ & --- & $5\times10^{4}$ & $5\times10^{4}$ & --- & --- & --- & --- & $306.95 \pm 2.21$ & $423.13 \pm 22.70$ & $3.48 \pm 0.12$ \\
 & D & $5\times10^{-4}$ & $1\times10^{-5}$ & 2 & $1\times10^{-2}$ & --- & $1\times10^{6}$ & $1\times10^{6}$ & $1\times10^{9}$ & --- & --- & --- & $380.81 \pm 8.42$ & $490.05 \pm 10.92$ & $3.44 \pm 0.05$ \\
\midrule
% LwF
\multirow{3}{*}{LwF}
 & B & $5\times10^{-4}$ & $1\times10^{-5}$ & 2 & --- & --- & 0.30 & --- & --- & --- & --- & --- & $180.74 \pm 4.80$ & $307.63 \pm 33.62$ & $3.49 \pm 0.09$ \\
 & C & $5\times10^{-4}$ & $1\times10^{-5}$ & 2 & --- & --- & 0.40 & 0.45 & --- & --- & --- & --- & $190.67 \pm 5.00$ & $312.65 \pm 22.43$ & $3.49 \pm 0.08$ \\
 & D & $5\times10^{-5}$ & $1\times10^{-5}$ & 2 & --- & --- & 0.15 & 0.15 & 0.20 & --- & --- & --- & $199.56 \pm 4.91$ & $316.31 \pm 10.90$ & $3.51 \pm 0.08$ \\
\midrule
% GEM
\multirow{3}{*}{GEM}
 & B & $5\times10^{-4}$ & $1\times10^{-5}$ & 2 & 0.0 & --- & --- & --- & --- & 16 & --- & --- & $416.92 \pm 2.62$ & $630.46 \pm 34.66$ & $3.49 \pm 0.10$ \\
 & C & $1\times10^{-3}$ & $1\times10^{-5}$ & 2 & $1\times10^{-4}$ & --- & --- & --- & --- & 16 & 1 & --- & $468.37 \pm 9.71$ & $670.73 \pm 23.47$ & $3.46 \pm 0.06$ \\
 & D & $5\times10^{-3}$ & $1\times10^{-5}$ & 1 & $1\times10^{-5}$ & --- & --- & --- & --- & 16 & 1 & 1 & $879.56 \pm 6.82$ & $710.84 \pm 11.91$ & $3.45 \pm 0.05$ \\
\midrule
% LoRA
\multirow{3}{*}{LoRA}
 & B & $1\times10^{-3}$ & --- & 2 & $1\times10^{-5}$ & --- & --- & --- & --- & --- & --- & --- & $100.17 \pm 2.86$ & $175.45 \pm 0.00$ & $4.23 \pm 0.10$ \\
 & C & $1\times10^{-3}$ & --- & 4 & $1\times10^{-4}$ & --- & --- & --- & --- & --- & --- & --- & $64.51 \pm 0.82$ & $269.01 \pm 0.00$ & $4.16 \pm 0.08$ \\
 & D & $5\times10^{-2}$ & --- & 4 & 0.0 & --- & --- & --- & --- & --- & --- & --- & $64.25 \pm 0.75$ & $269.01 \pm 0.00$ & $4.17 \pm 0.08$ \\
\midrule
% OGD
\multirow{3}{*}{OGD}
 & B & $1\times10^{-4}$ & $1\times10^{-5}$ & 2 & $1\times10^{-4}$ & 1.0 & --- & --- & --- & --- & --- & --- & $1407.74 \pm 50.85$ & $352.87 \pm 44.83$ & $3.38 \pm 0.11$ \\
 & C & $1\times10^{-4}$ & $1\times10^{-5}$ & 2 & $1\times10^{-4}$ & 1.0 & --- & --- & --- & --- & --- & --- & $1418.22 \pm 43.53$ & $356.63 \pm 33.10$ & $3.40 \pm 0.09$ \\
 & D & $5\times10^{-4}$ & $1\times10^{-5}$ & 1 & 0.0 & 1.0 & --- & --- & --- & --- & --- & --- & $2884.86 \pm 110.99$ & $335.15 \pm 23.27$ & $3.40 \pm 0.07$ \\
\midrule
% PiggyBack
\multirow{3}{*}{PiggyBack}
 & B & $1\times10^{-2}$ & $1\times10^{-4}$ & 2 & --- & --- & --- & --- & --- & --- & --- & --- & $162.76 \pm 0.64$ & $202.34 \pm 12.18$ & $3.86 \pm 0.14$ \\
 & C & $1\times10^{-2}$ & $1\times10^{-3}$ & 1 & --- & --- & --- & --- & --- & --- & --- & --- & $322.76 \pm 0.68$ & $172.46 \pm 6.04$ & $4.00 \pm 0.10$ \\
 & D & $1\times10^{-2}$ & $1\times10^{-4}$ & 1 & --- & --- & --- & --- & --- & --- & --- & --- & $322.50 \pm 1.08$ & $174.42 \pm 0.18$ & $3.96 \pm 0.03$ \\
\midrule
% Replay K-means
\multirow{3}{*}{\makecell[l]{Replay\\K-means}}
 & B & $1\times10^{-3}$ & $1\times10^{-5}$ & 1 & $1\times10^{-2}$ & 1.0 & --- & --- & --- & 16 & --- & --- & $1070.95 \pm 18.71$ & $278.47 \pm 34.21$ & $3.70 \pm 0.08$ \\
 & C & $1\times10^{-3}$ & $1\times10^{-5}$ & 1 & $1\times10^{-2}$ & 1.0 & --- & --- & --- & 16 & 1 & --- & $1115.35 \pm 17.44$ & $282.56 \pm 23.23$ & $3.71 \pm 0.02$ \\
 & D & $1\times10^{-3}$ & $1\times10^{-5}$ & 1 & $1\times10^{-2}$ & 1.0 & --- & --- & --- & 16 & 1 & 1 & $1159.54 \pm 21.17$ & $286.63 \pm 11.30$ & $3.67 \pm 0.13$ \\
\midrule
% Replay Random
\multirow{3}{*}{\makecell[l]{Replay\\Reservoir}}
 & B & $1\times10^{-3}$ & $1\times10^{-5}$ & 1 & $1\times10^{-2}$ & 1.0 & --- & --- & --- & 16 & --- & --- & $1024.69 \pm 51.87$ & $278.47 \pm 34.21$ & $3.61 \pm 0.17$ \\
 & C & $1\times10^{-3}$ & $1\times10^{-5}$ & 1 & $1\times10^{-2}$ & 1.0 & --- & --- & --- & 16 & 1 & --- & $1068.58 \pm 61.82$ & $282.56 \pm 23.23$ & $3.56 \pm 0.16$ \\
 & D & $1\times10^{-3}$ & $1\times10^{-5}$ & 1 & $1\times10^{-2}$ & 1.0 & --- & --- & --- & 16 & 1 & 1 & $1120.07 \pm 51.98$ & $286.63 \pm 11.30$ & $3.65 \pm 0.13$ \\
\midrule
% SLE-FNO
\multirow{3}{*}{SLE-FNO}
 & B & $5\times10^{-3}$ & $1\times10^{-5}$ & 4 & $1\times10^{-6}$ & 2.0 & --- & --- & --- & --- & --- & --- & $29.37 \pm 0.63$ & $174.30 \pm 11.56$ & $3.62 \pm 0.31$ \\
 & C & $5\times10^{-4}$ & $1\times10^{-5}$ & 1 & $1\times10^{-7}$ & 2.0 & --- & --- & --- & --- & --- & --- & $140.51 \pm 1.14$ & $152.39 \pm 6.23$ & $3.65 \pm 0.02$ \\
 & D & $5\times10^{-3}$ & $2\times10^{-5}$ & 1 & $1\times10^{-6}$ & 0.5 & --- & --- & --- & --- & --- & --- & $139.92 \pm 3.77$ & $155.00 \pm 0.00$ & $3.69 \pm 0.04$ \\
\bottomrule
\end{tabular}%
}
}
\end{sidewaystable}

\subsection{Number of KPCA Modes}
\label{subsec:KPCAmodes}

\begin{figure}[h!]
    \centering
    \includegraphics[width=0.5\textwidth]{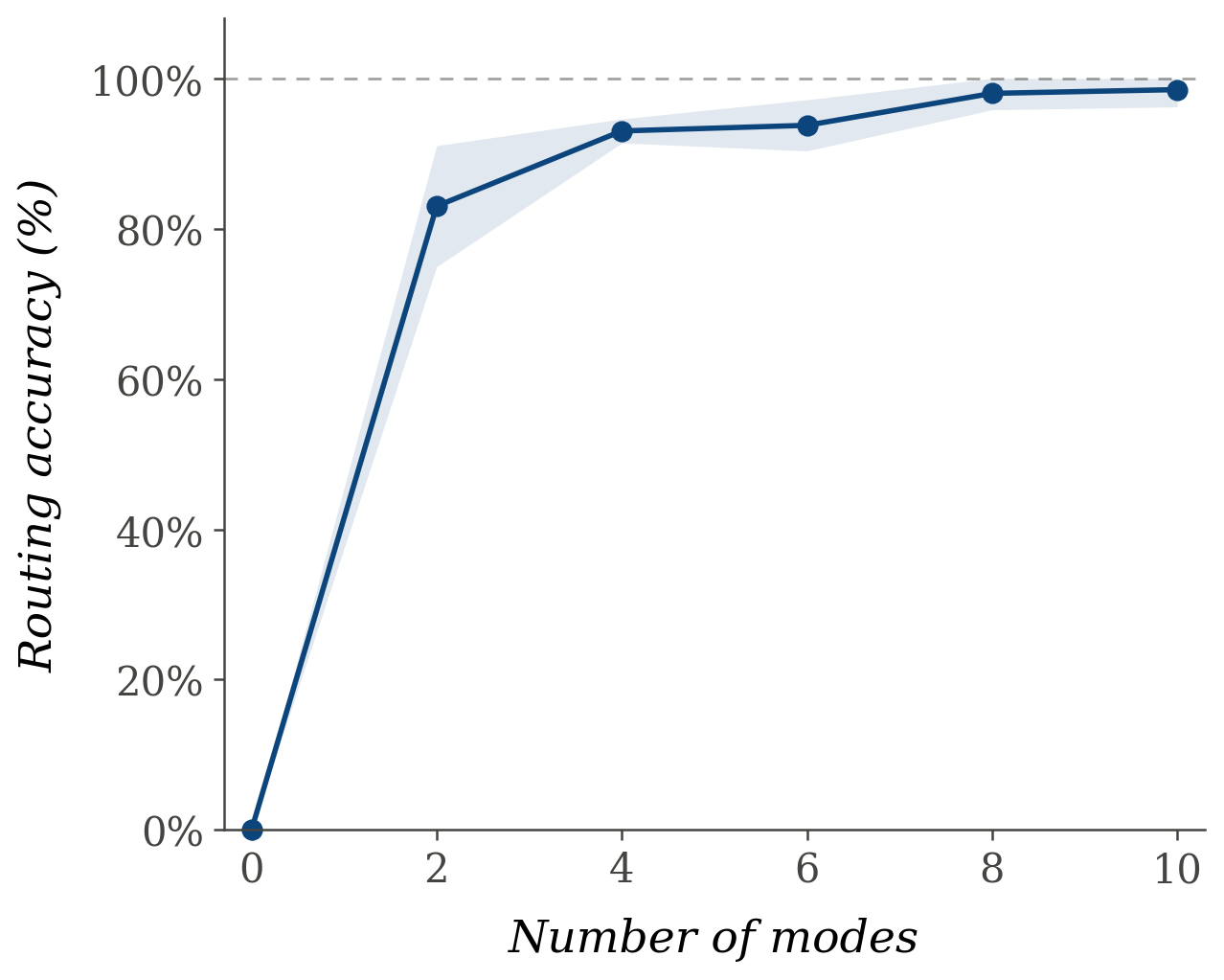}
    \caption{\edit{ Task A routing accuracy (\%) shown as the mean (solid line) with a shaded region indicating $\pm 1$ standard deviation, plotted as a function of the number of retained KPCA modes. The x-axis represents the number of retained modes, while the y-axis reports the routing accuracy on the Task A test set.}}
    \label{fig:KPCAModes}
\end{figure}

\edit{
In this part, we examine the classification error based on the number of modes kept. We perform inference on all 4 KPCAs (A,B,C and D) and assign the task to the class that is giving the smallest reconstruction error.}

We examine the number of KPCA modes required for accurate task identification. Since RFF are used to lift the data into a nonlinear feature space, a fixed number of nonlinear features must be chosen and shared across all tasks. This constraint is necessary to ensure a fair comparison of reconstruction and \edit{routing} performance across tasks. Accordingly, we fix the number of RFF features to 4096 for all experiments.

Under this setting, we observe that tasks~B, C, and~D achieve \edit{the highest routing} accuracy on the test set using \edit{4} KPCA mode each. In contrast, task~A requires \edit{10} modes to reach comparable performance. This behavior is expected, as task~A exhibits significantly higher variability due to both geometric diversity and the presence of multiple distinct inlet profiles. Figure~\ref{fig:KPCAModes} illustrates the relationship between the number of retained KPCA modes and the resulting \edit{routing} accuracy for task~A.

\edit{
\subsection{KPCA-Based Classification: Cross-Validation Results}
\label{subsec:kpca_cv_results}
Building on the same assumption as previous subsection, the KPCA classifier is evaluated using a cross-validation protocol consisting of 5 splits and 2 random seeds used to generate RFF resulting in a total of 10 runs. Across all runs, the classifier achieves an overall mean accuracy of 99.1\% with 456 correctly classified samples out of 460 (the misclassified cases belong to Task D as discussed later). The observed performance ranges from 97.8\% to 100.0\%, demonstrating highly consistent classification behavior across different splits and seeds. These results confirm that the classifier is able to reliably route samples across all groups under varying data partitions. The implications of incorrect routing, i.e., misclassify a sample to the wrong group, are discussed next in Section~\ref{Without OOD}.
}

\edit{

\subsection{Performance Without Routing Mechanism}
\label{Without OOD}
To further analyze the role of the routing mechanism, we evaluate the model without using the distribution detector at inference time and without knowing the input task class. In this setting, we consider two alternatives. First, each test sample is passed through all branches separately, including both cross-branch and task-specific evaluations. Second, we use an ensemble strategy, where predictions from all branches are averaged across tasks without any routing.

The results in Table~\ref{tab:ood_no_detector_transformed} show a clear degradation in performance when routing is removed. While the task-specific SLE branch achieves the best performance for each task, applying mismatched branches leads to significant increases in error. The ensemble approach partially stabilizes the predictions through averaging, but still suffers from interference between branches. These results highlight the importance of an accurate routing mechanism to effectively utilize task-specific adaptations.

\begin{table}[h!]
\centering
\caption{\edit{Performance accuracy without distribution detector. Each entry is reported as mean $\pm$ std (higher is better).}}
\label{tab:ood_no_detector_transformed}

\edit{
\begin{tabular}{llc}
\toprule
Branch & Score $\uparrow$ \\
\midrule

\multicolumn{2}{c}{\textbf{Task A}} \\
Backbone  & \textbf{0.985} \\
SLE-B & 0.637 $\pm$ 0.011 \\
SLE-C & 0.464 $\pm$ 0.033 \\
SLE-D & 0.551 $\pm$ 0.017 \\
Ensemble   & 0.665 $\pm$ 0.006 \\

\midrule
\multicolumn{2}{c}{\textbf{Task B}} \\
Backbone  & 0.522 \\
SLE-B & \textbf{0.898 $\pm$ 0.024} \\
SLE-C & 0.494 $\pm$ 0.093 \\
SLE-D & 0.299 $\pm$ 0.053 \\
Ensemble   & 0.654 $\pm$ 0.030 \\

\midrule
\multicolumn{2}{c}{\textbf{Task C}} \\
Backbone  & 0.304 \\
SLE-B & 0.444 $\pm$ 0.088 \\
SLE-C & \textbf{0.789 $\pm$ 0.050} \\
SLE-D & 0.169 $\pm$ 0.106 \\
Ensemble   & 0.452 $\pm$ 0.111 \\

\midrule
\multicolumn{2}{c}{\textbf{Task D}} \\
Backbone  & 0.513 \\
SLE-B & 0.109 $\pm$ 0.051 \\
SLE-C & 0.014 $\pm$ 0.015 \\
SLE-D & \textbf{0.830 $\pm$ 0.022} \\
Ensemble   & 0.240 $\pm$ 0.071 \\

\bottomrule
\end{tabular}
}

\end{table}

}

\edit{
\subsection{End-to-End Inference}
\label{subsec:endtoend}

We evaluate the full pipeline with the KPCA selector active during
inference. At each CL stage, incoming test samples are
first routed by the KPCA classifier to the most appropriate corrector,
which then produces the final prediction.
The routing pool grows sequentially: at stage~B the selector chooses
between correctors~$\{A, B\}$; at stage~C among $\{A, B, C\}$; and at
stage~D among all four $\{A, B, C, D\}$.
For each group, we report the accuracy and the average accuracy across all groups available at that stage, as mean~$\pm$~std over the 10 seed/split runs.

As shown in Table~\ref{tab:kpca_routed}, the KPCA selector achieves
perfect routing accuracy on groups~A, B, and~C across all stages.
Group~D reaches $80\%$ routing accuracy at stage~D, where all four
misrouted samples (across all 10 runs) are assigned to group~A.
This behaviour is attributed to the size of group~A being the largest dataset in the CL sequence, its KPCA model spans a broader region of the feature space, making it more likely to absorb samples from later, smaller groups whose feature distributions partially overlap with it. In other words, the larger the training set used to fit a KPCA model, the higher the probability that its reconstruction subspace resembles out-of-distribution samples from subsequent tasks, thereby confusing
the selector. When such misrouting occurs, the inference is forwarded along the main backbone without any task-specific correction, since group~A corresponds to the pretrained model operating without a corrector head.
This is consistent with the analysis in Section~\ref{Without OOD},
which shows that routing failures degrade performance reasonably rather than catastrophically, as the backbone alone still provides a reasonable prediction.

\begin{table}[h!]
\edit{
\centering
\caption{\edit{End-to-end inference accuracy (mean $\pm$ std over 10 runs) with the KPCA selector active during inference. Routing accuracy per group across all runs is shown in parentheses. A dash indicates the corrector is not yet available at that stage.}}
\label{tab:kpca_routed}
\resizebox{\columnwidth}{!}{%
\begin{tabular}{llccccc}
\hline
\textbf{Stage} & \textbf{Pool} & \textbf{A} & \textbf{B} & \textbf{C} & \textbf{D} & \textbf{Avg.\ Acc.} \\
\hline
B & $\{A,B\}$     & $0.985\pm0.000$ {\tiny(100\%)} & $0.898\pm0.033$ {\tiny(100\%)} & ---                            & ---                           & 0.942 \\
C & $\{A,B,C\}$   & $0.985\pm0.000$ {\tiny(100\%)} & $0.898\pm0.033$ {\tiny(100\%)} & $0.789\pm0.050$ {\tiny(100\%)} & ---                           & 0.891 \\
D & $\{A,B,C,D\}$ & $0.985\pm0.000$ {\tiny(100\%)} & $0.898\pm0.033$ {\tiny(100\%)} & $0.789\pm0.050$ {\tiny(100\%)} & $0.778\pm0.054$ {\tiny(80\%)} & 0.863 \\
\hline
\end{tabular}}

}
\end{table}

}

\edit{

\subsection{Ablation Study on SLE Layer Size}
\label{app:sleablation}

To evaluate the impact of the SLE layer capacity, we perform an ablation study across three tiers (Small, Medium, Large), each defined by different hidden channel sizes and Fourier modes. Specifically, the Small configuration uses $h_c=6$ with modes $(4,4)$ (2,032 parameters), the Medium configuration uses $h_c=12$ with modes $(8,8)$ (20,200 parameters), and the Large configuration uses $h_c=24$ with modes $(32,32)$ (1,183,408 parameters). These correspond to approximately 0.08\%, 0.84\%, and 49.1\% of the backbone model (2.41M parameters), respectively. Notably, the Large configuration accounts for nearly half of the backbone size, making it less practical from a parameter-efficiency perspective, but still useful for assessing the upper bound of achievable performance.

The results in Table~\ref{tab:sle_ablation_accuracy} show that increasing the SLE capacity generally improves performance, particularly for more complex tasks such as C and D. The Medium configuration achieves strong and stable performance, while the Large configuration often yields the best accuracy, indicating that higher-capacity SLE layers can better capture task-specific corrections. However, this improvement comes at the cost of significantly increased model size, highlighting a clear trade-off between performance and efficiency.

\begin{table}[h!]
\centering
\caption{\edit{Ablation study on SLE layer size. Performance accuracy is reported at the final step using transformed accuracy (higher is better).}}
\label{tab:sle_ablation_accuracy}

\edit{
\begin{tabular}{lccc}
\toprule
Tier & B & C & D \\
\midrule

Small 
& 0.875 $\pm$ 0.020 
& 0.652 $\pm$ 0.096 
& 0.759 $\pm$ 0.056 \\

Medium 
& 0.887 $\pm$ 0.015 
& \textbf{0.766 $\pm$ 0.043} 
& 0.801 $\pm$ 0.021 \\

Large 
& \textbf{0.905 $\pm$ 0.021} 
& 0.748 $\pm$ 0.066 
& \textbf{0.841 $\pm$ 0.034} \\

\bottomrule
\end{tabular}
}

\end{table}

}

\edit{
\subsection{Continual Learning (CL) Techniques}
\label{app:cl}

More detailed explanations about each benchmark CL technique are presented here. 

\subsubsection{Elastic Weight Consolidation (EWC)}
EWC~\cite{kirkpatrick2017overcoming} is motivated by the biological idea of synaptic consolidation~\cite{ziegler2015synaptic}, where the brain protects previously learned knowledge by reducing plasticity in important neural connections. In CL, EWC follows a similar principle where the identified important parameters for previous tasks should not change too much when learning a new task Fig.~\ref{fig:continual_learning_algorithms}a.

To estimate how important each parameter is, EWC uses the Fisher Information Matrix (FIM), which measures how sensitive the model's output distribution is to changes in parameters, and is formally defined as:

\begin{equation}
\mathbf{F} = \mathbb{E}_{(x,y)\sim p(x,y)}\left[\nabla_{\theta}\log p(y\mid x, \theta)\,\nabla_{\theta}\log p(y\mid x, \theta)^{\top}\right] \;.
\end{equation}
In this equation, $\mathbf{F}$ denotes the FIM with respect to the model parameters $\theta$. The expectation $\mathbb{E}_{(x,y)\sim p(x,y)}[\cdot]$ is taken over the joint data distribution $p(x,y)$ of inputs $x$ and outputs $y$. The term $p(y \mid x, \theta)$ represents the conditional likelihood of observing output $y$ given input $x$ under the parameterized model. The gradient $\nabla_{\theta} \log p(y \mid x, \theta)$ is the score function, i.e., the derivative of the log-likelihood with respect to the parameters $\theta$. The outer product $\nabla_{\theta} \log p(y \mid x, \theta)\,\nabla_{\theta} \log p(y \mid x, \theta)^{\top}$ captures the second-order sensitivity of the log-likelihood, and its expectation defines the curvature of the parameter space encoded by the FIM.

Computing previously mentioned theoretical (expected) FIM is generally impractical as it requires expectation over the true data distribution, which is unknown in practice, and the matrix scales with the number of model parameters~\cite{fujita2022fisher,van2025computation,spall2003monte}. To address this, the metric is approximated empirically using Monte Carlo sampling over available data:

\begin{equation}
\mathbf{F}^{(t)} \;\approx\; \frac{1}{n_t} \sum_{j=1}^{n_t}
\left(\nabla_{\theta}\log p(y_j^{(t)} \mid x_j^{(t)},\theta)\right)
\left(\nabla_{\theta}\log p(y_j^{(t)} \mid x_j^{(t)},\theta)\right)^{\top} \;,
\end{equation}
where $\mathbf{F}^{(t)}$ denotes the empirical FIM for task $t$. The approximation replaces the expectation over the data distribution with a finite-sample average over the $n_t$ training examples of task $t$.

Even with this approximation, storing the full FIM remains expensive because it captures correlations between every pair of parameters. To make the method scalable, EWC assumes that parameters are locally independent, which allows using only the diagonal entries:

\begin{equation}
F_i^{(t)} = \left[\mathbf{F}^{(t)}\right]_{ii} \;.
\end{equation}

This diagonal Fisher approximation is widely used not only in EWC but also across other CL methods. Prior work has compared exact, sampling-based, and empirical Fisher estimates, analyzing trade-offs between computational cost and fidelity~\cite{van2025computation,soen2024trade,spall2003monte}.

Once the Fisher values are computed, EWC adds a quadratic penalty to the loss of the new task to prevent large deviations from previously learned parameter values. For a previous task's parameters \(\theta^*\) and Fisher entries \(\{F_i\}\), the single-task EWC objective is:

\begin{equation}
\mathcal{L}_{\mathrm{EWC}}(\theta)
=
\mathcal{L}_{\text{new}}(\theta)
+
\frac{\lambda}{2}\sum_{i} F_i\big(\theta_i - \theta_i^{*}\big)^2 \;,
\end{equation}
where $\mathcal{L}_{\text{EWC}}(\theta)$ denotes the total loss used for training on a new task under EWC, $\mathcal{L}_{\text{new}}(\theta)$ is the standard task-specific loss for the current task, and the scalar $\lambda$ is a regularization coefficient that controls the trade-off between learning the new task and preserving previously learned knowledge. The summation index $i$ runs over all model parameters, where $\theta_i$ denotes the $i$-th parameter and $\theta_i^{*}$ is its value after training on previous tasks. The quantity $F_i$ represents the diagonal element of the FIM associated with parameter $\theta_i$, measuring its importance to earlier tasks. The quadratic penalty $(\theta_i - \theta_i^{*})^{2}$ discourages deviations of important parameters, thereby mitigating catastrophic forgetting.

Extensions of EWC accumulate penalties across multiple previously learned tasks:

\begin{equation}
\mathcal{L}_{\mathrm{EWC}}(\theta)
= \mathcal{L}_k(\theta)
+ \frac{\lambda}{2}\sum_{t=1}^{k-1}\sum_i F^{(t)}_i\,\big(\theta_i - \theta_i^{*(t)}\big)^2 \;,
\end{equation}
where \( \mathcal{L}_k \) is the loss for the current task \(k\) and \( F_i^{(t)} \) is the diagonal Fisher entry for parameter \(i\) from task \(t\)~\cite{aich2021elastic}.

In practice, computing and storing one Fisher value per task introduces overhead, and large values of \(F_i^{(t)}\) or \(\lambda\) can overly restrict parameter updates, making learning on later tasks slower or more difficult. For example, when training task \(k=4\), the regularization term sums over the Fisher matrices from previous three tasks \(t = 1, 2, 3\).

\subsubsection{Learning without Forgetting (LwF)}
The inspiration behind LwF~\cite{li2017learning} (Fig.~\ref{fig:continual_learning_algorithms}b) is knowledge distillation~\cite{gou2021knowledge} with the help of an auxiliary loss to guide model training by encouraging the current model to align its predictions with those of another model. Originally proposed in~\cite{hinton2015distilling}, the framework transfers knowledge from a large and expressive teacher model to a smaller and more efficient student model. While the original method focused on classification, the same concept extends naturally to regression tasks, where the goal is to match continuous-valued outputs rather than probability scores~\cite{kang2021data,saputra2019distilling}.

In regression-based distillation, the learning objective combines the standard supervised loss with a distillation loss that minimizes the discrepancy between the teacher and student outputs. A general formulation is:

\begin{equation}
\mathcal{L}_{KD} = \mathcal{L}_{task}(y,\, f_s(x)) + \lambda\, \mathcal{L}_{distill}\big(f_t(x),\, f_s(x)\big) \;,
\end{equation}
where \(f_t(x)\) and \(f_s(x)\) are the teacher and student predictions, \(\mathcal{L}_{task}\) is the supervised task loss (e.g., mean squared error), \(\mathcal{L}_{distill}\) measures the discrepancy between teacher and student, and $\lambda$ controls the balance between learning from the ground-truth data and imitating the teacher. 

%when tasks are far apart, it cannot optimize both simultaneously because the regularizer pulls new predictions toward the old model

Through this auxiliary alignment, the student benefits from the teacher’s potentially more informative predictions. In CL context, the teacher may be a stored snapshot of the model from a previous task, enabling distillation without requiring a separate pretrained network. LwF uses knowledge distillation by freezing the previous model as a (teacher) and training the current model (student) to match its predictions while learning the new task, typically trading off plasticity and retention. The LwF loss is defined as~\cite{li2017learning,hinton2015distilling}

\begin{equation}
\mathcal{L}_{\mathrm{LwF}}
=\mathcal{L}_{\text{task}}(\theta;\mathcal{D}_k)
+\frac{\lambda}{|\mathcal{D}_k|}\sum_{x\in\mathcal{D}_k}\big\|f_{\theta}(x)-f_{\theta^{*}}(x)\big\|_2^{2} \;,
\end{equation}
where $\theta^{*}$ is the frozen teacher (after task $k-1$), $\theta$ is the student (new model) on task $k$, and $\mathcal{D}_k$ is the current-task data.

\subsubsection{Replay-Based}
Replay-based CL stores a small number of samples from previous tasks in a replay buffer so the model can reuse them during later training. These stored samples act as a lightweight memory of earlier data and help reduce forgetting when the model learns new tasks~\cite{rolnick2019experience,bagus2021investigation}.

Instead of having access to the full data from previous tasks, the model keeps a compact buffer
\begin{equation}
\mathcal{B} = \{(x_j, y_j)\}_{j=1}^{M} \;,
\end{equation}
where $\mathcal{B}$ denotes the replay buffer, $(x_j, y_j)$ are the $j$-th stored input–output pair, $x_j$ represents an input sample, $y_j$ its corresponding target, $j$ is the sample index, and $M$ is the fixed buffer capacity. During training on a new task samples from the buffer are mixed with the new task data so that the loss contains information from both old and new examples.

By replaying stored samples during training, the model is encouraged to retain knowledge from earlier tasks while still being able to learn new information. Replay-based CL also introduces several challenges. The first challenge is how to select samples that represent the past data distribution most efficiently. Since ideally only a small number of samples can be stored, the buffer must capture the structure of the past data with limited memory~\cite{chaudhry2019tiny}.

Another challenge is data size. In many real-world applications, data may be too large to store or transfer, and this becomes a practical obstacle. Privacy is an additional limitation, because in some domains (for example medical or personal data), storing or reusing old samples may not be possible~\cite{bagus2021investigation}. To avoid storing real samples, some work proposes using a generative model instead, which is trained on previous task data and later used to generate synthetic samples when needed \cite{shin2017continual,wu2018memory}. 
%In this case, the replay buffer becomes:
%\begin{equation}
%\tilde{\mathcal{B}} = \{G(z_j)\}_{j=1}^{M},
%\end{equation}
%where \(G(\cdot)\) is the generative model and \(z_j\) is a noise input sampled from a %prior distribution.
Finally, the replay buffer can be fixed in size. If the number of tasks grows and the memory stays constant, then the number of samples stored per task decreases over time. 

% For example, if allocating memory equally per task:
% \begin{equation}
% M_t = \frac{M}{k},
% \end{equation}
% where \(k\) is the number of tasks seen so far, then replay becomes weaker as more tasks accumulate. This creates a trade-off between storage limits and the ability to protect knowledge from earlier tasks.

To select the optimal samples to store in the replay buffer, different strategies have been proposed. A simple and widely used option is random selection (or reservoir sampling), where each incoming sample has the same probability of being stored. This strategy has been shown to be effective in some examples~\cite{chaudhry2019tiny,rolnick2019experience}. Other methods try to select more representative samples. One common idea is to fit a clustering model in a feature space and choose samples that are close to cluster centers \cite{bagus2021investigation}. 
%Let \(\phi(x)\) be the feature vector of input \(x\), and let \(\{c_k\}_{k=1}^K\) be cluster centers. The buffer \(\mathcal{B}\) is then chosen to keep samples that stay close to these centers, for example
%\begin{equation}
%\mathcal{B} \approx \arg\min_{\mathcal{B}:\,|\mathcal{B}|\le M}
%\sum_{k=1}^K \min_{x \in \mathcal{B}} \|\phi(x) - c_k\|_2^2.
%\end{equation}
More recent work uses diversity-based or optimization-based selection, where the goal is to keep samples that cover different regions of the input space or produce diverse gradients during training \cite{aljundi2019gradient,tiwari2022gcr}. 
In our work we adapted the following two algorithms to choose old samples.
\paragraph{Random (Reservoir) Replay.}
In random replay, also known as reservoir replay, samples are selected uniformly at random from the incoming data stream and stored in a fixed-size replay buffer. Given a data stream $\{\mathbf{x}_t\}_{t=1}^{\infty}$ and a replay buffer of capacity $B$,
each sample $\mathbf{x}_t$ is retained in the buffer with probability
\begin{equation}
p_t = \min\!\left(1,\frac{B}{t}\right) \;,
\end{equation}
where $t$ denotes the index of the current sample in the stream.
If the buffer is already full, an existing sample is replaced uniformly at random.
This strategy ensures that, at any time, the buffer contains an unbiased random subset
of all previously observed samples, without requiring task boundaries or feature-based selection.
Due to its simplicity and low computational overhead, reservoir replay is widely used as a baseline
in CL~\cite{vitter1985random,chaudhry2019tiny}.

\paragraph{K-means Replay.}
In k-means replay, samples are selected based on their representativeness with respect to the input data distribution.
A k-means clustering model is fitted directly on the input samples
$\{\mathbf{x}_n\}_{n=1}^{N}$, where $\mathbf{x}_n \in \mathbb{R}^d$ denotes the input vector of sample $n$
(e.g., the vectorized input field).
The cluster centers $\{\boldsymbol{\mu}_k\}_{k=1}^{K}$ are obtained by minimizing the within-cluster
sum of squared distances
\begin{equation}
\sum_{n=1}^{N} \min_{k \in \{1,\dots,K\}}
\left\| \mathbf{x}_n - \boldsymbol{\mu}_k \right\|_2^2  \;,
\end{equation}
where $K$ is the number of clusters and $\boldsymbol{\mu}_k \in \mathbb{R}^d$ denotes the centroid of cluster $k$.
The replay buffer is then populated with samples whose input representations are closest to the cluster centers,
i.e., samples minimizing $\|\mathbf{x}_n - \boldsymbol{\mu}_k\|_2$ for each cluster $k$.
By selecting prototypical and diverse inputs that capture the structure of the input space,
this approach aims to reduce redundancy in the replay buffer and improve memory efficiency compared to random selection.
K-means based replay has been adapted in several CL works to improve replay efficiency~\cite{rebuffi2017icarl,chaudhry2019tiny}.

For task A, we stored only 10\% of the available samples, which corresponds to 16 samples. For tasks B and C, we stored one sample per task (out of the 8 training samples). An overview of the Replay-based methods used is shown in Fig.~\ref{fig:continual_learning_algorithms}c-d.

\subsubsection{Orthogonal Gradient Descent (OGD)}

OGD is a CL method that constrains parameter updates so that learning a new task does not interfere with directions important to previous tasks~\cite{farajtabar2020ogd}. In contrast to optimization-based methods that depend on stored past samples \cite{lopez2017gradient,chaudhry2019tiny}, OGD avoids data storage and uses gradient constraints to protect previously acquired knowledge.

The main principle of OGD is to ensure that the gradient of a new task remains compatible with gradients from older tasks. This can be written as:

\begin{equation}
\min_{\mathbf{g}_k} \mathcal{L}_k
\quad \text{s.t.} \quad 
\langle \mathbf{g}_k, \mathbf{g}_t \rangle \ge 0 \quad \forall t<k \;,
\end{equation}
where \(\mathbf{g}_k\) is the current gradient and \(\mathbf{g}_t\) are stored gradients from earlier tasks. For a network with parameters \(\theta = [w_1,\dots,w_D]\), a task gradient has the form:

\begin{equation}
\mathbf{g}_t =
\begin{bmatrix}
\dfrac{\partial \mathcal{L}_t}{\partial w_1} \\
\vdots \\
\dfrac{\partial \mathcal{L}_t}{\partial w_D}
\end{bmatrix}
\in \mathbb{R}^D \;.
\end{equation}

OGD projects the current gradient onto the orthogonal complement of all stored gradients:

\begin{equation}
\mathbf{g}_k^{\perp} =
\mathbf{g}_k - 
\sum_{t=1}^{k-1}
\frac{\langle \mathbf{g}_k, \mathbf{g}_t \rangle}{\|\mathbf{g}_t\|_2^2} \, \mathbf{g}_t \;.
\end{equation}
In this equation, $\mathbf{g}_k$ denotes the gradient of the loss with respect to the model parameters $\theta$ for the current task indexed by $k$. The index $k$ refers to the current task in a sequential learning setting, while $t \in \{1,\dots,k-1\}$ indexes previously learned tasks. The vector $\mathbf{g}_k^{\perp}$ represents the component of the current gradient that is orthogonal to the subspace spanned by the gradients of past tasks. The inner product $\langle \mathbf{g}_k, \mathbf{g}_t \rangle$ measures the alignment between the current-task gradient and the gradient associated with task $t$, and $\|\mathbf{g}_t\|_2^2$ denotes the squared $\ell_{2}$-norm of the past-task gradient. Subtracting the projection of $\mathbf{g}_k$ onto each $\mathbf{g}_t$ removes directions that would interfere with previously learned tasks, yielding an update direction that mitigates catastrophic forgetting. Then we perform the update as:

\begin{equation}
\theta \leftarrow \theta - \eta\, \mathbf{g}_k^{\perp} \;,
\end{equation}
where The scalar $\eta > 0$ is the learning rate. The previous equation guarantees that $ \langle \mathbf{g}_k^{\perp}, \mathbf{g}_t\rangle = 0$ for  $\forall t<k $.

A direct implementation of OGD requires storing all past gradient vectors, resulting in \(O(kD)\) memory, where k is the number of previous tasks and D is number of parameters in the model. This becomes prohibitive for large models \cite{de2021continual}. In our experiments, storing full gradients and performing exact projection in \(\mathbb{R}^D\) led to GPU out-of-memory failures and caused overly stiff optimization, similar to observations in projection-heavy CL methods~\cite{chaudhry2020continual,dong2022gdod}. To make OGD practical, we adapt a more scalable approach inspired by basis construction and orthogonal subspace regularization~\cite{chaudhry2020continual}. Rather than storing full gradients, we apply OGD per layer. For each layer \(\ell\), we maintain an orthonormal basis of past gradients:

\begin{equation}
\mathcal{B}_\ell = \{b_{\ell,1}, \dots, b_{\ell,r_\ell}\} \;,
\quad \|b_{\ell,i}\|_2 = 1 \;,
\end{equation}
where $\mathcal{B}_\ell$ denotes an orthonormal basis of past gradient directions for layer $\ell$, $b_{\ell,i}$ are unit-norm basis vectors, and $r_\ell$ is the number of retained directions defining the gradient subspace at that layer. The basis vectors are computed via incremental Gram--Schmidt orthogonalization, similar to~\cite{tang2021gradient}. During training, the layerwise gradient is modified as:

\begin{equation}
g_\ell^{\perp}
= g_\ell - \alpha \sum_{i=1}^{r_\ell} \langle g_\ell, b_{\ell,i} \rangle b_{\ell,i} \;,
\end{equation}
where \(b_{\ell,i}\) are basis vectors spanning the subspace of gradients associated with previously learned tasks and \(\alpha \in [0,1]\) controls the strength of the orthogonalization. 
When \(\alpha = 1\), the update fully removes the components of \(g_\ell\) that interfere with past tasks, recovering standard OGD.
Smaller values of \(\alpha\) only partially remove these components, allowing controlled interference that improves stability–plasticity trade-offs, as explored in relaxed gradient-projection methods~\cite{chaudhry2018riemannian}.

As tasks accumulate, the size of each basis may still grow. To control memory and projection cost, we compress each basis using singular value decomposition (SVD) and retain only dominant directions explaining at least \(95\%\) of the energy, capped at 32 vectors per layer. This approach follows low-rank knowledge preservation strategies explored in~\cite{chaudhry2020continual,wright2023sketchogd}. The compressed basis provides a good trade-off between memory, computation, and learning flexibility.

In summary, OGD enforces gradient orthogonality to preserve prior knowledge. Our implementation uses layerwise projection, orthonormal basis storage, and SVD compression to avoid memory blowup and excessive constraint stiffness, making OGD practical and stable for CL in large architectures such as FNO (see Fig.~\ref{fig:continual_learning_algorithms}e for an overview).

\subsubsection{Gradient Episodic Memory (GEM)}
GEM is a hybrid CL method that combines replay with gradient-level protection against forgetting~\cite{lopez2017gradient}. Similar to memory-based replay methods, GEM stores a small buffer of samples from earlier tasks instead of keeping the full dataset~\cite{nesterov2022gradient,guo2020improved}. During training on a new task \(k\), the model computes two gradients; one from the current task batch and another from stored memory samples representing past tasks:

\begin{equation}
g_k = \nabla_\theta \mathcal{L}_k(\theta) \;,
\qquad
g_t = \nabla_\theta \mathcal{L}_t(\theta) \;, \; t < k \;.
\end{equation}
Here, $g_k$ denotes the gradient of the loss $\mathcal{L}_k(\theta)$ with respect to the model parameters for the current task $k$, while $g_t$ represents the corresponding gradient for a previous task $t<k$, both evaluated at the current parameter values $\theta$.

To avoid increasing the loss of past tasks, GEM requires the final update direction \(\tilde{g}\) to satisfy:

\begin{equation}
\langle \tilde{g}, g_t \rangle \ge 0 \quad \forall \, t < k \;.
\end{equation}
This condition enforces that the update direction $\tilde{g}$ has a non-negative inner product with the gradient $g_t$ of every previous task $t<k$, ensuring that the parameter update does not increase the loss of earlier tasks.

If the current gradient already satisfies these constraints, the model updates normally (\(\tilde{g} = g_k\)). Otherwise, GEM projects the gradient by solving:

\begin{equation}
\tilde{g} = \arg\min_{g} \frac{1}{2} \|g - g_k\|^2 
\quad \text{s.t.} \quad 
\langle g, g_t \rangle \ge 0 \;,
\end{equation}
which prevents interference while allowing learning of the new task. 
% This makes GEM different from loss-regularization approaches such as EWC \cite{kirkpatrick2017overcoming}, because the protection happens directly at the optimizer level.

Several extensions have improved GEM's efficiency and flexibility. A-GEM~\cite{chaudhry2018efficient} replaces multiple per-task constraints with a single one computed from a random memory batch. Instead of checking all prior \(g_t\), A-GEM uses a single reference gradient:

\begin{equation}
g_{\text{ref}} = \nabla_\theta \mathcal{L}(\theta;\mathcal{M}_{<k}) \;,
\end{equation}
where $g_{\text{ref}}$ denotes the reference gradient computed on a randomly sampled mini-batch from the episodic memory $\mathcal{M}_{<k}$ containing examples from tasks learned prior to task $k$. $\mathcal{L}(\theta;\mathcal{M}_{<k})$ is the loss evaluated on this memory batch and $\nabla_\theta$ denotes the gradient with respect to the model parameters $\theta$, enforcing:

\begin{equation}
\langle \tilde{g}, g_{\text{ref}} \rangle \ge 0 \;.
\label{eq:GEM_Constraint}
\end{equation}

If Eq.~\eqref{eq:GEM_Constraint} is violated, the closed-form projection is used to get the projected gradients, as follows:

\begin{equation}
\tilde{g} =
g_k - \frac{\langle g_k, g_{\text{ref}} \rangle}{\|g_{\text{ref}}\|_2^2} \, g_{\text{ref}} \;,
\end{equation}
where \(g_k\) is the current task gradient, \(g_{\text{ref}}\) is the reference gradient of past tasks, and 
\(\tilde{g}\) the projection of \(g_k\) orthogonal to \(g_{\text{ref}}\).

It should be noted that other relaxed variants such as $\epsilon$-SOFT-GEM \cite{hu2020gradient} allow a small negative margin:

\begin{equation}
\langle \tilde{g}, g_{\text{ref}} \rangle \ge -\epsilon \;,
\end{equation}
where $\epsilon$ is a small relaxation parameter that allows limited negative alignment between the updated directions. 

In our work, we use an A-GEM-style implementation to reduce computation and memory while still enforcing the core GEM constraint. We maintain a small episodic memory buffer (similar in sample size as we used for replay based methods), and to choose these samples we used k-means clustering approach to select more representative samples, Then computed a single averaged reference gradient, and use projection only when the update would increase memory loss. This makes the approach scalable for our FNO-based setting while preserving GEM’s forgetting-resistant behavior (see Fig.~\ref{fig:continual_learning_algorithms}f for an overview).

\subsubsection{PiggyBack}
% Another strategy to reduce forgetting in CL is to freeze the shared backbone and only learn small task-specific parameters \cite{mallya2018PiggyBack}. The PiggyBack method follows this idea by attaching a task-specific mask to a pretrained backbone rather than updating the backbone weights themselves~\cite{mallya2018PiggyBack}. The motivation is that large pretrained networks already contain rich representational structure, so instead of modifying parameters (\(\theta\)), the model learns which connections should be active for each task.

Another strategy to mitigate forgetting in CL is to freeze the shared backbone and learn only a small set of task-specific parameters. The PiggyBack method follows this idea by attaching a task-specific mask to a pretrained backbone rather than updating the backbone weights themselves~\cite{mallya2018PiggyBack}. The motivation is that large pretrained networks already contain rich representational structure, so instead of modifying the backbone parameters $\theta$, the model learns which connections should be active for each task.

Let the frozen backbone parameters be $\theta \in \mathbb{R}^D$. For task $k$, PiggyBack learns a point-wise mask with elements $m_{k,i}$, where $k$ denotes the task number in the sequential learning setting and $i$ indexes the backbone weights $\theta_i$. During training on task $k$, the backbone remains fixed and only the mask is optimized:
\begin{equation}
\nabla_\theta \mathcal{L}_k = 0 \;,
\qquad
\nabla_{m_{k,i}} \mathcal{L}_k \neq 0 \;.
\end{equation}
The mask modulates the backbone through elementwise multiplication, yielding the effective parameters for task $k$:
\begin{equation}
\theta_{k,i}^{\mathrm{eff}} = m_{k,i} \odot \theta_i  \;.
\label{eq:piggyback_effective_weights}
\end{equation}

In the original PiggyBack formulation, the mask values are binarized using a threshold $\tau$ to enforce sparsity and reduce storage~\cite{mallya2018PiggyBack}:
\begin{equation}
m_{k,i} =
\begin{cases}
1, & m_{k,i} \ge \tau \;, \\[4pt]
0, & m_{k,i} < \tau \;.
\end{cases}
\label{eq:piggyback_binarization}
\end{equation}
This produces a compact task-specific connectivity pattern while keeping the backbone unchanged, which helps avoid catastrophic forgetting.

In our experiments, we found that binarization destroyed fine-grained connectivity patterns learned during training and caused a large accuracy drop on new tasks. For that reason, we retain the continuous (non-binarized) mask values throughout. An overview of PiggyBack is shown in Fig.~\ref{fig:continual_learning_algorithms}g.

\subsubsection{Low-Rank Adaptation (LoRA)}

LoRA was originally introduced by Hu \textit{et al.}~\cite{hu2022lora} as a parameter-efficient method for adapting large pretrained models. The core idea is that when a model is fine-tuned, the change in the weight matrix is often low-rank. Instead of updating the full parameter matrix, LoRA freezes the backbone weights and learns a low-rank decomposition that represents only the task-specific update.

Let \(W \in \mathbb{R}^{k \times d}\) be a pretrained weight matrix. Standard fine-tuning updates \(W\) directly. LoRA instead reparameterizes the adapted weight as:

\begin{equation}
W' = W + \Delta W \;,
\end{equation}
where the update \(\Delta W\) is constrained to be low-rank:

\begin{equation}
\Delta W = A B, \qquad A \in \mathbb{R}^{k \times r}, \; B \in \mathbb{R}^{r \times d}, \; r \ll \min(k,d) \;.
\end{equation}
In this formulation, $\Delta W$ denotes a low-rank update to a weight matrix $W$, factorized as the product of two matrices $A \in \mathbb{R}^{k \times r}$ and $B \in \mathbb{R}^{r \times d}$. The dimensions $k$ and $d$ correspond to the output and input dimensions of $W$, respectively, while $r$ is the rank of the update. The condition $r \ll \min(k,d)$ enforces a low-rank constraint, significantly reducing the number of trainable parameters while preserving expressive capacity. Only the matrices \(A\) and \(B\) are trainable, while \(W\) remains frozen. During initialization, \(B\) is typically initialized with a small random distribution $B \sim \mathcal{N}(0, \sigma^2)$ and \(A\) is initialized to zero, ensuring that early training behaves like the frozen pretrained model.

To scale the adaptation effect, LoRA introduces a tunable factor \(\alpha\), modifying the above model to allows control of adaptation strength without changing model architecture:

\begin{equation}
W' = W + \frac{\alpha}{r} A B \;.
\end{equation}

In CL settings, LoRA allows each task to have its own low-rank parameters while sharing the frozen pretrained backbone. Formally, for task \(k\), the adapted weights become:

\begin{equation}
W_k' = W + \frac{\alpha}{r} A_k B_k \;,
\end{equation}
where \((A_k, B_k)\) are task-specific parameters. This avoids interference between tasks because past tasks' LoRA modules are not modified during training of new tasks, addressing catastrophic forgetting without requiring replay or explicit gradient constraints.

Recent CL studies have shown that LoRA-style modular adapters provide a good balance between memory efficiency, stability, and plasticity, making LoRA a strong option when the pretrained backbone is large and expressive. An overview of the LoRA method is shown in Fig.~\ref{fig:continual_learning_algorithms}h.

}

%\clearpage

\bibliographystyle{unsrt}
\bibliography{refs}

\begin{thebibliography}{100}

\bibitem{thiyagalingam2022scientific}
J.~Thiyagalingam, M.~Shankar, G.~Fox, and T.~Hey.
\newblock Scientific machine learning benchmarks.
\newblock {\em Nature Reviews Physics}, 4(6):413--420, 2022.

\bibitem{karniadakis2021physics}
G.~E. Karniadakis, I.~G. Kevrekidis, L.~Lu, P.~Perdikaris, S.~Wang, and
  L.~Yang.
\newblock Physics-informed machine learning.
\newblock {\em Nature Reviews Physics}, 3(6):422--440, 2021.

\bibitem{brunton2022data}
S.~L. Brunton and J.~N. Kutz.
\newblock {\em Data-driven science and engineering: Machine learning, dynamical
  systems, and control}.
\newblock Cambridge University Press, 2022.

\bibitem{li2020fourier}
Z.~Li, N.~Kovachki, K.~Azizzadenesheli, B.~Liu, K.~Bhattacharya, A.~Stuart, and
  A.~Anandkumar.
\newblock Fourier neural operator for parametric partial differential
  equations.
\newblock {\em arXiv preprint arXiv:2010.08895}, 2020.

\bibitem{lu2021learning}
L.~Lu, P.~Jin, G.~Pang, Z.~Zhang, and G.~E. Karniadakis.
\newblock Learning nonlinear operators via deeponet based on the universal
  approximation theorem of operators.
\newblock {\em Nature Machine Intelligence}, 3(3):218--229, 2021.

\bibitem{kovachki2023neural}
N.~Kovachki, Z.~Li, B.~Liu, K.~Azizzadenesheli, K.~Bhattacharya, A.~Stuart, and
  A.~Anandkumar.
\newblock Neural operator: Learning maps between function spaces with
  applications to {PDEs}.
\newblock {\em Journal of Machine Learning Research}, 24(89):1--97, 2023.

\bibitem{raissi2019physics}
M.~Raissi, P.~Perdikaris, and G.~E. Karniadakis.
\newblock Physics-informed neural networks: A deep learning framework for
  solving forward and inverse problems involving nonlinear partial differential
  equations.
\newblock {\em Journal of Computational Physics}, 378:686--707, 2019.

\bibitem{goswami2023physics}
S.~Goswami, A.~Bora, Y.~Yu, and G.~E. Karniadakis.
\newblock Physics-informed deep neural operator networks.
\newblock In {\em Machine Learning in Modeling and Simulation: Methods and
  Applications}, pages 219--254. Springer, 2023.

\bibitem{wang2021understanding}
S.~Wang, Y.~Teng, and P.~Perdikaris.
\newblock Understanding and mitigating gradient flow pathologies in
  physics-informed neural networks.
\newblock {\em SIAM Journal on Scientific Computing}, 43(5):A3055--A3081, 2021.

\bibitem{kiyani2025optimizer}
E.~Kiyani, K.~Shukla, J.~F. Urb{\'a}n, J.~Darbon, and G.~E. Karniadakis.
\newblock Which optimizer works best for physics-informed neural networks and
  {Kolmogorov-Arnold} networks?
\newblock {\em arXiv preprint arXiv:2501.16371}, 2025.

\bibitem{choi2025defining}
Y.~Choi, S.~W. Cheung, Y.~Kim, P.~Tsai, A.~N. Diaz, I.~Zanardi, S.~W. Chung,
  D.~M. Copeland, C.~Kendrick, W.~Anderson, et~al.
\newblock Defining foundation models for computational science: A call for
  clarity and rigor.
\newblock {\em arXiv preprint arXiv:2505.22904}, 2025.

\bibitem{menonscientific}
S.~S. Menon, T.~Mondal, S.~Brahmachary, A.~Panda, S.~M. Joshi, K.~Kalyanaraman,
  and A.~D. Jagtap.
\newblock On scientific foundation models: Rigorous definitions, key
  applications, and a survey.
\newblock {\em SSRN}, 2025.

\bibitem{quinonero2008dataset}
J.~Qui{\~n}onero-Candela, M.~Sugiyama, A.~Schwaighofer, and N.~D. Lawrence.
\newblock {\em Dataset shift in machine learning}.
\newblock {MIT} Press, 2008.

\bibitem{hendrycks2016baseline}
D.~Hendrycks and K.~Gimpel.
\newblock A baseline for detecting misclassified and out-of-distribution
  examples in neural networks.
\newblock {\em arXiv preprint arXiv:1610.02136}, 2016.

\bibitem{yang2024generalized}
J.~Yang, K.~Zhou, Y.~Li, and Z.~Liu.
\newblock Generalized out-of-distribution detection: A survey.
\newblock {\em International Journal of Computer Vision}, 132(12):5635--5662,
  2024.

\bibitem{yuan2022towards}
L.~Yuan, H.~S. Park, and E.~Lejeune.
\newblock Towards out of distribution generalization for problems in mechanics.
\newblock {\em Computer Methods in Applied Mechanics and Engineering},
  400:115569, 2022.

\bibitem{arzani2025interpreting}
A.~Arzani, L.~Yuan, P.~Newell, and B.~Wang.
\newblock Interpreting and generalizing deep learning in physics-based problems
  with functional linear models.
\newblock {\em Engineering with Computers}, 41(1):135--157, 2025.

\bibitem{church2021emerging}
Y.~M. K.~W.~Church, Z.~Chen.
\newblock Emerging trends: A gentle introduction to fine-tuning.
\newblock {\em Natural Language Engineering}, 27(6):763--778, 2021.

\bibitem{french1999catastrophic}
R.~M. French.
\newblock Catastrophic forgetting in connectionist networks.
\newblock {\em Trends in Cognitive Sciences}, 3(4):128--135, 1999.

\bibitem{gama2014survey}
J.~Gama, I.~{\v{Z}}liobait{\.e}, A.~Bifet, M.~Pechenizkiy, and A.~Bouchachia.
\newblock A survey on concept drift adaptation.
\newblock {\em ACM computing surveys {(CSUR)}}, 46(4):1--37, 2014.

\bibitem{liang2017enhancing}
S.~Liang, Y.~Li, and R.~Srikant.
\newblock Enhancing the reliability of out-of-distribution image detection in
  neural networks.
\newblock {\em arXiv preprint arXiv:1706.02690}, 2017.

\bibitem{ovadia2019can}
Y.~Ovadia, E.~Fertig, J.~Ren, Z.~Nado, D.~Sculley, S.~Nowozin, J.~Dillon,
  B.~Lakshminarayanan, and J.~Snoek.
\newblock Can you trust your model's uncertainty? evaluating predictive
  uncertainty under dataset shift.
\newblock {\em Advances in Neural Information Processing Systems}, 32, 2019.

\bibitem{koh2021wilds}
P.~W. Koh, S.~Sagawa, H.~Marklund, S.~M. Xie, M.~Zhang, A.~Balsubramani, W.~Hu,
  M.~Yasunaga, R.~L. Phillips, I.~Gao, et~al.
\newblock {Wilds}: A benchmark of in-the-wild distribution shifts.
\newblock In {\em International Conference on Machine Learning}, pages
  5637--5664. PMLR, 2021.

\bibitem{hadsell2020embracing}
R.~Hadsell, D.~Rao, A.~A. Rusu, and R.~Pascanu.
\newblock Embracing change: Continual learning in deep neural networks.
\newblock {\em Trends in Cognitive Sciences}, 24(12):1028--1040, 2020.

\bibitem{de2021continual}
M.~De~Lange, R.~Aljundi, M.~Masana, S.~Parisot, X.~Jia, A.~Leonardis,
  G.~Slabaugh, and T.~Tuytelaars.
\newblock A continual learning survey: Defying forgetting in classification
  tasks.
\newblock {\em IEEE Transactions on Pattern Analysis and Machine Intelligence},
  44(7):3366--3385, 2021.

\bibitem{wang2024comprehensive}
L.~Wang, X.~Zhang, H.~Su, and J.~Zhu.
\newblock A comprehensive survey of continual learning: Theory, method and
  application.
\newblock {\em IEEE Transactions on Pattern Analysis and Machine Intelligence},
  46(8):5362--5383, 2024.

\bibitem{vandeven2019three}
G.~M. Van~de Ven and A.~S. Tolias.
\newblock Three scenarios for continual learning.
\newblock {\em arXiv preprint arXiv:1904.07734}, 2019.

\bibitem{li2017learning}
Z.~Li and D.~Hoiem.
\newblock Learning without forgetting.
\newblock {\em IEEE Transactions on Pattern Analysis and Machine Intelligence},
  40(12):2935--2947, 2017.

\bibitem{kirkpatrick2017overcoming}
J.~Kirkpatrick, R.~Pascanu, N.~Rabinowitz, J.~Veness, G.~Desjardins, A.~A.
  Rusu, K.~Milan, J.~Quan, T.~Ramalho, A.~Grabska-Barwinska, et~al.
\newblock Overcoming catastrophic forgetting in neural networks.
\newblock {\em Proceedings of the National Academy of Sciences},
  114(13):3521--3526, 2017.

\bibitem{rolnick2019experience}
D.~Rolnick, A.~Ahuja, J.~Schwarz, T.~Lillicrap, and G.~Wayne.
\newblock Experience replay for continual learning.
\newblock {\em Advances in Neural Information Processing Systems}, 32, 2019.

\bibitem{shin2017continual}
H.~Shin, J.~K. Lee, J.~Kim, and J.~Kim.
\newblock Continual learning with deep generative replay.
\newblock {\em Advances in Neural Information Processing Systems}, 30, 2017.

\bibitem{farajtabar2020ogd}
M.~Farajtabar, N.~Azizan, A.~Mott, and A.~Li.
\newblock Orthogonal gradient descent for continual learning.
\newblock {\em arXiv preprint arXiv:1910.07104}, 2020.

\bibitem{lopez2017gradient}
D.~Lopez-Paz and M.~Ranzato.
\newblock Gradient episodic memory for continual learning.
\newblock {\em Advances in Neural Information Processing Systems}, 30, 2017.

\bibitem{saha2021gradient}
G.~Saha, I.~Garg, and K.~Roy.
\newblock Gradient projection memory for continual learning.
\newblock {\em arXiv preprint arXiv:2103.09762}, 2021.

\bibitem{mallya2018packnet}
S.~Lazebnik A.~Mallya.
\newblock Packnet: Adding multiple tasks to a single network by iterative
  pruning.
\newblock In {\em Proceedings of the IEEE conference on Computer Vision and
  Pattern Recognition}, pages 7765--7773, 2018.

\bibitem{mallya2018PiggyBack}
A.~Mallya, D.~Davis, and S.~Lazebnik.
\newblock Piggyback: Adapting a single network to multiple tasks by learning to
  mask weights.
\newblock In {\em Proceedings of the European Conference on Computer Vision
  {(ECCV)}}, pages 67--82, 2018.

\bibitem{rusu2016progressive}
A.~A. Rusu, N.~C. Rabinowitz, G.~Desjardins, H.~Soyer, J.~Kirkpatrick,
  K.~Kavukcuoglu, R.~Pascanu, and R.~Hadsell.
\newblock Progressive neural networks.
\newblock {\em arXiv preprint arXiv:1606.04671}, 2016.

\bibitem{rao2019continual}
D.~Rao, F.~Visin, A.~Rusu, R.~Pascanu, Y.~W. Teh, and R.~Hadsell.
\newblock Continual unsupervised representation learning.
\newblock {\em Advances in Neural Information Processing Systems}, 32, 2019.

\bibitem{cha2021co2l}
H.~Cha, J.~Lee, and J.~Shin.
\newblock {Co2l}: Contrastive continual learning.
\newblock In {\em Proceedings of the IEEE/CVF International Conference on
  Computer Vision}, pages 9516--9525, 2021.

\bibitem{Zhou_2024}
D.~Zhou, Q.~Wang, Z.~Qi, H.~Ye, D.~Zhan, and Z.~Liu.
\newblock Class-incremental learning: A survey.
\newblock {\em IEEE Transactions on Pattern Analysis and Machine Intelligence},
  2024.

\bibitem{van2022three}
G.~M. Van~de Ven, T.~Tuytelaars, and A.~S. Tolias.
\newblock Three types of incremental learning.
\newblock {\em Nature Machine Intelligence}, 4(12):1185--1197, 2022.

\bibitem{he2021clear}
Y.~He and B.~Sick.
\newblock {CLeaR}: An adaptive continual learning framework for regression
  tasks.
\newblock {\em AI Perspectives}, 3(1):2, 2021.

\bibitem{besnard2024forecasting}
Q.~Besnard and N.~Ragot.
\newblock Continual learning for time series forecasting: A first survey.
\newblock {\em Engineering Proceedings}, 68(1):49, 2024.

\bibitem{ding2024understanding}
M.~Ding, K.~Ji, D.~Wang, and J.~Xu.
\newblock Understanding forgetting in continual learning with linear
  regression.
\newblock {\em arXiv preprint arXiv:2405.17583}, 2024.

\bibitem{howard2024multifidelity}
A.~Howard, Y.~Fu, and P.~Stinis.
\newblock A multifidelity approach to continual learning for physical systems.
\newblock {\em Machine Learning: Science and Technology}, 5(2):025042, 2024.

\bibitem{tripura2023ncwno}
T.~Tripura and S.~Chakraborty.
\newblock Neural combinatorial wavelet neural operator for catastrophic
  forgetting free in-context operator learning of multiple partial differential
  equations.
\newblock {\em Computer Physics Communications}, page 109882, 2025.

\bibitem{samuel2025cl3d}
K.~M. Samuel and F.~Ahmed.
\newblock Continual learning strategies for {3D} engineering regression
  problems: A benchmarking study.
\newblock {\em Journal of Computing and Information Science in Engineering},
  25(10):101003, 2025.

\bibitem{kang2024continual}
H.~Kang, J.~Yoon, S.~J. Hwang, and C.~D. Yoo.
\newblock Continual learning: Forget-free winning subnetworks for video
  representations.
\newblock {\em IEEE Transactions on Pattern Analysis and Machine Intelligence},
  2024.

\bibitem{mahmoudi2021story}
M.~Mahmoudi, A.~Farghadan, D.~R. McConnell, A.~J. Barker, J.~J. Wentzel, M.~J.
  Budoff, and A.~Arzani.
\newblock The story of wall shear stress in coronary artery atherosclerosis:
  biochemical transport and mechanotransduction.
\newblock {\em Journal of Biomechanical Engineering}, 143(4):041002, 2021.

\bibitem{raghavan2005quantified}
M.~L. Raghavan, B.~Ma, and R.~E. Harbaugh.
\newblock Quantified aneurysm shape and rupture risk.
\newblock {\em Journal of Neurosurgery}, 102(2):355--362, 2005.

\bibitem{womersley1955method}
J.~R. Womersley.
\newblock Method for the calculation of velocity, rate of flow and viscous drag
  in arteries when the pressure gradient is known.
\newblock {\em The Journal of Physiology}, 127(3):553, 1955.

\bibitem{ku1985pulsatile}
D.~N. Ku, D.~P. Giddens, C.~K. Zarins, and S.~Glagov.
\newblock Pulsatile flow and atherosclerosis in the human carotid bifurcation.
  positive correlation between plaque location and low oscillating shear
  stress.
\newblock {\em Arteriosclerosis: An Official Journal of the American Heart
  Association, Inc.}, 5(3):293--302, 1985.

\bibitem{malek1999hemodynamic}
A.~M. Malek, S.~L. Alper, and S.~Izumo.
\newblock Hemodynamic shear stress and its role in atherosclerosis.
\newblock {\em Jama}, 282(21):2035--2042, 1999.

\bibitem{arzani2018accounting}
A.~Arzani.
\newblock Accounting for residence-time in blood rheology models: do we really
  need non-newtonian blood flow modelling in large arteries?
\newblock {\em Journal of The Royal Society Interface}, 15(146):20180486, 2018.

\bibitem{staarmann2019shear}
B.~Staarmann, M.~Smith, and C.~J. Prestigiacomo.
\newblock Shear stress and aneurysms: a review.
\newblock {\em Neurosurgical Focus}, 47(1):E2, 2019.

\bibitem{ayachit2015paraview}
U.~Ayachit.
\newblock {\em The paraview guide: a parallel visualization application}.
\newblock Kitware, Inc., 2015.

\bibitem{czarnecki2017sobolev}
W.~M. Czarnecki, S.~Osindero, M.~Jaderberg, G.~Swirszcz, and R.~Pascanu.
\newblock Sobolev training for neural networks.
\newblock {\em Advances in Neural Information Processing Systems}, 30, 2017.

\bibitem{goodfellow2013empirical}
I.~J. Goodfellow, M.~Mirza, D.~Xiao, A.~Courville, and Y.~Bengio.
\newblock An empirical investigation of catastrophic forgetting in
  gradient-based neural networks.
\newblock {\em arXiv preprint arXiv:1312.6211}, 2013.

\bibitem{luo2023empirical}
Y.~Luo, Z.~Yang, F.~Meng, Y.~Li, J.~Zhou, and Y.~Zhang.
\newblock An empirical study of catastrophic forgetting in large language
  models during continual fine-tuning.
\newblock {\em arXiv preprint arXiv:2308.08747}, 2023.

\bibitem{pratt1992discriminability}
L.~Y. Pratt.
\newblock Discriminability-based transfer between neural networks.
\newblock {\em Advances in Neural Information Processing Systems}, 5, 1992.

\bibitem{buzzega2020dark}
P.~Buzzega, M.~Boschini, A.~Porrello, D.~Abati, and S.~Calderara.
\newblock Dark experience for general continual learning: a strong, simple
  baseline.
\newblock {\em Advances in Neural Information Processing Systems},
  33:15920--15930, 2020.

\bibitem{prabhu2020gdumb}
A.~Prabhu, P.~H. Torr, and P.~K. Dokania.
\newblock Gdumb: A simple approach that questions our progress in continual
  learning.
\newblock In {\em European Conference on Computer Vision}, pages 524--540.
  Springer, 2020.

\bibitem{chaudhry2018efficient}
A.~Chaudhry, M.~Ranzato, M.~Rohrbach, and M.~Elhoseiny.
\newblock Efficient lifelong learning with {A-GEM}.
\newblock {\em arXiv preprint arXiv:1812.00420}, 2018.

\bibitem{hu2022lora}
E.~J. Hu, Y.~Shen, P.~Wallis, Z.~Allen-Zhu, Y.~Li, S.~Wang, L.~Wang, W.~Chen,
  et~al.
\newblock {LoRA}: Low-rank adaptation of large language models.
\newblock {\em ICLR}, 1(2):3, 2022.

\bibitem{zhang2020side}
J.~O. Zhang, A.~Sax, A.~Zamir, L.~Guibas, and J.~Malik.
\newblock Side-tuning: a baseline for network adaptation via additive side
  networks.
\newblock In {\em European Conference on Computer Vision}, pages 698--714.
  Springer, 2020.

\bibitem{rahaman2019spectral}
N.~Rahaman, A.~Baratin, D.~Arpit, F.~Draxler, M.~Lin, F.~Hamprecht, Y.~Bengio,
  and A.~Courville.
\newblock On the spectral bias of neural networks.
\newblock In {\em International Conference on Machine Learning}, pages
  5301--5310. PMLR, 2019.

\bibitem{serra2018overcoming}
J.~Serra, D.~Suris, M.~Miron, and A.~Karatzoglou.
\newblock Overcoming catastrophic forgetting with hard attention to the task.
\newblock In {\em International Conference on Machine Learning}, pages
  4548--4557. PMLR, 2018.

\bibitem{ha2016hypernetworks}
D.~Ha, A.~Dai, and Q.~V. Le.
\newblock Hypernetworks.
\newblock {\em arXiv preprint arXiv:1609.09106}, 2016.

\bibitem{aljundi2017expert}
R.~Aljundi, P.~Chakravarty, and T.~Tuytelaars.
\newblock Expert gate: Lifelong learning with a network of experts.
\newblock In {\em Proceedings of the IEEE Conference on Computer Vision and
  Pattern Recognition}, pages 3366--3375, 2017.

\bibitem{zhu2024tame}
H.~Zhu, M.~Majzoubi, A.~Jain, and A.~Choromanska.
\newblock {TAME}: Task agnostic continual learning using multiple experts.
\newblock In {\em Proceedings of the IEEE/CVF Conference on Computer Vision and
  Pattern Recognition}, pages 4139--4148, 2024.

\bibitem{kim2022continual}
G.~Kim, S.~Esmaeilpour, C.~Xiao, and B.~Liu.
\newblock Continual learning based on ood detection and task masking.
\newblock In {\em Proceedings of the IEEE/CVF Conference on Computer Vision and
  Pattern Recognition}, pages 3856--3866, 2022.

\bibitem{wortsman2020supermasks}
M.~Wortsman, V.~Ramanujan, R.~Liu, A.~Kembhavi, M.~Rastegari, J.~Yosinski, and
  A.~Farhadi.
\newblock Supermasks in superposition.
\newblock {\em Advances in Neural Information Processing Systems},
  33:15173--15184, 2020.

\bibitem{scholkopf1998nonlinear}
B.~Sch{\"o}lkopf, A.~Smola, and K.~R. M{\"u}ller.
\newblock Nonlinear component analysis as a kernel eigenvalue problem.
\newblock {\em Neural Computation}, 10(5):1299--1319, 1998.

\bibitem{scholkopf1999kpca}
B.~Sch{\"o}lkopf, R.~C. Williamson, A.~Smola, J.~Shawe-Taylor, and J.~Platt.
\newblock Support vector method for novelty detection.
\newblock {\em Advances in Neural Information Processing Systems}, 12, 1999.

\bibitem{lee2018simple}
K.~Lee, K.~Lee, H.~Lee, and J.~Shin.
\newblock A simple unified framework for detecting out-of-distribution samples
  and adversarial attacks.
\newblock {\em Advances in Neural Information Processing Systems}, 31, 2018.

\bibitem{liu2020energy}
W.~Liu, X.~Wang, J.~Owens, and Y.~Li.
\newblock Energy-based out-of-distribution detection.
\newblock {\em Advances in Neural Information Processing Systems},
  33:21464--21475, 2020.

\bibitem{fang2024kernel}
K.~Fang, Q.~Tao, K.~Lv, M.~He, X.~Huang, and J.~Yang.
\newblock Kernel {PCA} for out-of-distribution detection.
\newblock {\em Advances in Neural Information Processing Systems},
  37:134317--134344, 2024.

\bibitem{rahimi2007random}
A.~Rahimi and B.~Recht.
\newblock Random features for large-scale kernel machines.
\newblock {\em Advances in Neural Information Processing Systems}, 20, 2007.

\bibitem{takamoto2022pdebench}
M.~Takamoto, T.~Praditia, R.~Leiteritz, D.~MacKinlay, F.~Alesiani,
  D.~Pfl{\"u}ger, and M.~Niepert.
\newblock Pdebench: An extensive benchmark for scientific machine learning.
\newblock {\em Advances in Neural Information Processing Systems},
  35:1596--1611, 2022.

\bibitem{semmelrock2025reproducibility}
H.~Semmelrock, T.~Ross-Hellauer, S.~Kopeinik, D.~Theiler, A.~Haberl,
  S.~Thalmann, and D.~Kowald.
\newblock Reproducibility in machine-learning-based research: Overview,
  barriers, and drivers.
\newblock {\em AI Magazine}, 46(2):e70002, 2025.

\bibitem{chaudhry2018riemannian}
A.~Chaudhry, P.~K. Dokania, T.~Ajanthan, and P.~H. Torr.
\newblock Riemannian walk for incremental learning: Understanding forgetting
  and intransigence.
\newblock In {\em Proceedings of the European Conference on Computer Vision
  (ECCV)}, pages 532--547, 2018.

\bibitem{bengio2017deep}
Y.~Bengio, I.~Goodfellow, A.~Courville, et~al.
\newblock {\em Deep learning}, volume~1.
\newblock MIT press Cambridge, MA, USA, 2017.

\bibitem{li2023large}
A.~Li, C.~Zhang, F.~Xiao, C.~Fan, Y.~Deng, and D.~Wang.
\newblock Large-scale comparison and demonstration of continual learning for
  adaptive data-driven building energy prediction.
\newblock {\em Applied Energy}, 347:121481, 2023.

\bibitem{resani2024miracle3d}
H.~Resani and B.~Nasihatkon.
\newblock {MIRACLE3D}: Memory-efficient integrated robust approach for
  continual learning on point clouds via shape model construction.
\newblock {\em arXiv preprint arXiv:2410.06418}, 2024.

\bibitem{raisch2025adapting}
F.~Raisch, M.~Langtry, F.~Koch, R.~Choudhary, C.~Goebel, and B.~Tischler.
\newblock Adapting to change: A comparison of continual and transfer learning
  for modeling building thermal dynamics under concept drifts.
\newblock {\em Energy and Buildings}, page 116868, 2025.

\bibitem{liang2024inflora}
Y.-S. Liang and W.-J. Li.
\newblock Inflora: Interference-free low-rank adaptation for continual
  learning.
\newblock In {\em Proceedings of the IEEE/CVF Conference on Computer Vision and
  Pattern Recognition}, pages 23638--23647, 2024.

\bibitem{he2025cl}
J.~He, Z.~Duan, and F.~Zhu.
\newblock {CL-LoRA}: Continual low-rank adaptation for rehearsal-free
  class-incremental learning.
\newblock In {\em Proceedings of the Computer Vision and Pattern Recognition
  Conference}, pages 30534--30544, 2025.

\bibitem{wei2025online}
X.~Wei, G.~Li, and R.~Marculescu.
\newblock Online-lora: Task-free online continual learning via low rank
  adaptation.
\newblock In {\em 2025 IEEE/CVF Winter Conference on Applications of Computer
  Vision (WACV)}, pages 6634--6645. IEEE, 2025.

\bibitem{smith2023continual}
J.~S. Smith, Y.-C. Hsu, L.~Zhang, T.~Hua, Z.~Kira, Y.~Shen, and H.~Jin.
\newblock Continual diffusion: Continual customization of text-to-image
  diffusion with c-lora.
\newblock {\em arXiv preprint arXiv:2304.06027}, 2023.

\bibitem{wang2022dualprompt}
Z.~Wang, Z.~Zhang, S.~Ebrahimi, R.~Sun, H.~Zhang, C.-Y. Lee, X.~Ren, G.~Su,
  V.~Perot, and J.~et~al. Dy.
\newblock Dualprompt: Complementary prompting for rehearsal-free continual
  learning.
\newblock In {\em European Conference on Computer Vision}, pages 631--648.
  Springer, 2022.

\bibitem{smith2023coda}
J.~S. Smith, L.~Karlinsky, V.~Gutta, P.~Cascante-Bonilla, D.~Kim, A.~Arbelle,
  R.~Panda, R.~Feris, and Z.~Kira.
\newblock Coda-prompt: Continual decomposed attention-based prompting for
  rehearsal-free continual learning.
\newblock In {\em Proceedings of the IEEE/CVF Conference on Computer Vision and
  Pattern Recognition}, pages 11909--11919, 2023.

\bibitem{gao2024parameter}
Z.~Gao, Q.~Wang, A.~Chen, Z.~Liu, B.~Wu, L.~Chen, and J.~Li.
\newblock Parameter-efficient fine-tuning with discrete fourier transform.
\newblock {\em arXiv preprint arXiv:2405.03003}, 2024.

\bibitem{zhang2024spectral}
F.~Zhang and M.~Pilanci.
\newblock Spectral adapter: Fine-tuning in spectral space.
\newblock {\em arXiv preprint arXiv:2405.13952}, 2024.

\bibitem{zhang2025f}
H.~Zhang, C.~Kang, Y.~Wang, and D.~Zou.
\newblock F-adapter: Frequency-adaptive parameter-efficient fine-tuning in
  scientific machine learning.
\newblock {\em arXiv preprint arXiv:2509.23173}, 2025.

\bibitem{yu2024boosting}
J.~Yu, Y.~Zhuge, L.~Zhang, P.~Hu, D.~Wang, H.~Lu, and Y.~He.
\newblock Boosting continual learning of vision-language models via
  mixture-of-experts adapters.
\newblock In {\em Proceedings of the IEEE/CVF Conference on Computer Vision and
  Pattern Recognition}, pages 23219--23230, 2024.

\bibitem{zohdi2020machine}
T.~Zohdi.
\newblock A machine-learning framework for rapid adaptive digital-twin based
  fire-propagation simulation in complex environments.
\newblock {\em Computer Methods in Applied Mechanics and Engineering},
  363:112907, 2020.

\bibitem{tasmurzayev2025digital}
N.~Tasmurzayev, B.~Amangeldy, B.~Imanbek, Z.~Baigarayeva, T.~Imankulov,
  G.~Dikhanbayeva, I.~Amangeldi, and S.~Sharipova.
\newblock Digital cardiovascular twins, {AI} agents, and sensor data: A
  narrative review from system architecture to proactive heart health.
\newblock {\em Sensors}, 25(17):5272, 2025.

\bibitem{ziegler2015synaptic}
L.~Ziegler, F.~Zenke, D.~B. Kastner, and W.~Gerstner.
\newblock Synaptic consolidation: from synapses to behavioral modeling.
\newblock {\em Journal of Neuroscience}, 35(3):1319--1334, 2015.

\bibitem{fujita2022fisher}
K.~Fujita, K.~Okada, and K.~Katahira.
\newblock The fisher information matrix: A tutorial for calculation for
  decision making models.
\newblock 2022.

\bibitem{van2025computation}
G.~M. van~de Ven.
\newblock On the computation of the fisher information in continual learning.
\newblock {\em arXiv preprint arXiv:2502.11756}, 2025.

\bibitem{spall2003monte}
J.~C. Spall.
\newblock Monte carlo-based computation of the fisher information matrix in
  nonstandard settings.
\newblock In {\em Proceedings of the 2003 American Control Conference, 2003.},
  volume~5, pages 3797--3802. IEEE, 2003.

\bibitem{soen2024trade}
A.~Soen and K.~Sun.
\newblock Trade-offs of diagonal fisher information matrix estimators.
\newblock {\em Advances in Neural Information Processing Systems},
  37:5870--5912, 2024.

\bibitem{aich2021elastic}
A.~Aich.
\newblock Elastic weight consolidation {EWC}: Nuts and bolts.
\newblock {\em arXiv preprint arXiv:2105.04093}, 2021.

\bibitem{gou2021knowledge}
J.~Gou, B.~Yu, S.~J. Maybank, and D.~Tao.
\newblock Knowledge distillation: A survey.
\newblock {\em International Journal of Computer Vision}, 129(6):1789--1819,
  2021.

\bibitem{hinton2015distilling}
G.~Hinton, O.~Vinyals, and J.~Dean.
\newblock Distilling the knowledge in a neural network.
\newblock {\em arXiv preprint arXiv:1503.02531}, 2015.

\bibitem{kang2021data}
M.~Kang and S.~Kang.
\newblock Data-free knowledge distillation in neural networks for regression.
\newblock {\em Expert Systems with Applications}, 175:114813, 2021.

\bibitem{saputra2019distilling}
M.~R.~U. Saputra, P.~P. De~Gusmao, Y.~Almalioglu, A.~Markham, and N.~Trigoni.
\newblock Distilling knowledge from a deep pose regressor network.
\newblock In {\em Proceedings of the IEEE/CVF International Conference on
  Computer Vision}, pages 263--272, 2019.

\bibitem{bagus2021investigation}
B.~Bagus and A.~Gepperth.
\newblock An investigation of replay-based approaches for continual learning.
\newblock In {\em 2021 International Joint Conference on Neural Networks
  {(IJCNN)}}, pages 1--9. IEEE, 2021.

\bibitem{chaudhry2019tiny}
A.~Chaudhry, M.~Rohrbach, M.~Elhoseiny, T.~Ajanthan, P.~K. Dokania, P.~H. Torr,
  and M.~Ranzato.
\newblock On tiny episodic memories in continual learning.
\newblock {\em arXiv preprint arXiv:1902.10486}, 2019.

\bibitem{wu2018memory}
C.~Wu, L.~Herranz, X.~Liu, J.~Van De~Weijer, B.~Raducanu, et~al.
\newblock Memory replay {GANs}: Learning to generate new categories without
  forgetting.
\newblock {\em Advances in Neural Information Processing Systems}, 31, 2018.

\bibitem{aljundi2019gradient}
R.~Aljundi, E.~Belilovsky, T.~Tuytelaars, L.~Charlin, M.~Caccia, M.~Lin, and
  L.~Page-Caccia.
\newblock Online continual learning with maximally interfered retrieval.
\newblock In {\em Advances in Neural Information Processing Systems
  {(NeurIPS)}}, volume~32, 2019.

\bibitem{tiwari2022gcr}
R.~Tiwari, K.~Killamsetty, R.~Iyer, and P.~Shenoy.
\newblock {GCR}: Gradient coreset based replay buffer selection for continual
  learning.
\newblock In {\em Proceedings of the IEEE/CVF Conference on Computer Vision and
  Pattern Recognition}, pages 99--108, 2022.

\bibitem{vitter1985random}
J.~S. Vitter.
\newblock Random sampling with a reservoir.
\newblock {\em ACM Transactions on Mathematical Software {(TOMS)}},
  11(1):37--57, 1985.

\bibitem{rebuffi2017icarl}
S.~Rebuffi, A.~Kolesnikov, G.~Sperl, and C.~H. Lampert.
\newblock {ICARL}: Incremental classifier and representation learning.
\newblock In {\em Proceedings of the IEEE conference on Computer Vision and
  Pattern Recognition}, pages 2001--2010, 2017.

\bibitem{chaudhry2020continual}
A.~Chaudhry, N.~Khan, P.~Dokania, and P.~Torr.
\newblock Continual learning in low-rank orthogonal subspaces.
\newblock {\em Advances in Neural Information Processing Systems},
  33:9900--9911, 2020.

\bibitem{dong2022gdod}
X.~Dong, R.~Wu, C.~Xiong, H.~Li, L.~Cheng, Y.~He, S.~Qian, J.~Cao, and L.~Mo.
\newblock {GDOD}: Effective gradient descent using orthogonal decomposition for
  multi-task learning.
\newblock In {\em Proceedings of the 31st ACM International Conference on
  Information \& Knowledge Management}, pages 386--395, 2022.

\bibitem{tang2021gradient}
S.~Tang, P.~Su, D.~Chen, and W.~Ouyang.
\newblock Gradient regularized contrastive learning for continual domain
  adaptation.
\newblock In {\em Proceedings of the AAAI Conference on Artificial
  Intelligence}, volume~35, pages 2665--2673, 2021.

\bibitem{wright2023sketchogd}
B.~Wright, Y.~Min, J.~Bernstein, and N.~Azizan.
\newblock {SketchOGD}: Memory-efficient continual learning.
\newblock {\em arXiv preprint arXiv:2305.16424}, 2023.

\bibitem{nesterov2022gradient}
Y.~Nesterov and M.~I. Florea.
\newblock Gradient methods with memory.
\newblock {\em Optimization Methods and Software}, 37(3):936--953, 2022.

\bibitem{guo2020improved}
Y.~Guo, M.~Liu, T.~Yang, and T.~Rosing.
\newblock Improved schemes for episodic memory-based lifelong learning.
\newblock {\em Advances in Neural Information Processing Systems},
  33:1023--1035, 2020.

\bibitem{hu2020gradient}
G.~Hu, W.~Zhang, H.~Ding, and W.~Zhu.
\newblock Gradient episodic memory with a soft constraint for continual
  learning.
\newblock {\em arXiv preprint arXiv:2011.07801}, 2020.

\end{thebibliography}

\includepdf[pages=-]{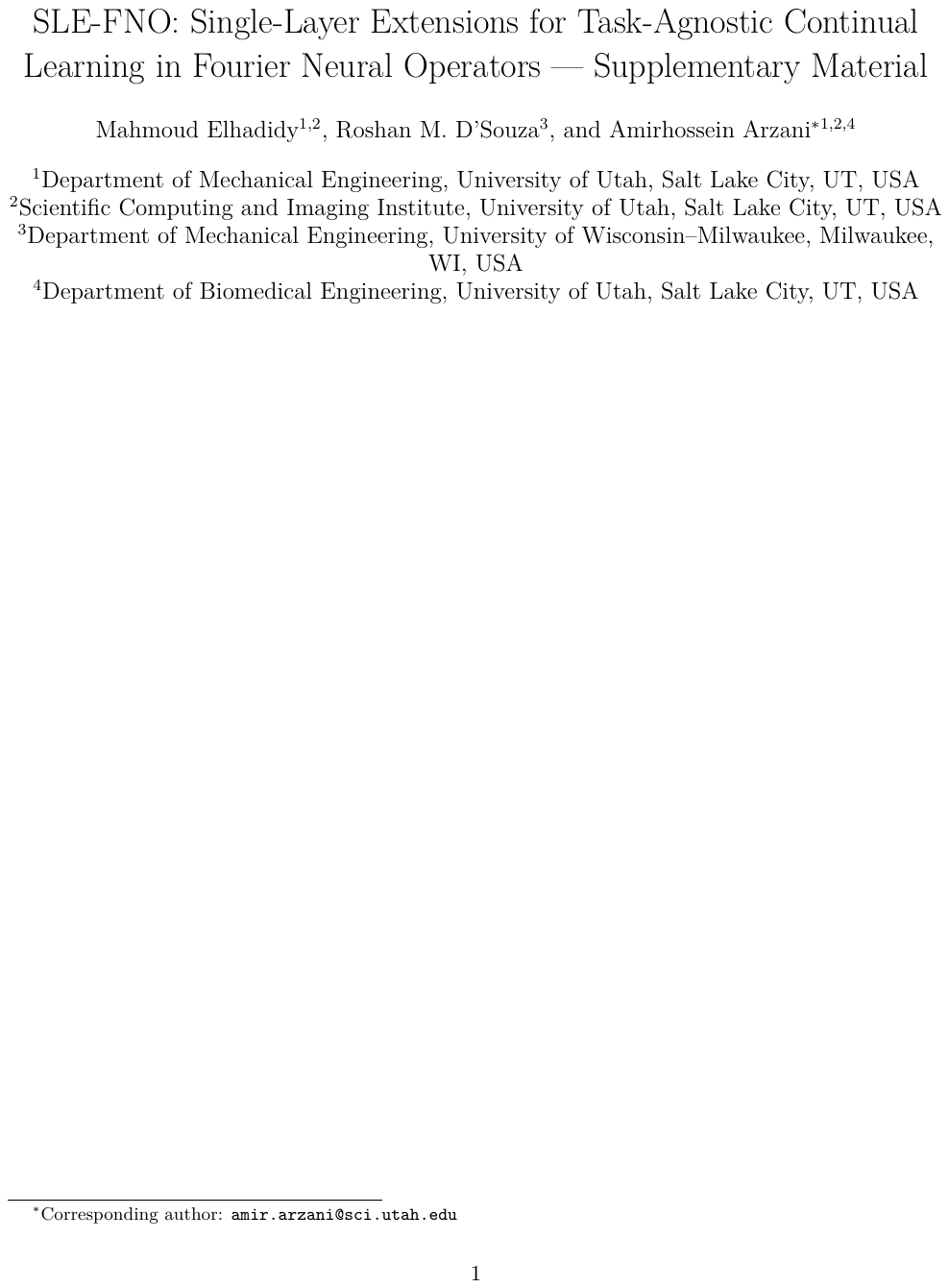}

\end{document}